\documentclass[10pt,twocolumn,letterpaper]{article}

\usepackage{iccv}
\usepackage{times}
\usepackage{epsfig}
\usepackage{graphicx}
\usepackage{amsmath}
\usepackage{amssymb}

\usepackage{subfig}
\usepackage{lipsum}
\usepackage{dsfont}
\usepackage{multicol, booktabs}
\usepackage{aliascnt}

\usepackage[pagebackref=true,breaklinks=true,letterpaper=true,colorlinks,bookmarks=false]{hyperref}
\usepackage[nameinlink,noabbrev]{cleveref}
\newaliascnt{inq}{equation}
\aliascntresetthe{inq}
\crefname{inq}{inequality}{inequalities}
\creflabelformat{inq}{#2\textup{(#1)}#3}
\usepackage{appendix}
\makeatletter

\def\endineq{\eqno \hbox{\@eqnnum}$$\@ignoretrue}
\makeatother

\usepackage{xcolor}
\usepackage[normalem]{ulem}

\newsavebox{\measurebox}

\usepackage{color, colortbl}
\definecolor{Gray}{gray}{0.9}
\newcolumntype{g}{>{\columncolor{Gray}}c}

\iccvfinalcopy

\begin{document}

%%%%%%%%% TITLE
\title{Entropy Maximization and Meta Classification for\\ Out-of-Distribution Detection in Semantic Segmentation}

\author{Robin Chan, Matthias Rottmann and Hanno Gottschalk\\
IZMD, Faculty of Mathematics and Natural Sciences, University of Wuppertal\\
{\tt\small \{\href{mailto:rchan@uni-wuppertal.de}{rchan},\href{mailto:rottmann@uni-wuppertal.de}{rottmann},\href{mailto:hgottsch@uni-wuppertal.de}{hgottsch}\}@uni-wuppertal.de}
}
\maketitle

\begin{abstract}
Deep neural networks (DNNs) for the semantic segmentation of images are usually trained to operate on a predefined closed set of object classes. This is in contrast to the ``open world'' setting where DNNs are envisioned to be deployed to. From a functional safety point of view, the ability to detect so-called ``out-of-distribution'' (OoD) samples, \ie, objects outside of a DNN's semantic space, is crucial for many applications such as automated driving.
A natural baseline approach to OoD detection is to threshold on the pixel-wise softmax entropy. We present a two-step procedure that significantly improves that approach. Firstly, we utilize samples from the COCO dataset as OoD proxy and introduce a second training objective to maximize the softmax entropy on these samples. Starting from pretrained semantic segmentation networks we re-train a number of DNNs on different in-distribution datasets and consistently observe improved OoD detection performance when evaluating on completely disjoint OoD datasets.
Secondly, we perform a transparent post-processing step to discard false positive OoD samples by so-called ``meta classification.'' To this end, we apply linear models to a set of hand-crafted metrics derived from the DNN's softmax probabilities. In our experiments we consistently observe a clear additional gain in OoD detection performance, cutting down the number of detection errors by 52\% when comparing the best baseline with our results.
We achieve this improvement sacrificing only marginally in original segmentation performance. Therefore, our method contributes to safer DNNs with more reliable overall system performance.
\end{abstract}

\section{Introduction}

\begin{figure}[t]
    \captionsetup[subfigure]{labelformat=empty}
    \centering
    \subfloat[Baseline segmentation mask]{\includegraphics[width=0.23\textwidth]{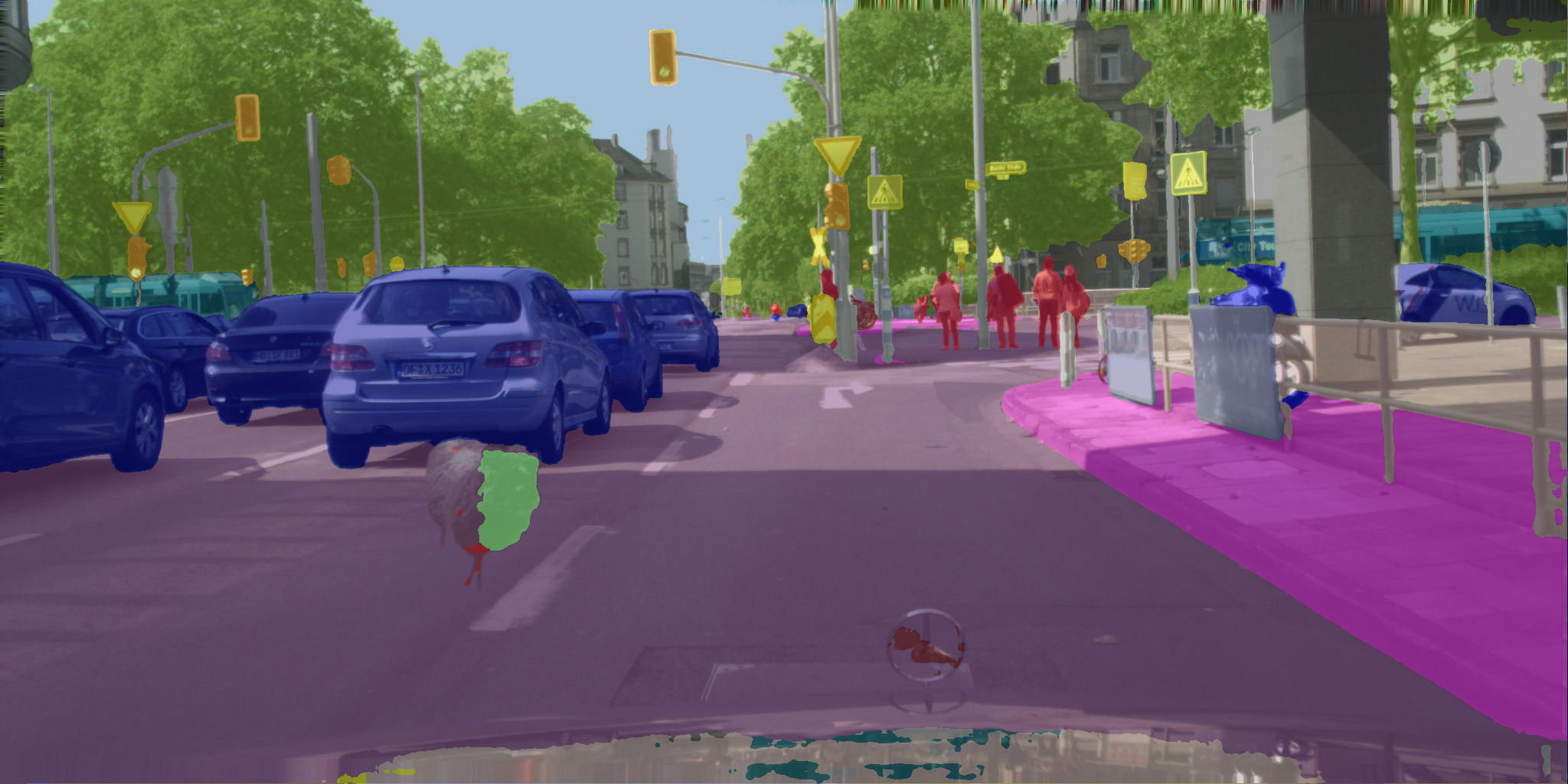}}~
    \subfloat[Baseline entropy heatmap]{\includegraphics[width=0.23\textwidth]{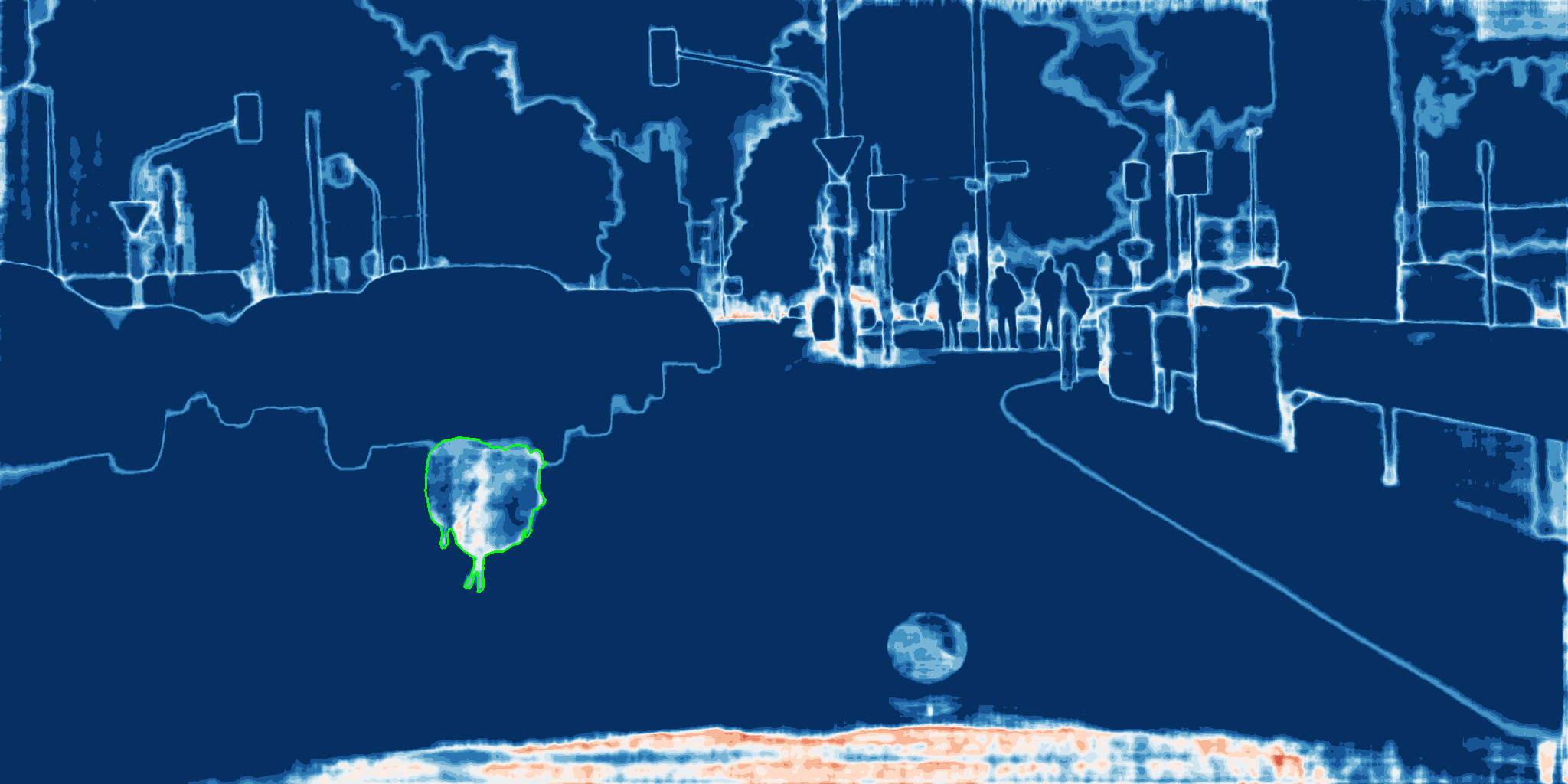}} \\
    \subfloat[Our segmentation mask]{\includegraphics[width=0.23\textwidth]{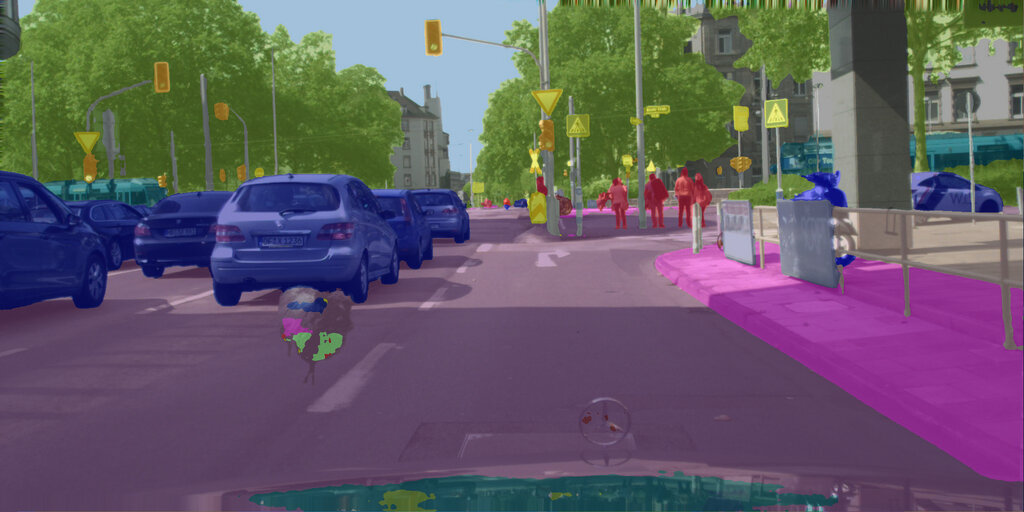}}~
    \subfloat[Our entropy heatmap]{\includegraphics[width=0.23\textwidth]{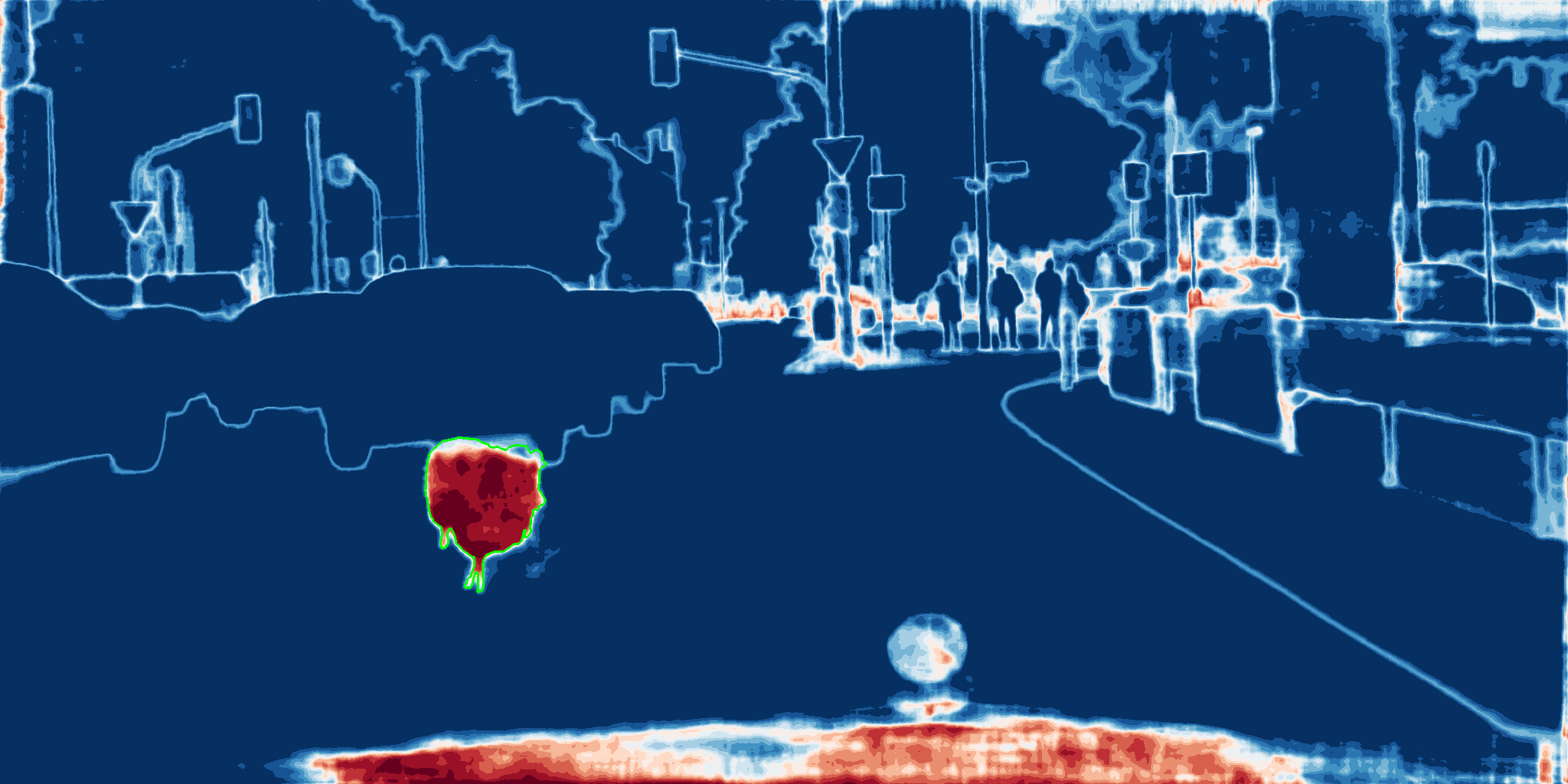}}
    \caption{Comparison of segmentation mask and softmax entropy before our OoD training (\emph{top row}) and after (\emph{bottom row}). While there are minor differences in the segmentation masks, the annotated unknown object (marked with green contours) becomes clearly recognizable in the entropy heatmap due to our OoD training. In the heatmap high values are red. 
    }
    \label{fig:page1}
\end{figure}

In recent years spectacular advances in the computer vision task semantic segmentation have been achieved by deep learning \cite{Wang19HRnet, Zhu19}. Deep convolutional neural networks (CNNs) are envisioned to be deployed to real world applications, where they are likely to be exposed to data that is substantially different from the model's training data. We consider data samples that are not included in the set of a model's semantic space as \textit{out-of-distribution} (OoD) samples. State-of-the-art neural networks for semantic segmentation, however, are trained to recognize a predefined closed set of object classes \cite{Cordts2016Cityscapes, Lin14COCO}, \eg for the usage in environment perception systems of autonomous vehicles \cite{janai20}. In open world settings there are countless possibly occurring objects. Defining additional classes requires a large amount of annotated data (cf. \cite{colling2021metabox, Zlateski18}) and may even lead to performance drops \cite{Deng10}. One natural approach is to introduce a \emph{none-of-the-known} output for objects not belonging to any of the predefined classes \cite{zhang17universum}. In other words, one uses a set of object classes that is sufficient for most scenarios and treats OoD objects by enforcing an alternative model output for such samples. From a functional safety point of view, it is a crucial but missing prerequisite that neural networks are capable of reliably indicating when they are operating out of their proper domain, \ie, detecting OoD objects, in order to initiate a fallback policy. 

As images from everyday scenes usually contain many different objects, of which only some could be out-of-distribution, knowing the location where the OoD object occurs is desired for practical application. Therefore, we address the problem of detecting anomalous regions in an image, which is the case if an OoD object is present (see \cref{fig:page1}) and which is a research area of high interest \cite{Blum19fishyscapes, hendrycks2019anomalyseg, Lis19, Pinggera16LostAndFound}. This so-called \emph{anomaly segmentation} \cite{baur2018deep, hendrycks2019anomalyseg} can be pursued, for instance, by incorporating sophisticated uncertainty estimates \cite{Atanov2019, gal2016} or by adding an extra class to the model's learnable set of classes \cite{zhang17universum}.

In this work, we detect OoD objects in semantic segmentation with a different approach which is composed of two steps:
As first step, we re-train the segmentation CNN to predict class labels with low confidence scores on OoD inputs, by enforcing the model to output high prediction uncertainty. In order to quantify uncertainty, we compute the softmax entropy which is maximized when a model outputs uniform probability scores over all classes \cite{lee18}. By deliberately including annotated OoD objects as \textit{known unknowns} into the re-training process and employing a modified multi-objective loss function, we observe that the segmentation CNN generalizes learned uncertainty to unseen OoD samples (\textit{unknown unknowns}) without significantly sacrificing in original
performance on the primary task, see \cref{fig:page1}.

The initial model for semantic segmentation is trained on the Cityscapes data \cite{Cordts2016Cityscapes}. As proxy for OoD samples we randomly pick images from the COCO dataset \cite{Lin14COCO} excluding the ones with instances that are also available in Cityscapes, cf. \cite{Hein19relu, hendrycks19oe, Meinke20} for a related approach in image classification. We evaluate the pixel-wise OoD detection performance via entropy thresholding for OoD samples from the LostAndFound \cite{Pinggera16LostAndFound} and Fishyscapes \cite{Blum19fishyscapes} dataset, respectively. Both datasets share the same setup as Cityscapes but include OoD objects.

The second step incorporates a \textit{meta classifier} flagging incorrect class predictions at segment level, similar as proposed in \cite{Maag20, Rottmann18, Rottmann19} for the detection of false positive instances in semantic segmentation. After increasing the sensitivity towards predicting OoD objects, we aim at removing false predictions which are produced due to the preceding entropy boost (cf. \cite{Chan20}). The removal of false positive OoD object predictions is based on aggregated dispersion measures and geometry features within segments (connected components of pixels), with all information derived solely from the CNN's softmax output. As meta classifier we employ a simple linear model which allows us to track and understand the impact of each metric.

To sum up our contributions, we are the first to successfully modify the training of segmentation CNNs to make them much more efficient at detecting OoD samples in LostAndFound and Fishyscapes. Re-training the CNNs with a specific choice of OoD images from COCO \cite{Lin14COCO} clearly outperforms the natural baseline approach of plain softmax entropy thresholding \cite{Hendrycks2017} as well as many state-of-the-art approaches from image classification. In addition, we are the first to demonstrate that entropy based OoD object predictions in semantic segmentation can be meta classified reliably, \ie, classified whether one considered OoD prediction is true positive or false positive without having access to the ground truth. For this meta task we employ simple logistic regression. Combining entropy maximization and meta classification therefore is an efficient and yet lightweight method, which is particularly suitable as an integrated monitoring system of safety-critical real world applications based on deep learning.

\section{Related Work}
\label{sec:RelatedWork}

Methods from prior works have already proven their efficiency in identifying OoD inputs for image data. The proposed methods are either modifications of the training procedure \cite{Hein19relu, hendrycks19oe, lee18, liang18, Meinke20} or post-processing techniques adjusting the estimated confidence \cite{DeVries2018, Hendrycks2017, lee18}. However, most of these works treat entire images as OoD.

When considering the semantic space to be fixed, one possible approach to anomaly segmentation, which we also pursue here, is to estimate uncertainty of CNNs.
Early approaches to uncertainty estimation involve Bayesian neural networks (BNNs) yielding posterior distributions over the model's weight parameters \cite{Mackay1992, neal2012}. In practice, approximations such as Monte-Carlo dropout \cite{gal2016} or stochastic batch normalization \cite{Atanov2019} are mainly used due to cheaper computational costs. Frameworks using dropout for uncertainty estimation applied to semantic segmentation have been developed in \cite{Badrinarayanan17BayesianSegNet, Kendall17}. Other approaches to model uncertainty consist of using an ensemble of neural networks \cite{Lakshminarayanan17}, which captures model uncertainty by averaging predictions over multiple models, and density estimation \cite{Blum19fishyscapes, Choi2018, nalisnick18, Ren19} via estimating the likelihood of samples with respect to the training distribution. Methods for OoD detection in semantic segmentation based on classification uncertainty and processing only monocular images have been analyzed in \cite{Angus2019, Bruegge20, Isobe17, jourdan20, Mehrtash20, Oberdiek20}.

Using BNNs for estimating uncertainty in deep neural networks is associated with prohibitive computational costs. Uncertainty estimates that are generated by multiple models or by multiple forward passes are still computationally expensive compared to single inference based ones. In our approach, we unite semantic segmentation and OoD detection in one model without any modifications of the underlying CNN's architecture. Therefore, our re-training approach can be even combined with existing OoD detection techniques and potentially enhance their efficiency.

Works with similar training approaches as ours use a different OoD proxy and are presented in \cite{Blum19fishyscapes, jourdan20}. They train neural networks on the unlabeled objects in Cityscapes as OoD approximation.
However, in our experiments we observe that the unlabeled data in Cityscapes lacks in diversity and therefore tends to be too dataset specific. With respect to other OoD datasets, such as LostAndFound and Fishyscapes, on which we perform our experiments, we observe that these mentioned methods fail to generalize. Furthermore, in contrast to those works we incorporate a post-processing step that significantly improves the OoD detection performance.

Another line of work detects OoD samples in semantic segmentation by incorporating autoencoders \cite{akccay2019skip, baur2018deep, Creusot15, Lis19}. Training such a model only on specific samples from a closed set of classes, it is assumed that the autoencoder model performs less accurately when fed with samples from never-seen-before classes. The identification of an OoD input then relies on the reconstruction quality. In this way, no OoD data is required, except for further adjusting the sensitivity of the method.

Autoencoders are in fact deep neural networks themselves and usually do not include a segmentation model. For the goal of safe real-time semantic segmentation, \eg necessary for automated driving \cite{janai20}, more lightweight approaches are favorable. We avoid incorporating deep auxiliary models at all and only employ a lightweight linear model instead. Usually the more complex a model, the greater the lack of interpretability. As monitoring systems are supposed to make deep learning models safer, one seeks for simpler and thereby more explainable approaches. We post-process our entropy boosted semantic segmentation CNN output via logistic regression whose computational overhead is negligible. This linear model is transparent as it allows us to analyze the impact of each single feature fed into the model and it demonstrates in our experiments to efficiently reduce the number of OoD detection errors.

\section{Entropy based OoD Detection} \label{sec:entropy}

In this section, we present our training method to improve the detection of OoD pixels in semantic segmentation via spatial entropy heatmapping.

\subsection{Training for high Entropy on OoD Samples} \label{sec:ood_training}
Let $f(x)\in(0,1)^q$ denote the softmax probabilities after processing the input image $x\in\mathcal{X}$ with some deep learning model $f:\mathcal{X}\rightarrow(0,1)^q$ and let $q = |\mathcal{C}|\in\mathbb{N}$ denote the number of classes. For the sake of brevity, we omit the consideration of image pixels in this section. We compute the softmax entropy via
\begin{equation}
	E(f(x)) = -\sum_{j\in\mathcal{C}} f_j(x) \log(f_j(x)) \label{eq:entropy} ~.
\end{equation}
By $(x,y(x)) \sim \mathcal{D}_{in}$ we denote an ``in-distribution'' example with $y(x)\in\mathcal{C}$ being the ground truth class label of input $x$, and by $x^\prime \sim \mathcal{D}_{out}$ we denote an ``out-distribution'' example for which no ground truth label is given. We aim at minimizing the overall objective
\begin{equation} \label{eq:obj}
\begin{split}
	L := (1-\lambda) \ &\mathbb{E}_{(x,y) \sim \mathcal{D}_{in}} \left[ \ell_{in} (f(x),y(x)) \right] \\
	+~\lambda\ &\mathbb{E}_{x^\prime \sim \mathcal{D}_{out}} \left[ \ell_{out} (f(x^\prime)) \right] \ ,~ \lambda \in [0,1]
\end{split}
\end{equation}
where
\begin{align}
	\ell_{in}(f(x),y(x)) &:= -\sum_{j\in\mathcal{C}} \mathds{1}_{j=y(x)} \log(f_j(x)) ~~ \textrm{and} \label{eq:in-loss} \\
	\ell_{out}(f(x^\prime)) &:= -\sum_{j\in\mathcal{C}} \frac{1}{q} \log(f_j(x^\prime)) \label{eq:out-loss}
\end{align}
with the indicator function $\mathds{1}_{j=y(x)}\in \{ 0,1 \} $ being equal to one if $j=y(x)$ and zero else.
In other words, for in-distribution samples we apply the commonly used empirical cross entropy loss, \ie, the negative log-likelihood of the target class. For out-distribution samples, we consider the negative log-likelihood averaged over all classes.

By that choice of out-distribution loss function, minimizing $\ell_{out}(f(x^\prime))$ is equivalent to maximizing the softmax entropy $E(f(x))$, see \cref{eq:entropy}. Since the softmax definition implies $f_j(x) \in (0,1)$ and $\sum_{j\in\mathcal{C}} f_j(x) = 1$, Jensen's inequality yields $\ell_{out}(f(x)) \geq  \log(q)$ and $E(f(x)) \leq \log(q)$, with equality (for both inequalities) if $f_j(x)=1/q ~ \forall ~ j\in\mathcal{C}$, \ie, if the softmax probabilities are uniformly distributed over all classes.

In order to control the impact of each single objective on the overall objective $L$, the convex combination between expected in-distribution loss and expected out-distribution loss is included, which can be adjusted by varying the parameter $\lambda$, see \cref{eq:obj}.

\subsection{OoD Object Prediction in Semantic Segmentation via Entropy Thresholding} \label{sec:ood_seg}

\begin{figure}[t]
    \captionsetup[subfigure]{labelformat=empty}
    \centering
    
    \sbox{\measurebox}{%
    \begin{minipage}[b]{0.1\linewidth}
    {\includegraphics[width=.9\linewidth]{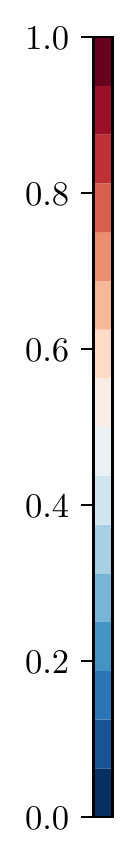}}
    \end{minipage}}
    \usebox{\measurebox}
    \begin{minipage}[b][\ht\measurebox][s]{.44\linewidth}
    \centering
    \subfloat[Entropy w/o OoD training]{\includegraphics[width=.99\linewidth]{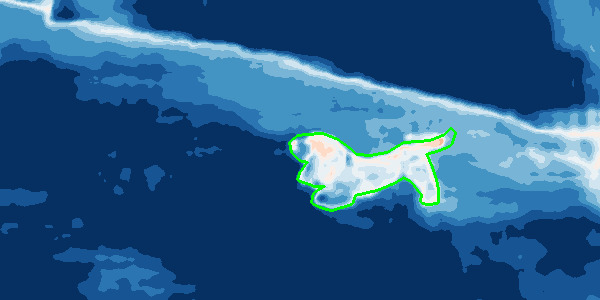}}
    \vfill
    \subfloat[Entropy w/ OoD training]{\includegraphics[width=.99\linewidth]{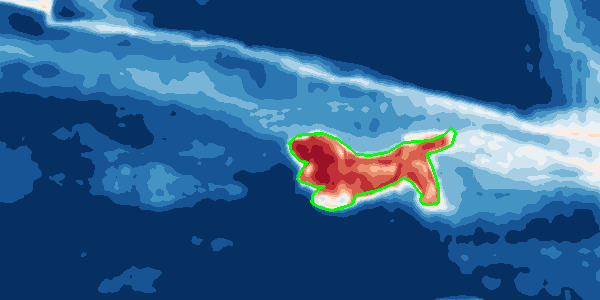}}
    \end{minipage}
    \begin{minipage}[b][\ht\measurebox][s]{.44\linewidth}
    \centering
    \subfloat[Prediction w/o OoD training]{\includegraphics[width=.99\linewidth]{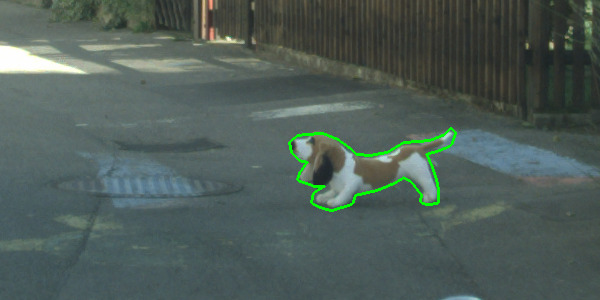}}
    \vfill
    \subfloat[Prediction w/ OoD training]{\includegraphics[width=.99\linewidth]{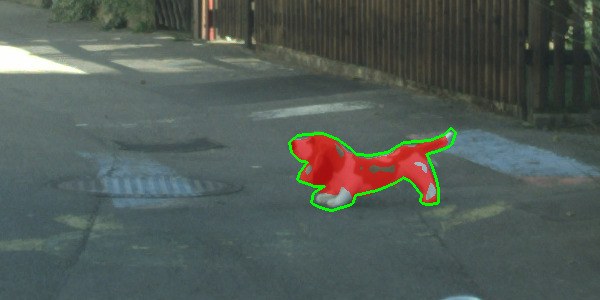}}
    \end{minipage}
    
    \caption{Comparison of softmax entropy heatmap and OoD prediction mask with our OoD training (\emph{bottom row}) and without (\emph{top row}). The green contours in the entropy heatmaps mark the annotation of the OoD object. The OoD object prediction is obtained by simply thresholding on the entropy heatmap (in this example at $t=0.7$ yielding the red pixels in the OoD prediction masks).}
    \label{fig:ood_pred}
\end{figure}

The softmax probabilities output of CNNs for semantic segmentation $f(x) \in (0,1)^{|\mathcal{Z}| \times q}, x\in\mathcal{X}~\subseteq [0,1]^{|\mathcal{Z}| \times 3}$ can be viewed as pixel-wise probability distributions that express how likely each potential class affiliation $j=1,\ldots,q$ at a given pixel $z \in \mathcal{Z}$ is, according to the model $f$. 
Let $f^{z}(x) \in (0,1)^q$ denote the softmax output at pixel location $z$ which we implicitly considered throughout the previous section. In semantic segmentation one minimizes the averaged pixel-wise classification loss over the image, cf. \cref{eq:obj}.
For the sake of simplicity, we consider the normalized entropy $\bar E(f^z(x))$ at pixel location $z$ in the following, that is $E(f^z(x))$ divided by $\log(q)$. One pixel is then assumed to be out-of-distribution (OoD) if the normalized entropy $\bar E(f^z(x))$ at that pixel location $z$ is greater than a threshold $t \in [0,1]$, \ie, $z$ \emph{is predicted to be OoD} if
\begin{equation} \label{eq:pred_ood_pixel}
    z \in \hat{\mathcal{Z}}_{out}(x) := \{ z' \in \mathcal{Z} : \bar E(f^{z'}(x)) \geq t \} ~ .
\end{equation}
A connected component $k \in \hat{\mathcal{K}}(x) \subseteq \mathcal{P}( \hat{\mathcal{Z}}_{out}(x) )$ (the latter being the power set of $\hat{\mathcal{Z}}_{out}(x)$) consisting of neighboring pixels fulfilling the condition in \cref{eq:pred_ood_pixel} gives us an \emph{OoD segment / object prediction}. An illustration can be viewed in \cref{fig:ood_pred}. Obviously, the better an in-distribution pixel can be separated from an out-distribution pixel by means of the entropy, the more accurate the OoD object prediction will be. 

\section{Meta Classifier in Semantic Segmentation} \label{sec:meta-classifier}

By training the segmentation CNN to output uniform confidence scores as presented in \cref{sec:entropy}, we increase the sensitivity towards predicting OoD objects, aiming for an ``entropy boost'' on OoD samples. However, it is not guaranteed that only OoD samples have a high entropy. Therefore, detecting OoD samples via entropy boosting potentially comes along with a considerable number of false OoD predictions, resulting in an unfavorable trade-off.

In this context, we consider one entire OoD object prediction (see \cref{sec:ood_seg}) as true positive if its intersection over union ($IoU$, \cite{Everingham15}) with a ground truth OoD object is greater than zero. More formally, let $\mathcal{Z}_{out}(x)$ be the set of pixel locations in $x$ which are labeled OoD according to ground truth. Then $k \in \hat{\mathcal{K}}(x)$ \emph{is true positive} (TP) if
\begin{align}\label{eq:seg_tp}
\begin{split}
    & \text{IoU}(k,\mathcal{Z}_{out}(x)) > 0 \\ \Leftrightarrow ~ & \exists~ z \in k: \bar E(f^z(x)) \geq t \land z \in \mathcal{Z}_{out}(x) ~.
\end{split}
\end{align}
One could also set a higher threshold on the IoU score, however in this work we treat every single pixel as a potential road hazard as this results in the least possible amount of overlooked OoD objects.

In \cite{Chan20} it has been demonstrated that false-positives due to increased prediction sensitivity can be removed based on a meta classifier's decision, achieving improved trade-offs between error rates. This meta classifier is essentially a binary classification model added on top of a segmentation CNN \cite{Maag20, Rottmann18, Rottmann19}. We construct hand-crafted metrics per connected component of pixels by aggregating different pixel-wise uncertainty measures derived from the softmax probabilities, one of which is the entropy. The entropy metric has proven to be highly correlated to the segment-wise IoU and therefore contributes greatly to the meta classifier's performance, cf. \cite{Rottmann18}. Therefore, we expect the learned entropy maximization on OoD objects to improve the meta classification performance. In contrast to existing approaches, that consider neighboring pixels sharing the same class label as segment, we generate metrics for segments above the given entropy threshold $t$ to adapt meta classification to OoD detection. Moreover, we additionally consider the variances within segments when aggregating pixel-wise measures instead of the means only.

Given the softmax output, further pixel-wise measures we integrate into the meta classifier are the variation ratio $V(f(x)) = 1 - f_{\hat{c}}(x), \hat{c} =\arg\max_{j\in\mathcal{C}} f_j(x)$ and probability margin $M(f(x)) = V(f(x)) + \max_{j\in\mathcal{C}\setminus\{\hat{c}\} }f_j(x)$.
Moreover, we also consider geometry features, such as the segment's size or its ratio between interior and boundary \cite{Rottmann18}. These metrics serve as inputs for the \emph{meta} model that classifies into \emph{true positive} and \emph{false positive} (FP) OoD object prediction, \ie, classifying $k \in \hat{\mathcal{K}}(x)$ into the sets 
\begin{align}\label{eq:meta_classes}
    \begin{split}
    & C_\mathrm{TP} := \{ k^\prime \in \hat{\mathcal{K}}(x) : \text{IoU}(k^\prime,\mathcal{Z}_{out}(x)) > 0 \} ~~ \textrm{and} \\ & C_\mathrm{FP} := \{ k^\prime \in \hat{\mathcal{K}}(x) : \text{IoU}(k^\prime,\mathcal{Z}_{out}(x)) = 0 \} ~.
    \end{split}
\end{align}
The outlined hand-crafted metrics form a structured dataset of features where the rows correspond to predicted segments and the columns to metrics.

\section{Setup of Experiments} \label{sec:setup}

\begin{figure}[t]
    \centering
    \includegraphics[width=0.98\linewidth]{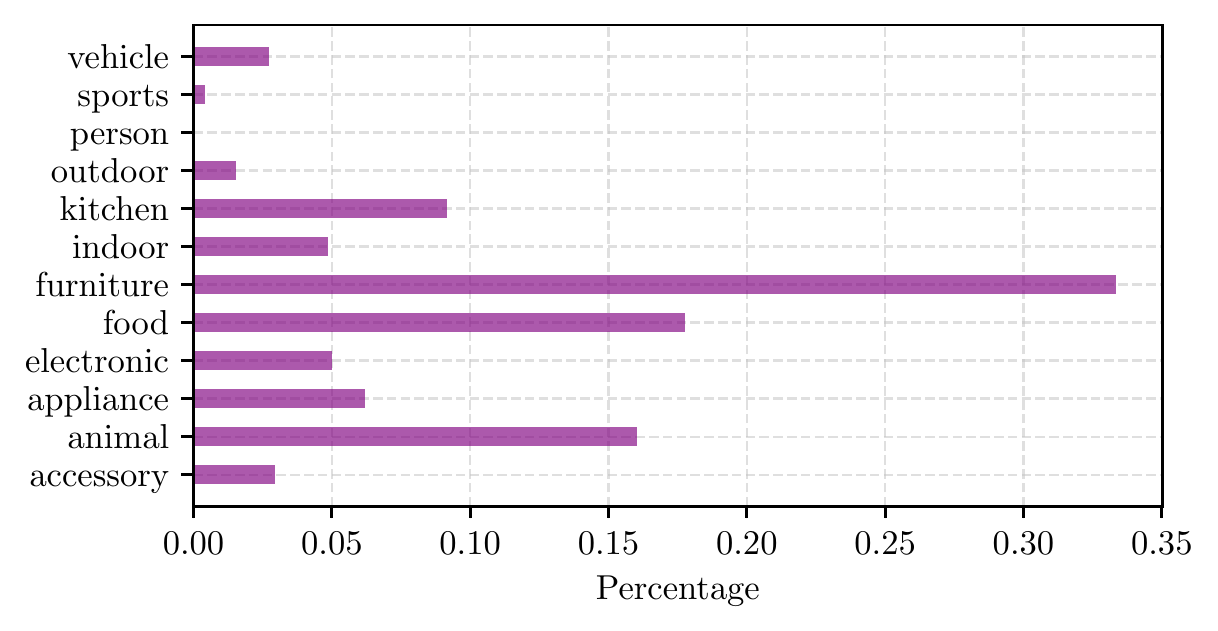}
    \caption{Relative number of pixels per supercategory in the COCO OoD proxy. In every epoch during OoD training 297 out of 46,751 images in total are randomly included.}
    \label{fig:coco-dist}
\end{figure}

We consider the semantic segmentation of the Cityscapes data \cite{Cordts2016Cityscapes} as original task, \ie, we consider Cityscapes as in-distribution $\mathcal{D}_{in}$. The training split consists of 2,975 pixel-annotated urban street scene images. As original model, we use the state-of-the-art semantic segmentation DeepLabv3+ model with a WideResNet38 backbone trained by Nvidia \cite{Zhu19}. This model is initialized with publicly available weights and serves as our \emph{baseline} model. For testing, we evaluate the OoD detection performance on two datasets comprising street scene images and unexpected objects. We consider images from the LostAndFound test split \cite{Pinggera16LostAndFound}, containing 1,203 images with annotations of road and small obstacles in front of the (ego-)car, and Fishyscapes Validation \cite{Blum19fishyscapes}, containing 30 images with annotated anomalous objects extracted from Pascal VOC \cite{Everingham15} which are then overlayed in Cityscapes images. Both datasets share the same setup as Cityscapes but include some unknown road objects.

In order to perform the \emph{OoD training} as proposed in \cref{sec:ood_training}, we approximate the out-distribution via images from the COCO \cite{Lin14COCO} dataset. This dataset contains images of objects captured in everyday scenes. Besides, we only consider COCO images with instances that are not included in Cityscapes (no persons, no cars, no traffic lights, \etc) and images that have a minimum height and width of at least 480 pixels. After filtering, there remain 46,751 images serving as our proxy for $\mathcal{D}_{out}$. The pixel frequencies per class is visualized in \cref{fig:coco-dist}. We emphasize that none of the OoD objects in the test data have been seen during our OoD training since we use disjoint datasets for training and testing, that are originally also designed for completely different applications. 
The used OoD proxy is a mixture of true unknown unknowns (pylon, bloated plastic bag, styrofoam, \etc) as well as known unknowns in terms of visual similarities (\eg dogs are available in the test data and share some visual features of cats which are available in the OoD proxy).
Employing this COCO subset as approximation of $\mathcal{D}_{out}$ is motivated by works on OoD detection \cite{hendrycks19oe, Meinke20} where 80 million tiny images \cite{Torralba2008tiny} serve as proxy for all possible images.

We finetune the DeepLabv3+ model with loss functions according to \cref{eq:in-loss} and \cref{eq:out-loss}. As training data we randomly sample 297 images from our COCO subset per epoch and mix them into all 2,975 Cityscapes training images (1:10 ratio of out-distribution to in-distribution images). We train the model's weight parameters on random squared crops of height / width of 480 pixels for 4 epochs in total and set the (out-distribution) loss weight $\lambda=0.9$ (see \cref{eq:obj}). As optimizer we use Adam \cite{Kingma14adam} with a learning rate of $10^{-5}$.

\begin{figure}[t!]
    \centering
    \begin{minipage}{0.49\linewidth}
    \includegraphics[width=\textwidth]{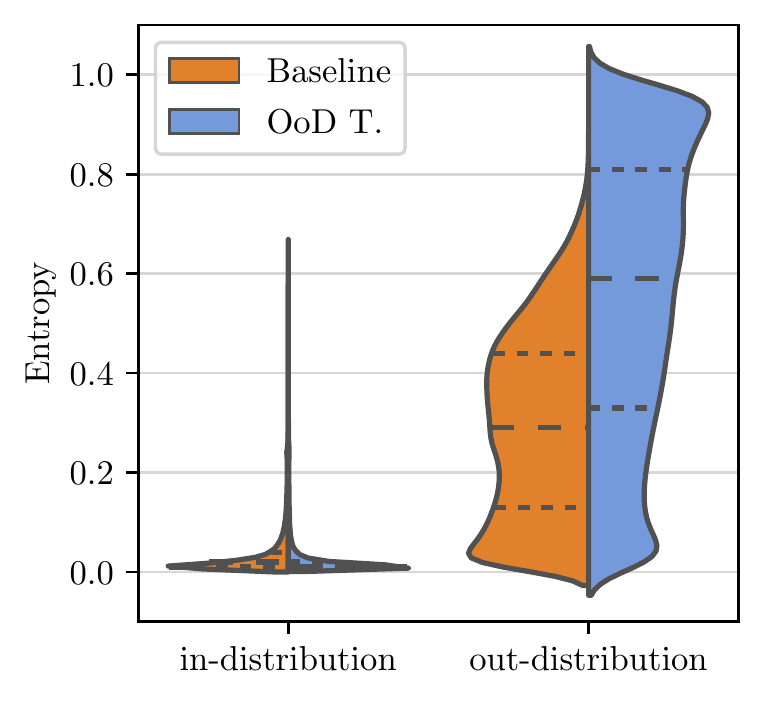}
    \centerline{\scriptsize ~~~~~ (a) LostAndFound}
    \end{minipage}
    \begin{minipage}{0.49\linewidth}
    \includegraphics[width=\textwidth]{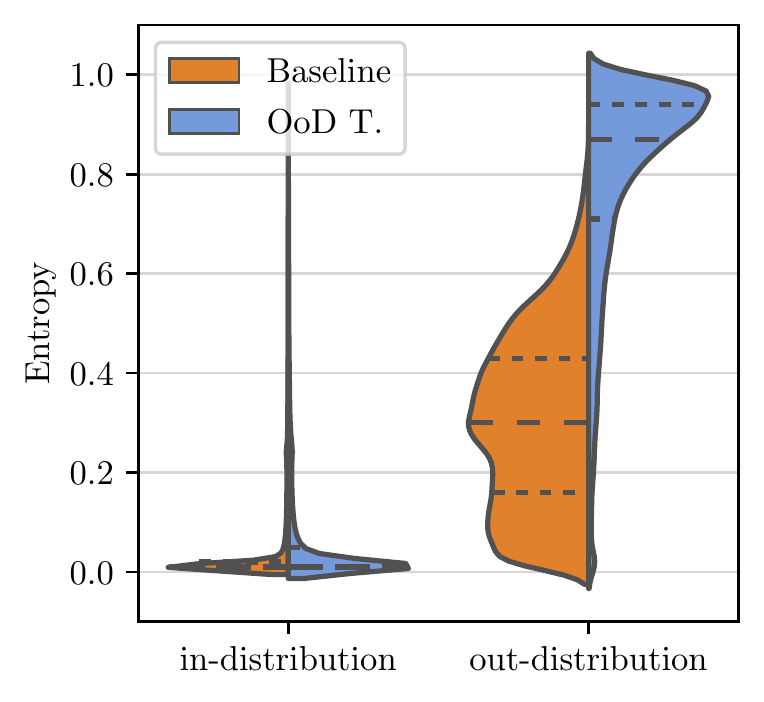}
    \centerline{\scriptsize ~~~~~ (b) Fishyscapes}
    \end{minipage}
    \caption{Relative pixel frequencies of (a) LostAndFound and (b) Fishyscapes OoD pixels, respectively. The density at different entropy values is displayed for the baseline model, \ie, before OoD training, and after OoD training. The inner lines of the violins represent the quartiles.}
    \label{fig:violins}
\end{figure}

\begin{figure*}[t]
    \centering
    ~~~~~~~\includegraphics[width=0.9\textwidth]{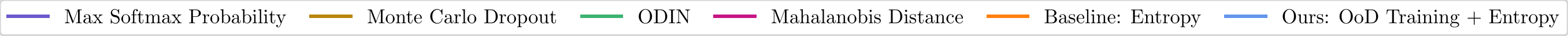}\\
    \vspace{-.3cm}
    \subfloat[LostAndFound (\emph{left:} AUROC, \emph{right:} AUPRC)]{\includegraphics[width=0.245\textwidth]{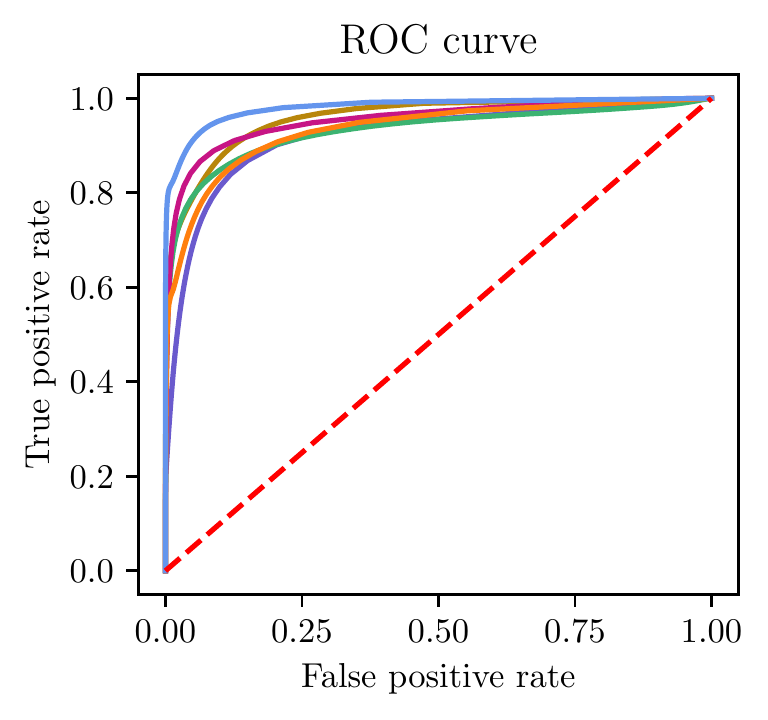} \includegraphics[width=0.245\textwidth]{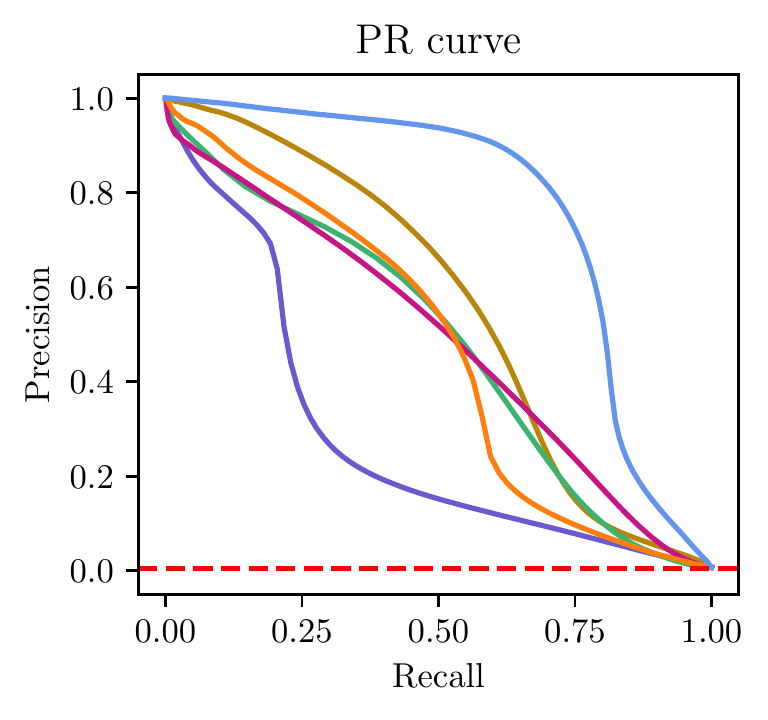}}~
    \subfloat[Fishyscapes (\emph{left:} AUROC, \emph{right:} AUPRC)]{\includegraphics[width=0.245\textwidth]{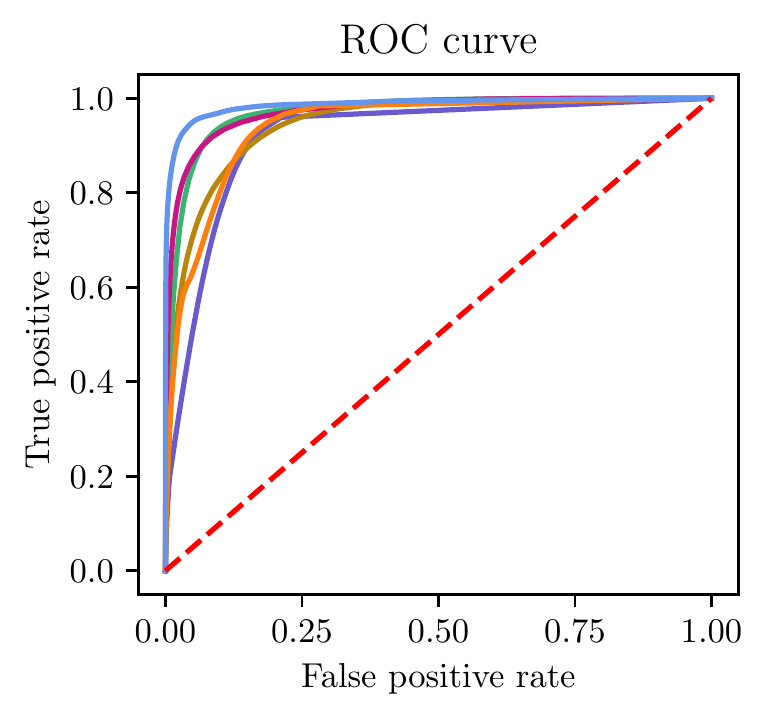} \includegraphics[width=0.245\textwidth]{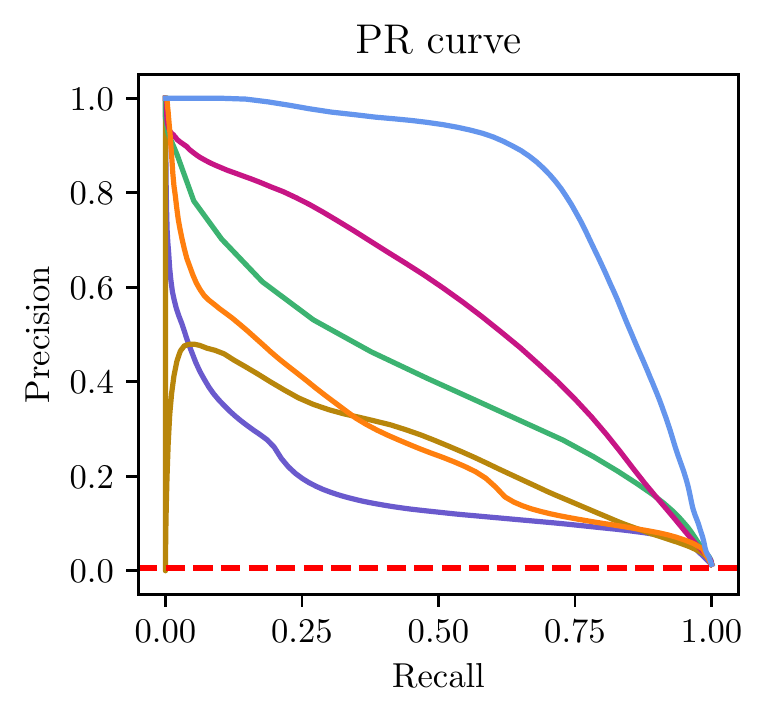}}
    \caption{Detection ability of LostAndFound (a) and Fishyscapes (b) OoD pixels, respectively, evaluated by means of receiver operating characteristic curve (\emph{a \& b left}) and precision recall curve (\emph{a \& b right}). The red lines indicate the performance according to random guessing, \ie, in the PR curves the red line indicate the fraction of OoD pixels.}
    \label{fig:auc}
\end{figure*}

\section{Pixel-wise Evaluation} \label{sec:results_pixel}
Based on the softmax probabilities, we compute the normalized entropy $\bar E$ for all pixels in the respective test dataset. This gives us per-pixel anomaly / OoD scores which we compare with the ground truth anomaly segmentation. For the sake of clarity, in this section we refer to in-distribution pixels as \emph{samples of the negative class} and to out-distribution pixels as \emph{samples of the positive class}.

\subsection{Separability by means of Area Under Curve}

On basis of the violin plots in \cref{fig:violins}, one already notices the beneficial effect of our OoD training over the baseline in separating in-distribution and out-distribution pixels as large masses of the distributions corresponding to the respective classes can be well separated for a larger range of entropy thresholds.
This effect can be further quantified with the aid of receiver operating characteristic (ROC) curves and precision recall (PR) curves. The area under the curve (AUC) then represents the degree of separability. The higher the AUC, the better the separability. In addition to the baseline, we include further scores of standard OoD detection methods. Namely these are: MSP \cite{Hendrycks2017}, MC dropout \cite{gal2016}, ODIN \cite{liang18} and Mahalanobis distance \cite{Lee2018mahala}.

By comparing the ROC curves for LostAndFound (\cref{fig:auc} (a) left), we observe that there is a performance gain over the baseline model when OoD training is applied. The baseline curve indicates that the corresponding model has a lower true positive rate across various fixed false positive rates, \ie, our model after OoD training assigns higher uncertainty / entropy values to OoD samples which is beneficial for OoD detection. Furthermore, also with respect to all other tested methods, entropy thresholding after OoD training shows the best degree of separability measured by the AUC of ROC curves (AUROC) with a score of $0.98$.
We observe the same effects for Fishyscapes (\cref{fig:auc} (b) left). From the Fishyscapes violins, the discrimination performance after OoD training seems already close to perfect. This is confirmed by the AUROC of $0.99$, again outperforming all other tested methods.

As the AUROC essentially measures the overlap of distributions corresponding to negative and positive samples, this score does not place more emphasis on one class over the other in case of class imbalance.
As there is a considerably strong class imbalance in LostAndFound and Fishyscapes ($0.7\%$ and $1.3\%$ OoD pixels), respectively, we also consider the PR curves, see \cref{fig:auc} (a) \& (b) right. Thus, true negatives are ignored and the emphasis shifts to the detection of the positive class (OoD samples). Now the AUC of PR curves (AUPRC) serves as measure of separability.
For LostAndFound as well as for Fishyscapes OoD pixels, the model after OoD training is superior not only over the baseline model but also any other tested method in terms of precision when we fix recall to any score. The AUPRC quantifies this performance gain and further clarifies the improved capability at detecting OoD pixels. Regarding LostAndFound, the OoD training increases the AUPRC over the baseline by $0.30$ up to a score of $0.76$. Regarding Fishyscapes, the performance gain is even more significant. We raise the AUC from $0.28$ up to $0.81$. We conclude that, measured by AUROC and AUPRC, our OoD training is highly beneficial for detecting OoD samples. 

Moreover, we conducted the same experiments as for the DeepLabv3+ model \cite{Zhu19} also for the weaker DualGCNNet \cite{Li19} which is re-trained with $\lambda=0.25$ for 11 epochs in total. We report all benchmark scores of all tested methods in \cref{tab:benchmark}. Besides AUPRC, we also provide the false positive rates at 95\% true positive rate ($\text{FPR}_{95}$) and the mean intersection over union (mIoU) for the semantic segmentation of the Cityscapes validation set. For further comparison, we additionally included scores of methods based on an auto-encoder \cite{Lis19} and on density estimation \cite{Blum19fishyscapes}.

\subsection{Original Task Performance} \label{sec:original_perf}

\begin{table}[t]
    \begin{center}
    \scalebox{0.67}{
    \begin{tabular}{l|cc|c}
    \toprule
    & $\text{FPR}_\text{95}$ $\downarrow$ & AUPRC $\uparrow$ & mIoU $\uparrow$  \\
    \midrule
    Network architecture and OoD score & \multicolumn{2}{c|}{LostAndFound Test} & \multicolumn{1}{c}{Cityscapes Val.} \\ 
    \midrule
    \midrule
    DualGCN \cite{Li19} + Entropy & 0.30 & 0.36 & 0.80 \\
    \rowcolor{Gray} Ours: DualGCN + OoD T. + Entropy  & 0.12 & 0.51 & 0.76 \\
    \midrule
    PSPNet \cite{Zhao2017PSP} + Image Resynthesis \cite{Lis19} & N/A & 0.41 & 0.80 \\
    DeepV3W + Max Softmax \cite{Hendrycks2017}  & 0.32 & 0.27 & \textbf{0.90} \\
    DeepV3W + ODIN \cite{liang18}  & 0.45  & 0.46 & \textbf{0.90} \\
    DeepV3W + MC Dropout \cite{gal2016} & 0.21 & 0.55 & 0.88 \\
    DeepV3W + Mahalanobis \cite{Lee2018mahala}  & 0.27 & 0.48 & \textbf{0.90} \\
    Baseline: DeepV3W \cite{Zhu19} + Entropy  & 0.35 & 0.46 & \textbf{0.90} \\
    \rowcolor{Gray} Ours: DeepV3W + OoD T. + Entropy  & \textbf{0.09} & \textbf{0.76} & 0.89 \\
    \midrule
    & \multicolumn{2}{c|}{Fishyscapes Val.} & \multicolumn{1}{c}{Cityscapes Val.} \\
    \midrule
    \midrule
    DualGCN \cite{Li19} + Entropy  & 0.46 & 0.07 & 0.80 \\
    \rowcolor{Gray} Ours: DualGCN + OoD T. + Entropy  & 0.21 & 0.38 & 0.76 \\
    \midrule
    DeepV3W + Max Softmax \cite{Hendrycks2017}  & 0.21 & 0.17 & \textbf{0.90} \\
    DeepV3W + ODIN \cite{liang18}  & 0.12  & 0.39 & \textbf{0.90} \\
    DeepV3W + MC Dropout \cite{gal2016} & 0.23 & 0.26 & 0.88 \\
    DeepV3W + Mahalanobis \cite{Lee2018mahala}  & 0.14 & 0.55 & \textbf{0.90} \\
    Baseline: DeepV3W \cite{Zhu19} + Entropy  & 0.18 & 0.28 & \textbf{0.90} \\
    \rowcolor{Gray} Ours: DeepV3W + OoD T. + Entropy  & \textbf{0.05} & \textbf{0.81} & 0.89 \\
    \midrule
    & \multicolumn{2}{c|}{Fishyscapes Static\footnotemark} & \multicolumn{1}{c}{Cityscapes Val.} \\
    \midrule
    DeepV3P \cite{Chen2018ECCV} + Image Resynthesis \cite{Lis19} & 0.27 & 0.30 & 0.80 \\
    DeepV3S \cite{Zhu19} + Learned Density \cite{Blum19fishyscapes} & 0.17 & 0.62 & 0.81 \\
    \rowcolor{Gray} Ours: DeepV3W + OoD T. + Entropy  & \textbf{0.09} & \textbf{0.87} & 0.89 \\
    \bottomrule
    \end{tabular}
    }
    \end{center}
    \caption{Results for LostAndFound and Fishyscapes.}\label{tab:benchmark}
\end{table}

\footnotetext{2nd best in public benchmark results: \href{https://fishyscapes.com/results}{https://fishyscapes.com/results}}

\begin{figure}[t]
    \centering
    \includegraphics[width=0.97\linewidth]{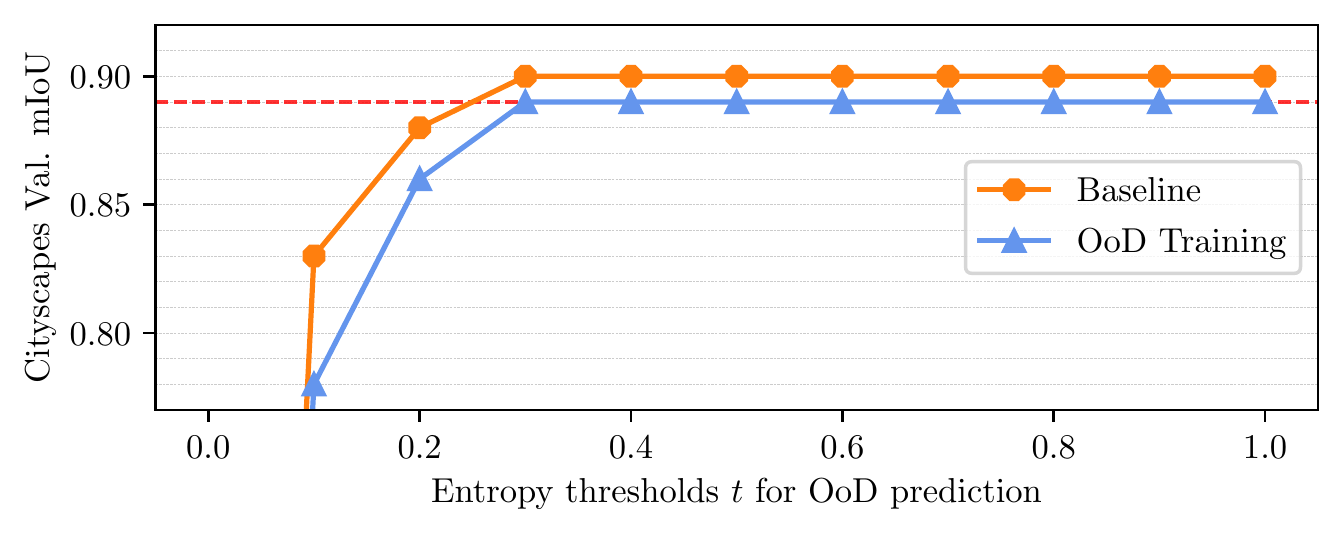}
    \caption{Mean intersection over union (mIoU) for the Cityscapes validation split with OoD predictions at entropy thresholds $t$. The dashed red line indicates the performance loss considered to be ``acceptable'' (1 percent point).}
    \label{fig:miou+ood}
\end{figure}

In order to monitor that the baseline model does not unlearn its original task due to OoD training, we evaluate the model's performance on in-distribution data with OoD predictions at different entropy thresholds. The original task is the semantic segmentation of the Cityscapes images and we evaluate by means of the most commonly used performance metric \emph{mean Intersection over Union} (mIoU, \cite{Everingham15}). Additionally to the Cityscapes class predictions, that is obtained via the standard maximum a posteriori (MAP) decision principle \cite{Chan18, Mehryar12mlfoundation}, we consider an extra OoD class prediction if the softmax entropy is above the given threshold $t$. We compute the mIoU for the Cityscapes validation dataset, but average only over the 19 Cityscapes class IoUs.

The state-of-the-art DeepLabv3+ model \cite{Zhu19}, which serves as our baseline throughout our experiments, achieves an mIoU score of $0.90$ on the Cityscapes validation dataset without OoD predictions (implying $t=1.0$). By re-training the CNN with entropy maximization on OoD inputs, we observe improved OoD-AUPRC scores.
This gain at detecting OoD samples comes with a marginal drop in Cityscapes validation mIoU down to $0.89$. These two mIoU scores remain nearly constant (deviations less than $1$ percent point) for the thresholds $t=0.3, \ldots, 1.0$. In general, the lower the entropy threshold, the more pixels are predicted to be OoD. For $t=0.2$ this results in a noticeable performance decrease, $0.05$ for the baseline model and $0.03$ for the re-trained model, respectively. As displayed in \cref{fig:miou+ood} further lowering the threshold leads to an even more significant sacrifice of original performance. Consequently, we consider in the following entropy thresholds of at least $t=0.3$ since the performance loss seems acceptable, especially in view of a substantially improved OoD detection capability.

\section{Segment-wise Evaluation} \label{sec: results_segment}

\begin{figure}[t]
    \captionsetup[subfigure]{labelformat=empty}
    \centering
    \begin{minipage}{0.49\linewidth}
    \includegraphics[width=\textwidth]{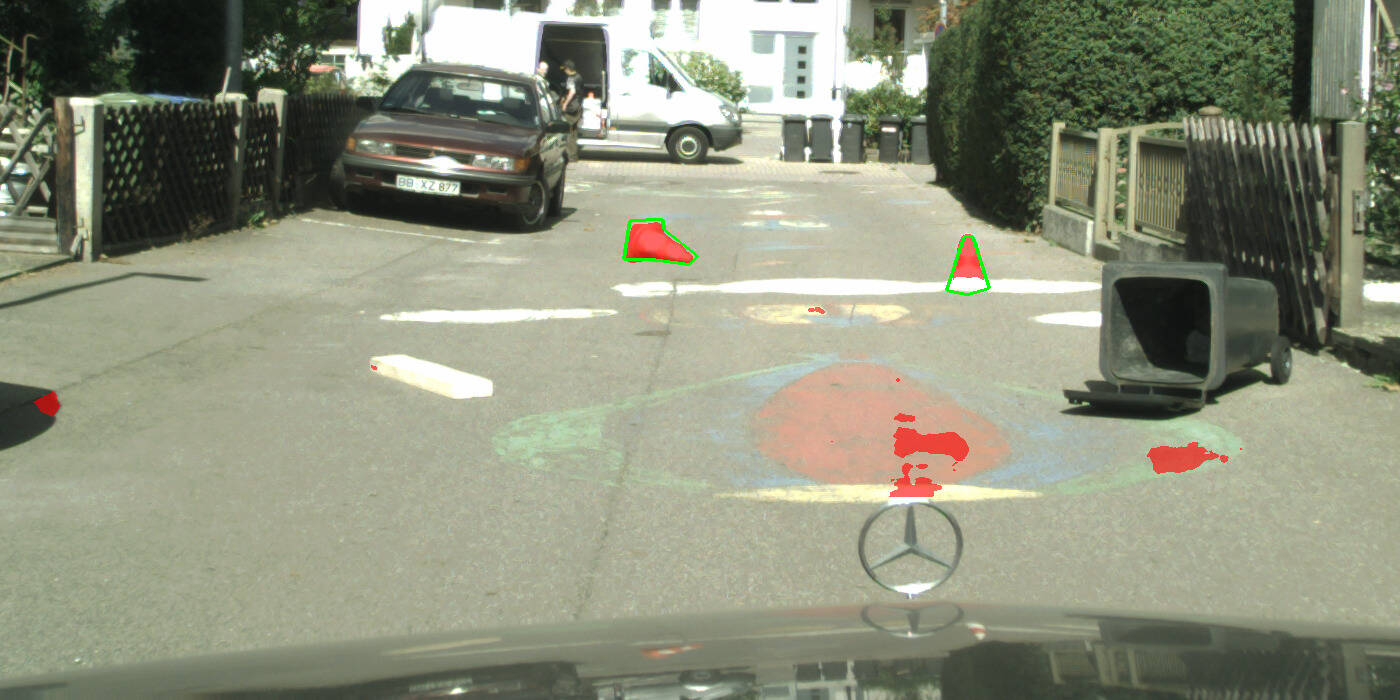}
    \centerline{\scriptsize OoD Training only}
    \end{minipage}
    \begin{minipage}{0.49\linewidth}
    \includegraphics[width=\textwidth]{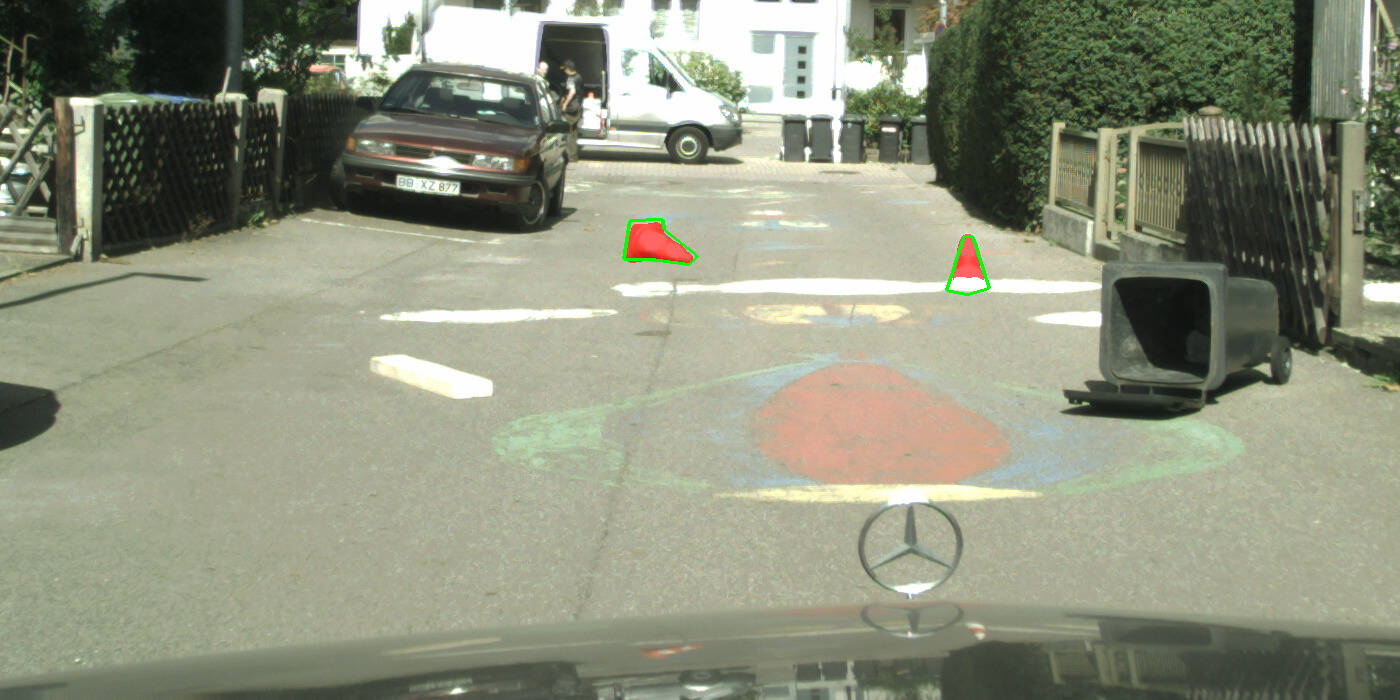}
    \centerline{\scriptsize OoD Training + meta classifier}
    \end{minipage}
    \caption{OoD detection with $t=0.5$ after OoD training and meta classification. The green contours mark the annotations of OoD objects. OoD predictions in the background according to the ground truth are ignored (this includes \eg the garbage bin even though it has been detected).}
    \label{fig:ood_metaclassif}
\end{figure}

In this section we evaluate the meta classification performance on LostAndFound. The main metrics for the segment-wise evaluation are the numbers of FPs and FNs with respect to an OoD object prediction, cf. \cref{eq:seg_tp}. The $F_1$-score $F_1 = 2\mathrm{TP}/(2\mathrm{TP}+\mathrm{FP}+\mathrm{FN}) \in [0,1]$ summarizes the error rates into an overall score. As the removal of FP OoD predictions should not come at cost of a significant loss in original performance, see \cref{fig:ood_metaclassif}, we additionally consider the \emph{miss rate of road pixels}:
\begin{equation} \label{eq:road_miss}
    \varepsilon := 1 - \left\vert \bigcup_{x\in\mathcal{X}} \left( \hat{\mathcal{Z}}_{in}(x) \cap \mathcal{Z}_{in}(x)\right) \right\vert \left\vert \bigcup_{x\in\mathcal{X}} \mathcal{Z}_{in}(x) \right\vert^{-1}
\end{equation}
with pixel locations predicted to be in-distribution in $\hat{\mathcal{Z}}_{in}$ and annotated as in-distribution in $\mathcal{Z}_{in}$. The road miss rate $\epsilon$ measures the fraction of actual road pixels in the whole dataset which are incorrectly identified.

\begin{figure}[t]
    \centering
    \includegraphics[width=0.99\linewidth]{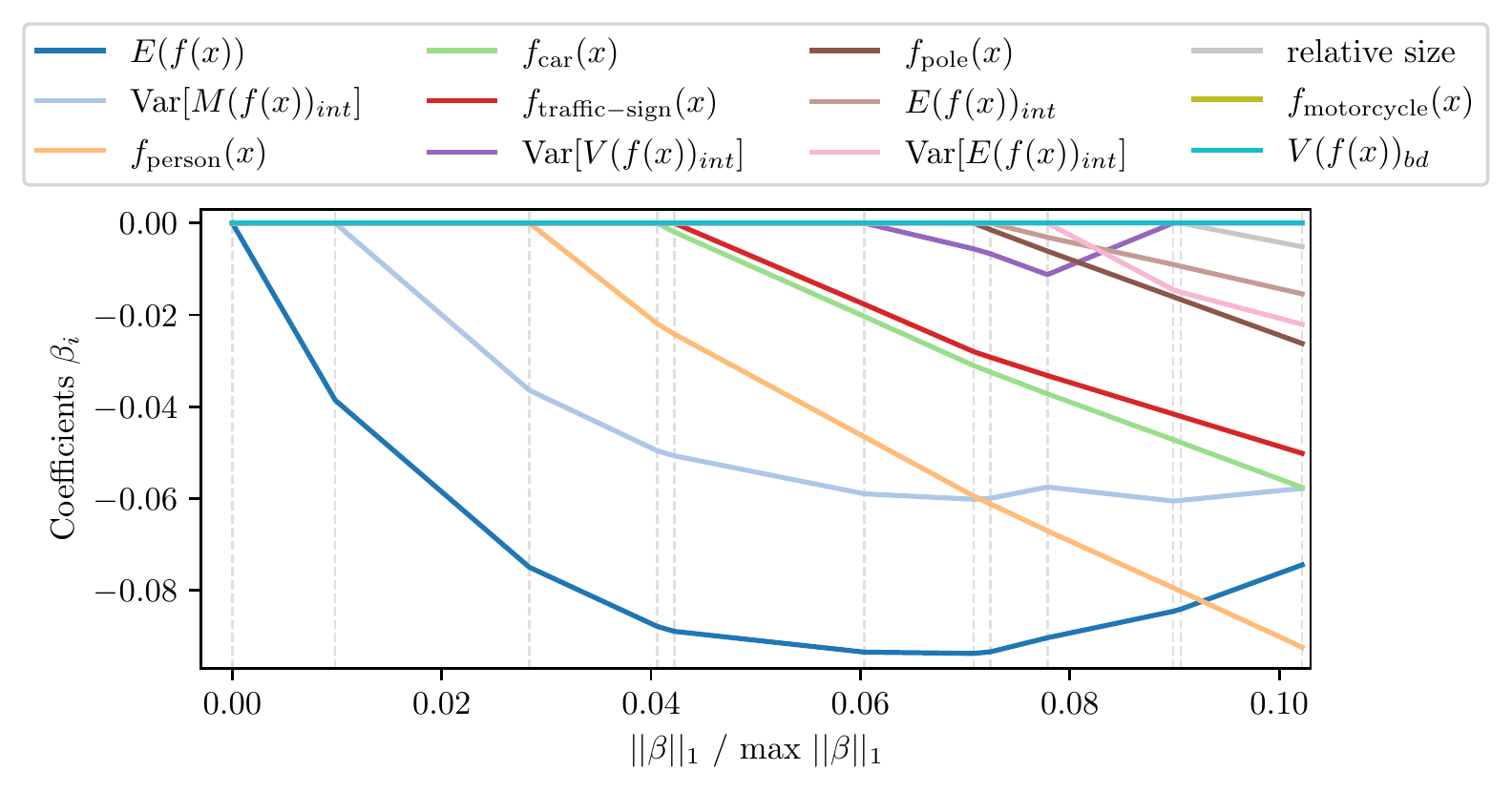}
    \caption{Least angle regression for the meta classifier with OoD training and an entropy threshold of $t=0.3$. The 12 features becoming active first are displayed. The suffixes $int$ and $bd$ refer to the restriction of a metric on the segment's interior and boundary, respectively. See \cref{sec:meta-classifier} for a description of the metrics.}
    \label{fig:lar}
\end{figure}

We compute per-segment metrics as outlined in \cref{sec:meta-classifier} for OoD object predictions in the LostAndFound test set and feed them through meta classification models, which are simple logistic regressions throughout our experiments. The segments are then leave-one-out cross validated whether they are TP or FP, see \cref{eq:meta_classes}. Via least angle regression we analyze the metrics having the most impact on the meta classification. The analysis shows that after OoD training the entropy metric $E(f(x))$ has the most impact, see \eg \cref{fig:lar} for $t=0.3$.

\begin{table*}[t]
    \begin{center}
    \scalebox{0.68}{
    \begin{tabular}{c||rr|r|r||rr|r|r||rr|r|r||rr|r|r}
    \toprule
    Entropy & \multicolumn{4}{c}{Baseline} & \multicolumn{4}{|c}{Baseline} & \multicolumn{4}{|c}{OoD Training} & \multicolumn{4}{|c}{OoD Training} \\
    Threshold & \multicolumn{4}{c}{} & \multicolumn{4}{|c}{+ Meta Classifier} & \multicolumn{4}{|c}{}  & \multicolumn{4}{|c}{+ Meta Classifier}  \\
    \midrule
    \midrule
    $\bar E \geq t$ & FP $\downarrow$  & FN $\downarrow$  & $F_1$ $\uparrow$ & $\varepsilon$ in \% $\downarrow$ & FP $\downarrow$ & FN $\downarrow$  & $F_1$ $\uparrow$ & $\varepsilon$ in \% $\downarrow$ & FP $\downarrow$ & FN $\downarrow$  & $F_1$ $\uparrow$ & $\varepsilon$ in \% $\downarrow$ & FP $\downarrow$ & FN $\downarrow$  & $F_1$ $\uparrow$ & $\varepsilon$ in \% $\downarrow$ \\
    \midrule
    $t=0.10$            & 33,584   & 77   &  0.09 &  7.60 & 386   & 314 & 0.80 & 3.24 & 21,967  & 99   & 0.12 &  5.22 & 245   & 302 & 0.83 & 2.70 \\
    $t=0.20$            & 19,456   & 136   &  0.13 &  2.48 & 454   & 307 & 0.78 & 0.93 & 17,000  & 127   & 0.15 &  2.14 & 271 & 303 & 0.83 & 0.18 \\
    \midrule
    $t=0.30$            & 7,349   & 218   &  0.28 &  0.38 & 412   & 302 & 0.79 & 0.09 & 8,068  & \textbf{191}   & 0.26 &  0.30 &  \textbf{290}   & 308 & \textbf{0.82} & \textbf{0.06} \\
    $t=0.40$            & 3,214   & 377   &  0.42 &  0.08 & 280   & 435 & 0.77 & 0.03 & 4,035  & \textbf{289}   & 0.39 & 0.11 & \textbf{251}   & 359 & \textbf{0.81} & \textbf{0.03} \\
    $t=0.50$            & 809   & 662   &   0.58 &  0.01 & \textbf{94}   & 686 & 0.71 & $<$ \textbf{0.01} & 1,215  & \textbf{415}   & 0.60  & 0.04 & 145   & 447 & \textbf{0.80} & 0.02 \\
    $t=0.60$            & 158   & 1,084  &   0.69 & $<$ 0.01 & \textbf{26}  & 1,093 & 0.50 & \textbf{$<$ 0.01} & 327   & \textbf{613}   & 0.69  & 0.02 & 49   & 619 & \textbf{0.76} & 0.02 \\
    $t=0.70$            & 10    & 1,511  &   0.16  & $<$ 0.01 & \textbf{3}  & 1,512 & 0.16 & $<$ \textbf{0.01} & 135   & \textbf{879}   & 0.61  & 0.01 & 21   & 881 & \textbf{0.63} & 0.01 \\
    \bottomrule
    \end{tabular}
    }
    \end{center}
    \caption{Detection errors for LostAndFound OoD objects at different entropy thresholds $t$. We consider the road miss rate $\varepsilon$, see \cref{eq:road_miss}, as further measure of loss in original performance (for Cityscapes mIoU, see \cref{fig:miou+ood}). Below the horizontal line, \ie, $t\geq0.3$, we consider the loss in original performance to be acceptable, see \cref{sec:original_perf} for further details.} \label{tab:res_segment}
\end{table*}

\begin{figure}[t]
    \centering
    \includegraphics[width=0.99\linewidth]{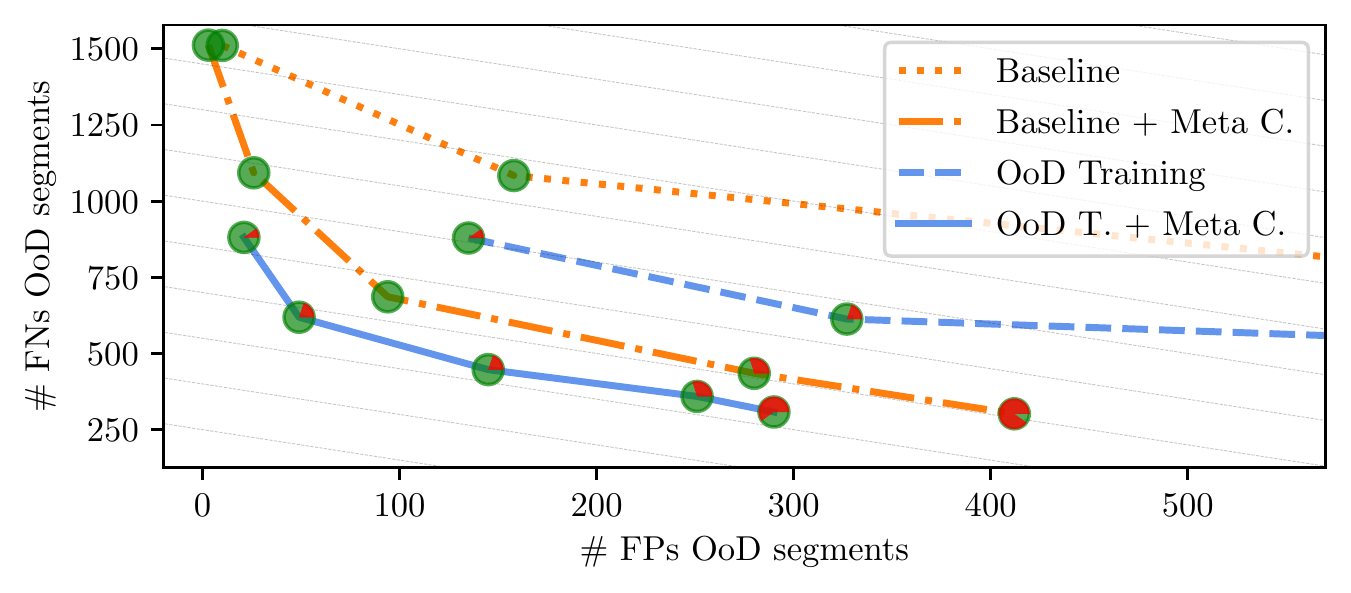}
    \caption{Detection errors of LostAndFound OoD objects. In this plot, the number of errors when $t=0.7, \ldots, 0.3$ are displayed (when in the axes' range). The pie-chart markers indicate the road miss rate $\varepsilon$, being entirely red if $\varepsilon\geq0.001$. See also \cref{tab:res_segment} for exact numbers.}
    \label{fig:error_rate}
\end{figure}

In general, the higher the entropy threshold, the less OoD objects are predicted and consequently less data is fed through the linear models. This explains the observation that meta classifiers identify FPs more reliably the lower $t$. Due to our OoD training, the meta classifiers demonstrate to be more effective, being most superior when $t=0.7$. In our experiments, OoD training in combination with meta classification at $t=0.3$ turns out to be the best OoD detection approach achieving the best result with only 598 errors in total and $F_1=0.82$ while having a road miss rate of marginally 0.06\%, see also \cref{fig:error_rate}. Compared to the best baseline at $t=0.6$ with $F_1=0.69$, we decrease the number of total errors by 52\% from 1,242 down to 598. More safety-relevantly, at the same time we significantly reduce the number of overlooked OoD objects by 70\% from 1,084 down to 308.

The numbers of detection errors, $F_1$ scores and road miss rates $\varepsilon$ at different entropy thresholds $t$ are summarized in \cref{tab:res_segment}. The FP OoD removal efficiency is given in \cref{tab:metaclass}.

\begin{table}[t]
    \begin{center}
    \scalebox{0.68}{
    \begin{tabular}{c||cc||cc||cc}
    \toprule
    Entropy & \multicolumn{2}{|c}{Baseline + MSP \cite{Hendrycks2017}} & \multicolumn{2}{|c}{Baseline + Meta C.} & \multicolumn{2}{|c}{OoD T. + Meta C.} \\ 
    Threshold $t$   & \multicolumn{1}{|c}{AUROC}  & \multicolumn{1}{c}{AUPRC} & \multicolumn{1}{|c}{AUROC}  & \multicolumn{1}{c}{AUPRC}  & \multicolumn{1}{|c}{AUROC}  & \multicolumn{1}{c}{AUPRC}  \\
    \midrule
    \midrule
    $t = 0.10$            & 0.8509 & 0.9817 & 0.9894 & 0.9993 & 0.9915 & 0.9993 \\
    $t = 0.20$            & 0.6470 & 0.9119 & 0.9859 & 0.9980 & 0.9898 & 0.9980 \\
    \midrule
    $t = 0.30$            & 0.5333 & 0.7376 & 0.9742 & 0.9884 & \textbf{0.9847} & \textbf{0.9953} \\
    $t = 0.40$            & 0.3847 & 0.4671 & 0.9715 & 0.9740 & \textbf{0.9808} & \textbf{0.9807} \\
    $t = 0.50$            & 0.4172 & 0.2286 & 0.9628 & 0.9214 & \textbf{0.9665} & \textbf{0.9536} \\
    $t = 0.60$            & 0.4906 & 0.1228 & 0.9291 & 0.7252 & \textbf{0.9511} & \textbf{0.8405} \\
    $t = 0.70$            & 0.5932 & 0.1334 & 0.9140 & 0.5283 & \textbf{0.9444} & \textbf{0.7185} \\
    \bottomrule
    \end{tabular}
    }
    \end{center}
    \caption{Meta classification performance on LostAndFound at different entropy thresholds $t$. As comparison to the meta classifier, we include the detection of OoD prediction errors via the maximum softmax probability (MSP, \cite{Hendrycks2017}).} \label{tab:metaclass}
\end{table}

\section{Conclusion \& Outlook}

In this work, we presented a novel re-training approach for deep neural networks that unites improved OoD detection capability and state-of-the-art semantic segmentation in one model. Up to now, only a small number of prior works exist for anomaly segmentation on LostAndFound and Fishyscapes, respectively.
We demonstrate that our OoD training significantly improves the detection efficiency via softmax entropy thresholding, leading to superior performance over existing OoD detection approaches.

Moreover, we introduced meta classifiers for entropy based OoD object predictions. By applying lightweight logistic regressions, we have demonstrated that entire LostAndFound OoD segments are meta classified reliably. This observation already holds for the tested CNN in its plain version. Due to the increased sensitivity of OoD predictions via entropy maximization, the meta classifiers' efficiency is even more pronounced. In view of emerging safety-critical deep learning applications, the combination of OoD training and meta classification has the potential to considerably improve the overall system's performance.

For future work, we plan to apply OoD training for the retrieval of OoD objects in order to assess the importance of their occurrence and whether a new concept is required to be learned. Our code is publicly available at \href{https://github.com/robin-chan/meta-ood}{https://github.com/robin-chan/meta-ood}.

\paragraph{Acknowledgement.}
The research leading to these results is funded by the German Federal Ministry for Economic Affairs and Energy within the project ``KI Absicherung – Safe AI for Automated Driving'', grant no.\ 19A19005R. The authors would like to thank the consortium for the successful cooperation. The authors gratefully also acknowledge the Gauss Centre for Supercomputing e.V.\ (\href{https://www.gauss-centre.eu}{https://www.gauss-centre.eu}) for funding this project by providing computing time through the John von Neumann Institute for Computing (NIC) on the GCS Supercomputer JUWELS at Jülich Supercomputing Centre (JSC).

{\small
\bibliographystyle{ieee_fullname}
\bibliography{egbib}
}

\section*{Appendix}
\begin{appendices}

\section{Maximized Entropy via the chosen Out-Distribution Loss Function}
In \cref{sec:ood_training} we state that by our choice of out-distribution loss function we maximize the softmax entropy. As a reminder, the softmax entropy is defined as
\begin{equation}
	E(f(x)) := -\sum_{j\in\mathcal{C}} f_j(x) \log(f_j(x))
\end{equation}
and the out-distribution loss function as
\begin{equation}
    \ell_{out}(f(x)) := -\sum_{j\in\mathcal{C}} \frac{1}{|\mathcal{C}|} \log(f_j(x)) ~,
\end{equation}
where $\mathcal{C}$ denotes the the set of trained classes and $f(x) \in (0,1)^{|\mathcal{C}|}$ the softmax probability vector. Then, \emph{minimizing} $\ell_{out}(f(x))$ is equivalent to \emph{maximizing} the softmax entropy $E(f(x))$. This statement can be proven straightforwardly using Jensen's inequality. Since the softmax definition implies $f_j(x) \in (0,1) ~ \forall j\in \mathcal{C}$ and $\sum_{j\in\mathcal{C}} f_j(x)=1$, Jensen's inequality applied to the convex function $-\log(\cdot)$ yields 
\begin{equation}
	\ell_{out}(f(x)) \geq -\log \left(\sum_{j\in\mathcal{C}} \frac{1}{|\mathcal{C}|} f_j(x) \right)= \log(|\mathcal{C}|)
\end{equation}
and applied to the concave function $\log(\cdot)$
\begin{equation}
	E(f(x)) \leq \log \left(\sum_{j\in\mathcal{C}} f_j(x) \frac{1}{f_j(x)} \right)= \log(|\mathcal{C}|)
\end{equation}
with equality if $f_j(x)=\frac{1}{|\mathcal{C}|} ~ \forall j\in \mathcal{C}$.

\section{Separability by means of Data Distribution}\label{app:1}
The violin plots in \cref{fig:violins} visualize the separability of in-distribution and out-distribution pixels (binary classification) in LostAndFound and Fishyscapes, respectively. 
These plots summarize different statistics such as median and interquartile ranges and also show the full distribution of the data. 
The density corresponds to the relative pixel frequency at a given entropy value of the considered class. In the following, we refer to the shape of the violin plots as distribution.

First, we focus on evaluating LostAndFound OoD objects, see \cref{fig:violins} (a). For the baseline model we observe that a large mass of data corresponding to the negative class is located at very low entropy values (median $0.02$), \ie, most road pixels are classified with high confidence. Moreover, the 75th percentile is located at an entropy value of $0.04$ and the sample of highest value at $0.57$. Regarding the pixels of the positive class, we see that the distribution is rather dispersed. The median is at $0.29$ and the interquartiles range from $0.13$ to $0.44$. We conclude that, on average, positive samples have higher entropy values than negative ones, \ie, pixels of an OoD object are classified with higher uncertainty than for road pixels. 
However, for perfect performance one seeks a threshold such that both distributions (of the positive and negative class) are separated. This is not the case for the baseline model since a substantial amount of samples still has very low entropy, \eg\ the 10th percentile of the positive samples is at $0.04$, which is also the median of negative samples.

After OoD training, the distribution of negative samples remains in large parts similar compared to the baseline only with little changes. Noteworthy, the median and upper quartile decrease down to entropy values of $0.1$ and $0.2$, respectively. The distribution's maximum is at $0.66$. On the contrary, the changes of the distribution for the positive samples are significant as a large mass is concentrated at very high entropy values. The median is located at $0.59$ which is roughly at the same magnitude as the maximum for negative samples. Moreover, the minimum value for positive pixels is at $0.01$ which equals the median for negative samples. In particular the latter underlines the significant improvement of separability due to our OoD training. We observe the same behavior for Fishyscapes OoD objects but even more pronounced, see \cref{fig:violins} (b). After the OoD training, the medians of the two classes, $0.01$ for negative samples and $0.87$ for positive samples, differ by $86$ percent points. Besides, the lower quartile of positive samples at an entropy value of $0.71$ as well as the 1st percentile at $0.03$ are still above the median of negative samples. Consequently, we conclude that our OoD training is beneficial for identifying OoD pixels.

\section{Segment-wise Metrics for Meta Classifiers}\label{app:2}
As outlined in \cref{sec:meta-classifier}, we train meta classifiers based on hand-crafted metrics. These metrics are derived from the softmax probabilities $f(x)\in(0,1)^{|\mathcal{Z}| \times |\mathcal{C}|}, x\in\mathcal{X}$ of deep convolutional neural networks, information we get in every forward pass.
As a reminder, let $\hat{\mathcal{Z}}_{out}(x)$ be the set of pixel locations in image $x\in\mathcal{X}$ that are predicted to be OoD, see \cref{sec:setup}. A connected component $k \in \hat{\mathcal{K}}(x) \subseteq \mathcal{P}( \hat{\mathcal{Z}}_{out}(x))$ represents an \emph{OoD segment / object prediction} due to the entropy being above the given threshold. This is different to other works dealing with segment-wise meta classification \cite{Chan20, Maag20, Rottmann18, Rottmann19} as they consider connected components sharing the same class label as segments.

We estimate uncertainty per OoD segment $k$ by averaging pixel-wise scores at the segment's pixel locations $z\in k$. In addition to the plain softmax probabilities $f^z(x)$, we also incorporate three pixel-wise dispersion measures, namely $\forall z\in k$ the (normalized) entropy
\begin{equation}
    \bar E(f^z(x)) = -\frac{1}{|\mathcal{C}|} \sum_{j\in\mathcal{C}} f_j^z(x) \log(f_j^z(x))~,
\end{equation}
the variation ratio
\begin{equation}
    V(f^z(x)) = 1 - f_{\hat{c}(z)}^z(x)~,
\end{equation}
and the probability margin
\begin{equation}
    M(f^z(x)) = V(f^z(x)) + \max_{j \in \mathcal{C} \setminus \{ \hat{c}(z) \} } f_j^z(x)
\end{equation}
with $\hat{c}(z) := \arg \max_{j\in\mathcal{C}} f_j^z(x)$ being the class label according to the maximum a posteriori principle.

The segment's size $S(k)=|k|$ is not only needed for averaging but also serves as meta classification input on its own. Moreover, let $k_{in} \subset k $ be the set of pixel locations in the interior of the segment $k$, \ie, $k_{in}=\{ (h,w) \in k: [h \pm 1] \times [w \pm 1] \in k \}$. This also gives us the pixel locations of the boundary $k_{bd} = k \setminus k_{in}$.
In order to capture geometry features of a segment, we consider the relative sizes
\begin{equation}
    \tilde{S} = S / S_{bd} ~~\text{and}~~ \tilde{S}_{in} = S_{in} / S_{bd}
\end{equation}
by treating the segment's boundary and interior separately. 

For all metrics outlined up to now, we additionally consider the variances of the pixel-wise scores. They measure the deviation from the segment score mean and in our experiments, and it turns out that they have a great impact on meta classifiers for OoD detection.

Let $k_{nb}=\{ z^\prime \in [h \pm 1] \times [w \pm 1] \subset |\mathcal{H}| \times |\mathcal{W}|: (h,w) \in k, z^\prime \notin k \}, \mathcal{Z}=\mathcal{H}\times\mathcal{W}$ be the neighborhood of $k$. As metric if one segment is misplaced we include  
\begin{equation}
    N(j|k) = \frac{1}{|k_{nb}|} \sum_{z \in k_{nb}} \mathds{1}_{\{ j = \hat{c}(z) \}} ~ \forall ~ j\in\mathcal{C}
\end{equation}
which is the proportion of neighborhood pixels, with class $j\in\mathcal{C}$ having the highest softmax score, to neighborhood size. Another metric for localization purposes is the segment's geometric center
\begin{equation}
    C_h(k) = \frac{1}{S} \sum_{i=1}^{S} h_i ~~\text{and}~~ C_w(k) = \frac{1}{S} \sum_{i=1}^{S} w_i
\end{equation}
with $ z_i=( h_i, w_i ) \in k ~ \forall ~ {i=1,\ldots,|k|}$, \ie, averaging over the segment's pixel coordinates in vertical and horizontal direction.

For each segment $k$ we then have $m=83$ metrics in total (as $|\mathcal{C}|=19$ in our experiments). This forms a structured dataset 
\begin{equation}
    \mu \subseteq \mathbb{R}^{|\cup_{x\in\mathcal{X}} \hat{\mathcal{K}}(x)| \times m} = \mathbb{R}^{|\cup_{x\in\mathcal{X}} \hat{\mathcal{K}}(x)| \times 83} \label{eq:meta-input}
\end{equation}
serving as input for the meta classification model $g:\mu \rightarrow [0,1]$, the latter being a simple logistic regression in our case. By means of this linear model, we learn to discriminate whether a segment $k$ has an intersection with the ground truth (while all inputs are independent of the ground truth segmentation), see also \cref{eq:meta_classes}.

\section{OoD Training with Cityscapes void Class}\label{app:3}
\begin{figure}[t]
    \captionsetup[subfigure]{labelformat=empty}
    \centering
    \subfloat[]{\footnotesize\rotatebox{90}{\hspace{1.4cm}Entropy}}
    \subfloat[Baseline performance]{\includegraphics[width=0.23\textwidth]{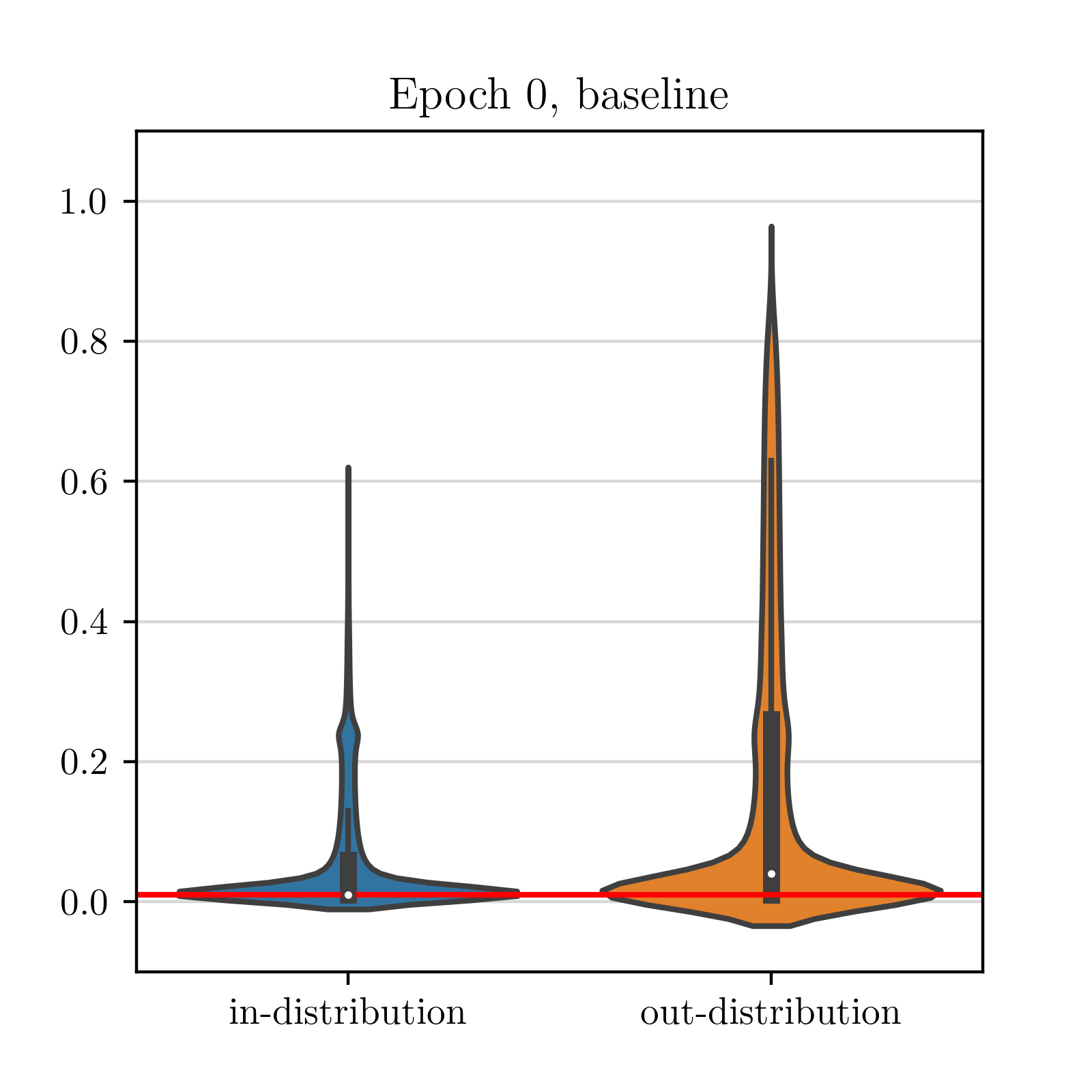}}
    \subfloat[Cityscapes void OoD training]{\includegraphics[width=0.23\textwidth]{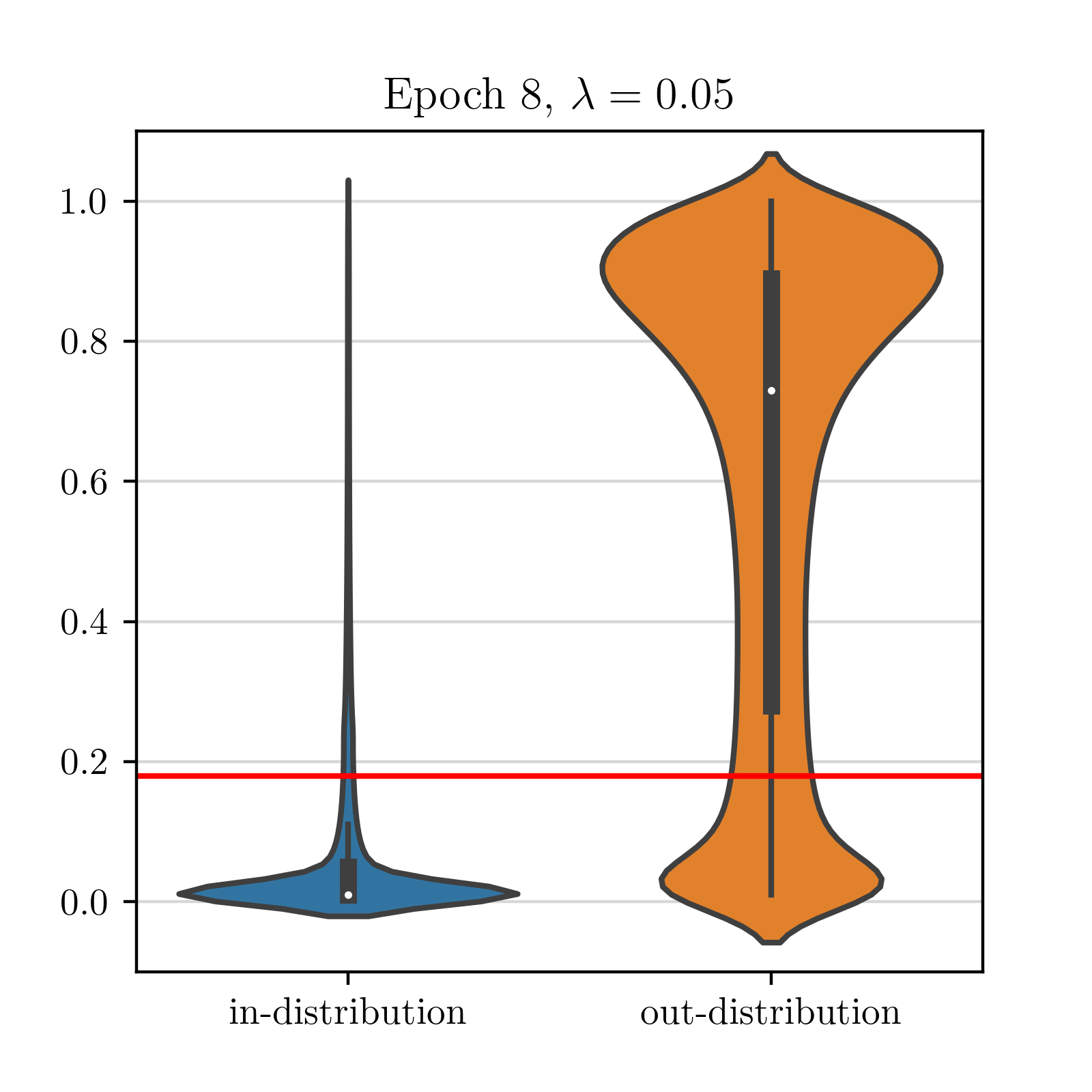}}
    \par\medskip
    \text{(\footnotesize Detecting Cityscapes unlabeled objects)}
    \caption{Separability between in-distribution and out-distribution pixels in Cityscapes. Pixels labeled as train class according to the ground truth are considered as in-distribution, pixels labeled with the void class as out-of-distribution. For the results with Cityscapes void OoD training the baseline model (left) was retrained with entropy maximization on the Cityscapes void class (right), \ie, using Cityscapes unlabeled objects as OoD proxy for $\mathcal{D}_{out}$.} \label{fig:cs_void_violins}
\end{figure}
\begin{figure}[t]
    \captionsetup[subfigure]{labelformat=empty}
    \centering
    \subfloat[]{\footnotesize\rotatebox{90}{\hspace{1.4cm}Entropy}}
    \subfloat[Baseline performance]{\includegraphics[width=0.23\textwidth]{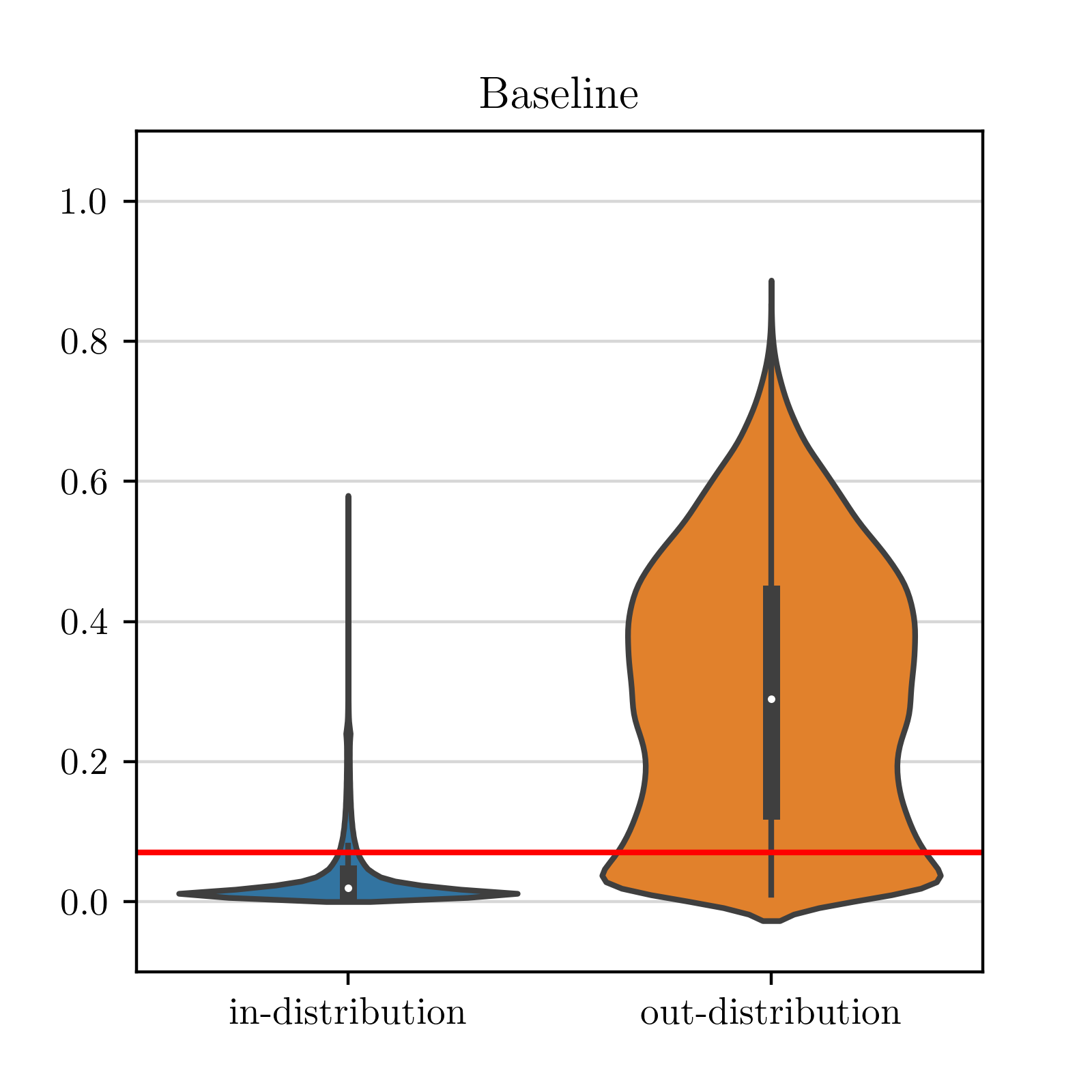}}
    \subfloat[Cityscapes void OoD training]{\includegraphics[width=0.23\textwidth]{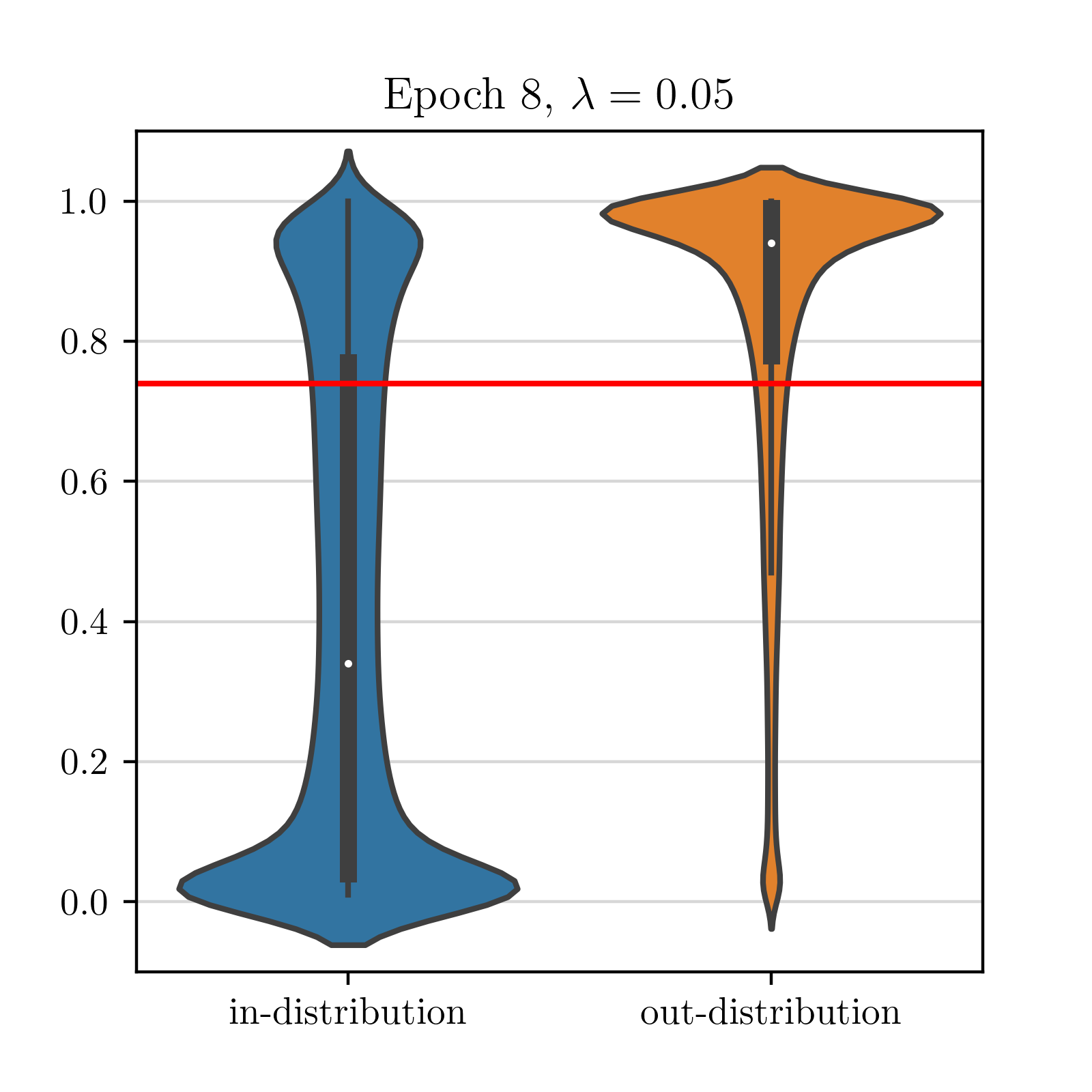}}
    \par\medskip
    \text{(\footnotesize Detecting LostAndFound OoD objects)}
    \caption{Separability between in-distribution and out-of-distribution pixels in the OoD dataset LostAndFound. For the results with Cityscapes void OoD training the baseline model (left) was retrained with entropy maximization on the Cityscapes void class (right).}
    \label{fig:laf_void_violins}
\end{figure}

\begin{figure*}[t]
    \captionsetup[subfigure]{labelformat=empty}
    \centering
    \subfloat[Cityscapes baseline segmentation]{\includegraphics[width=0.24\textwidth]{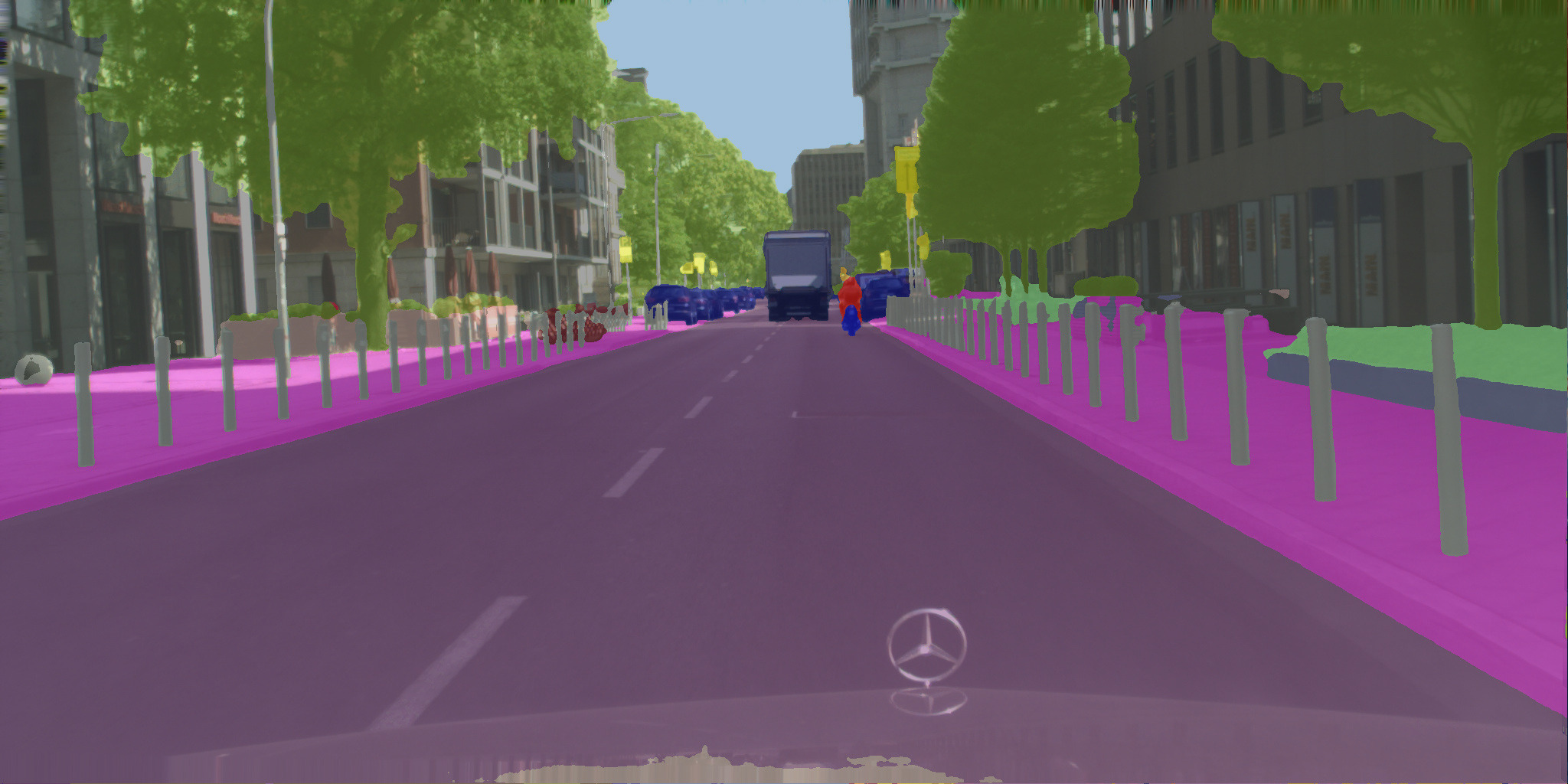}}~
    \subfloat[Cityscapes baseline entropy]{\includegraphics[width=0.24\textwidth]{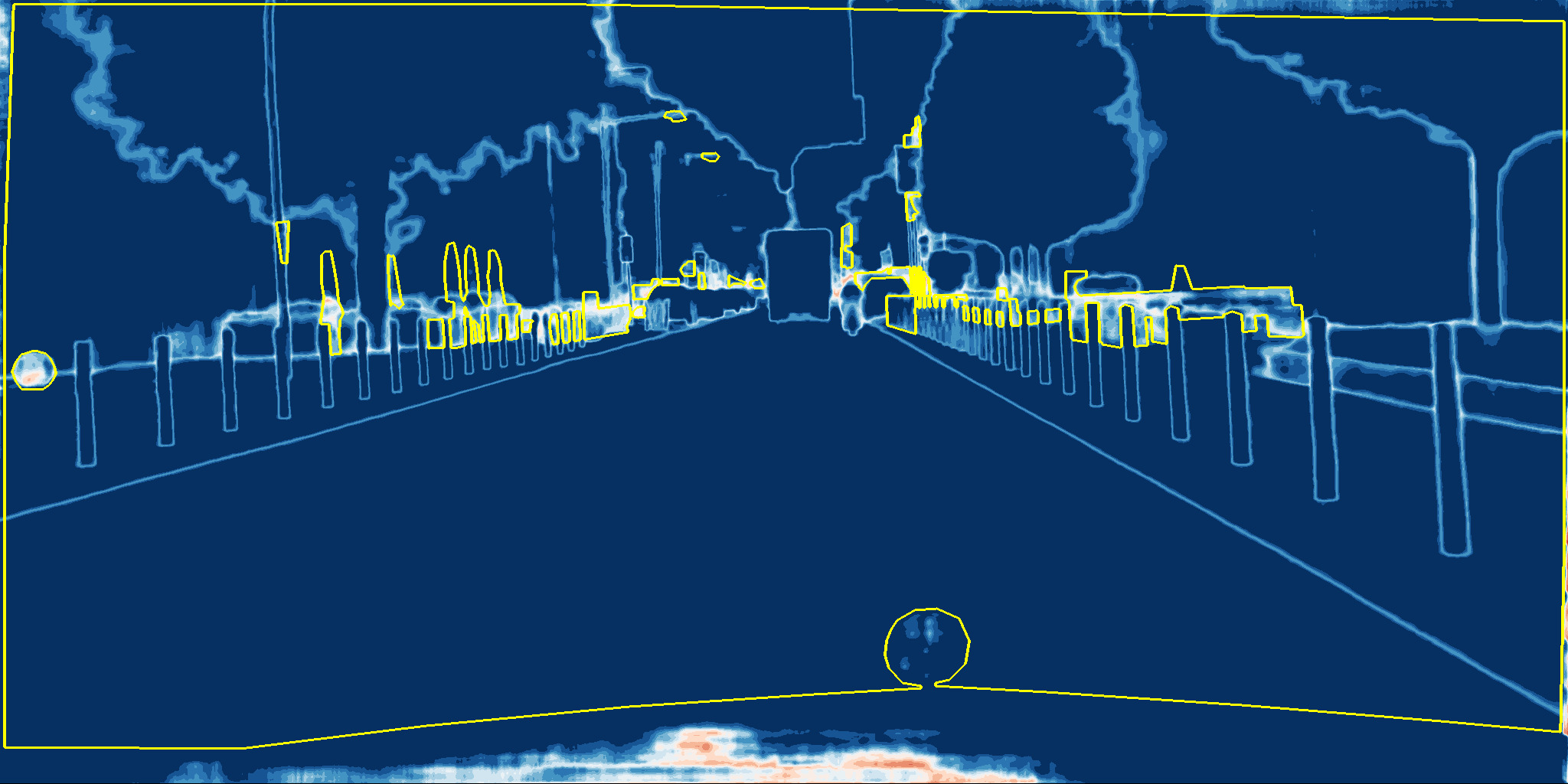}}~
    \subfloat[Lost\&Found baseline segmentation]{\includegraphics[width=0.24\textwidth]{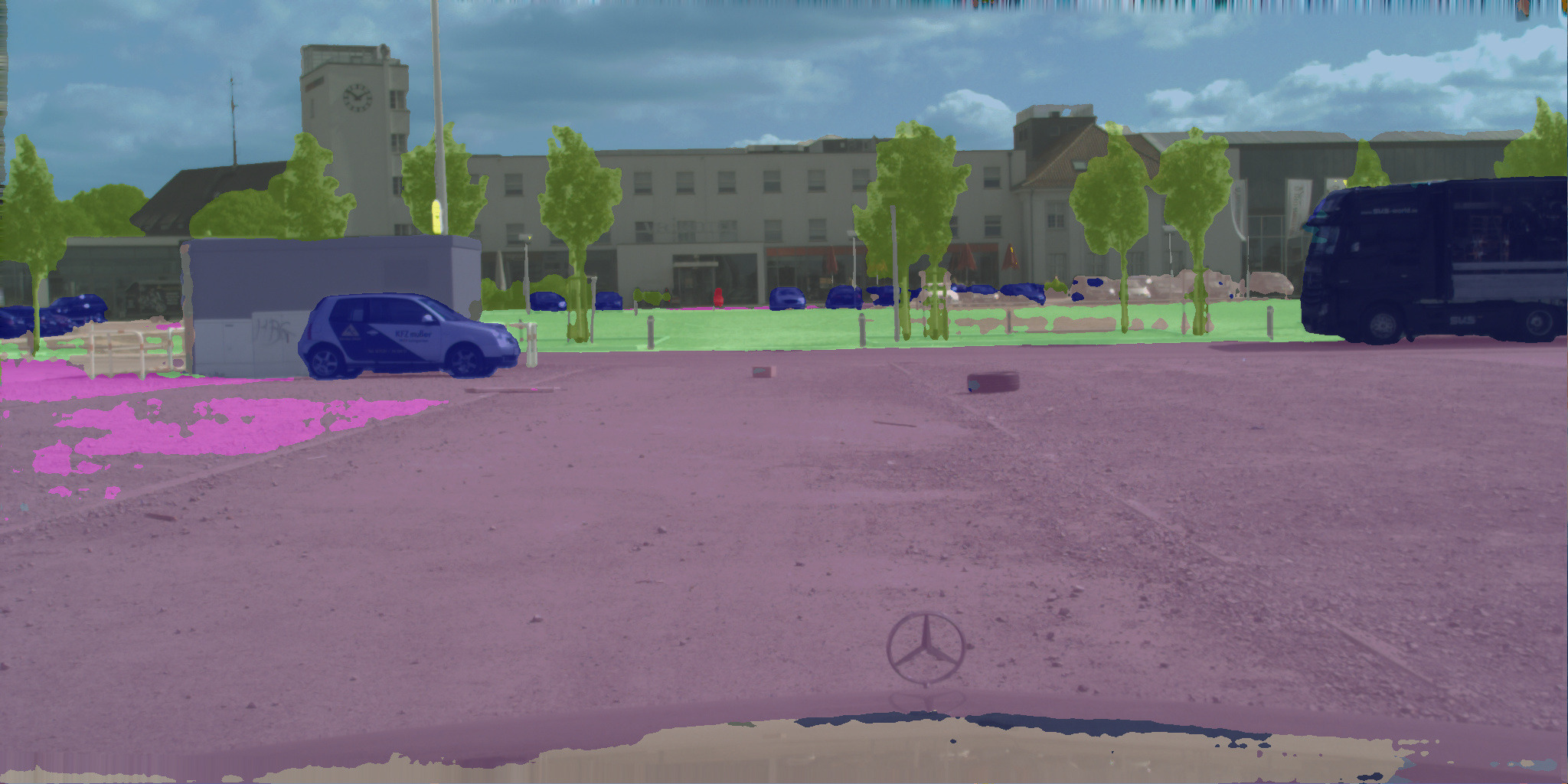}}~
    \subfloat[Cityscapes baseline entropy]{\includegraphics[width=0.24\textwidth]{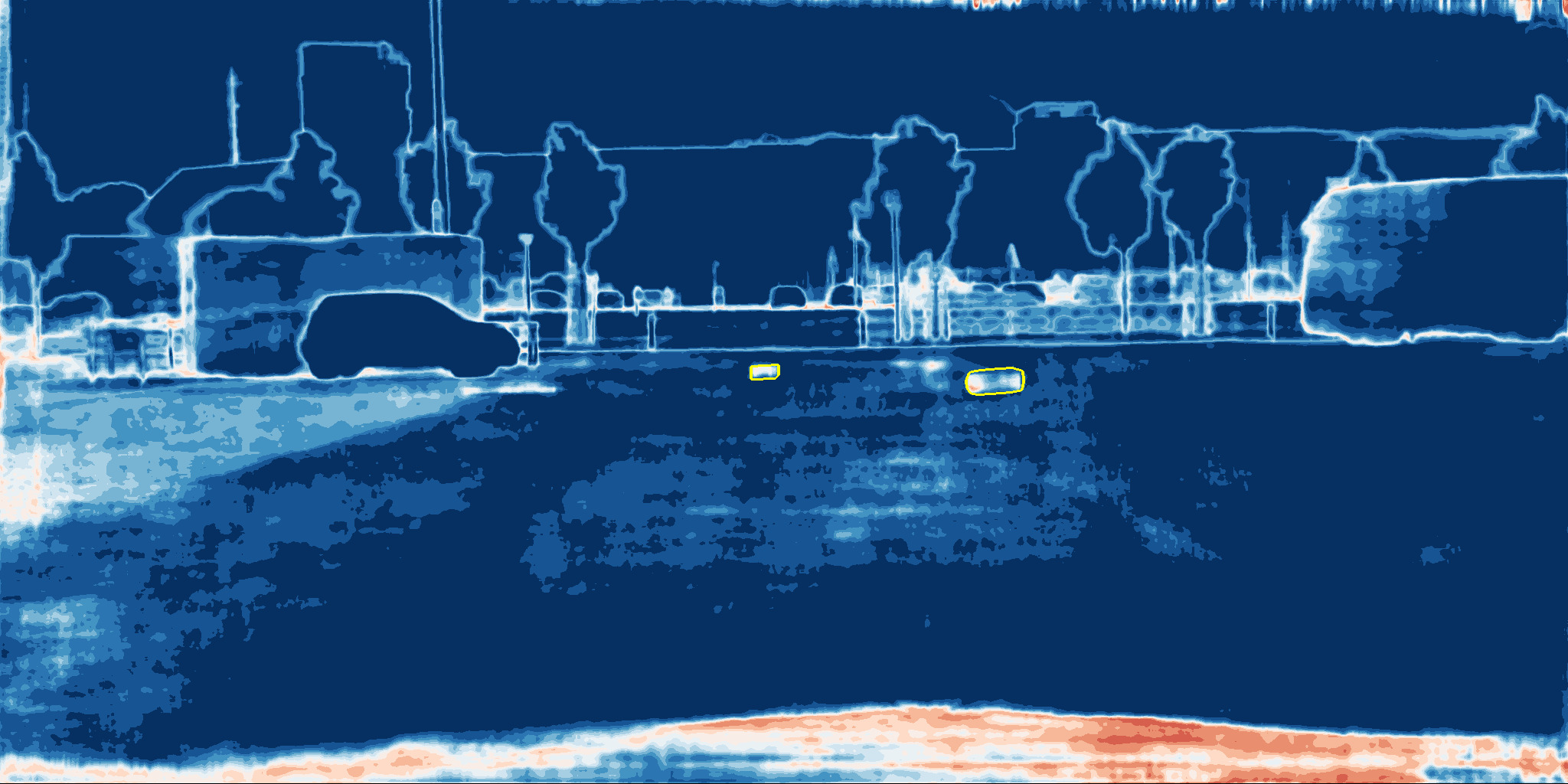}}\\
    \subfloat[Void OoD training segmentation]{\includegraphics[width=0.24\textwidth]{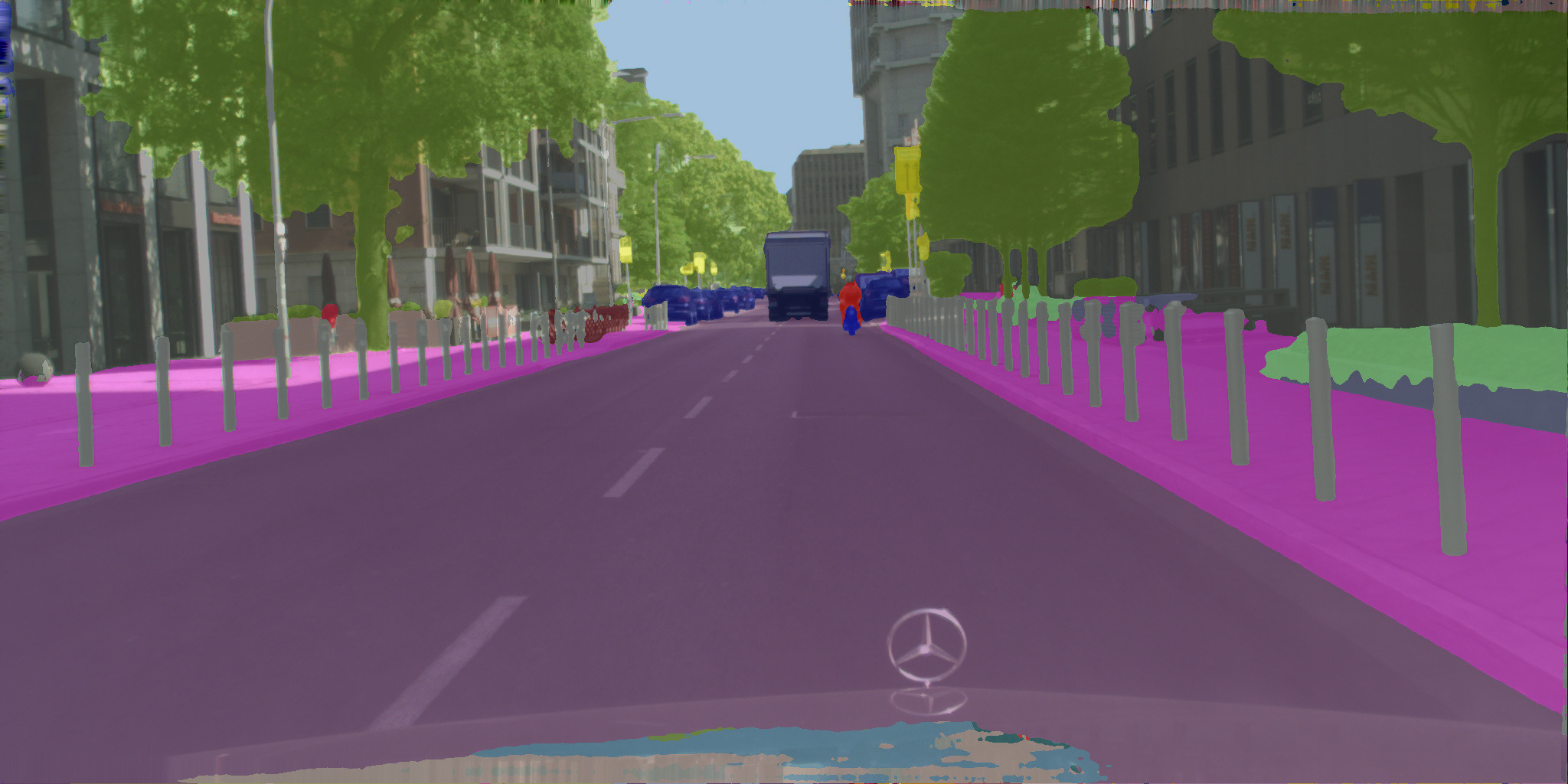}}~
    \subfloat[Void OoD training entropy]{\includegraphics[width=0.24\textwidth]{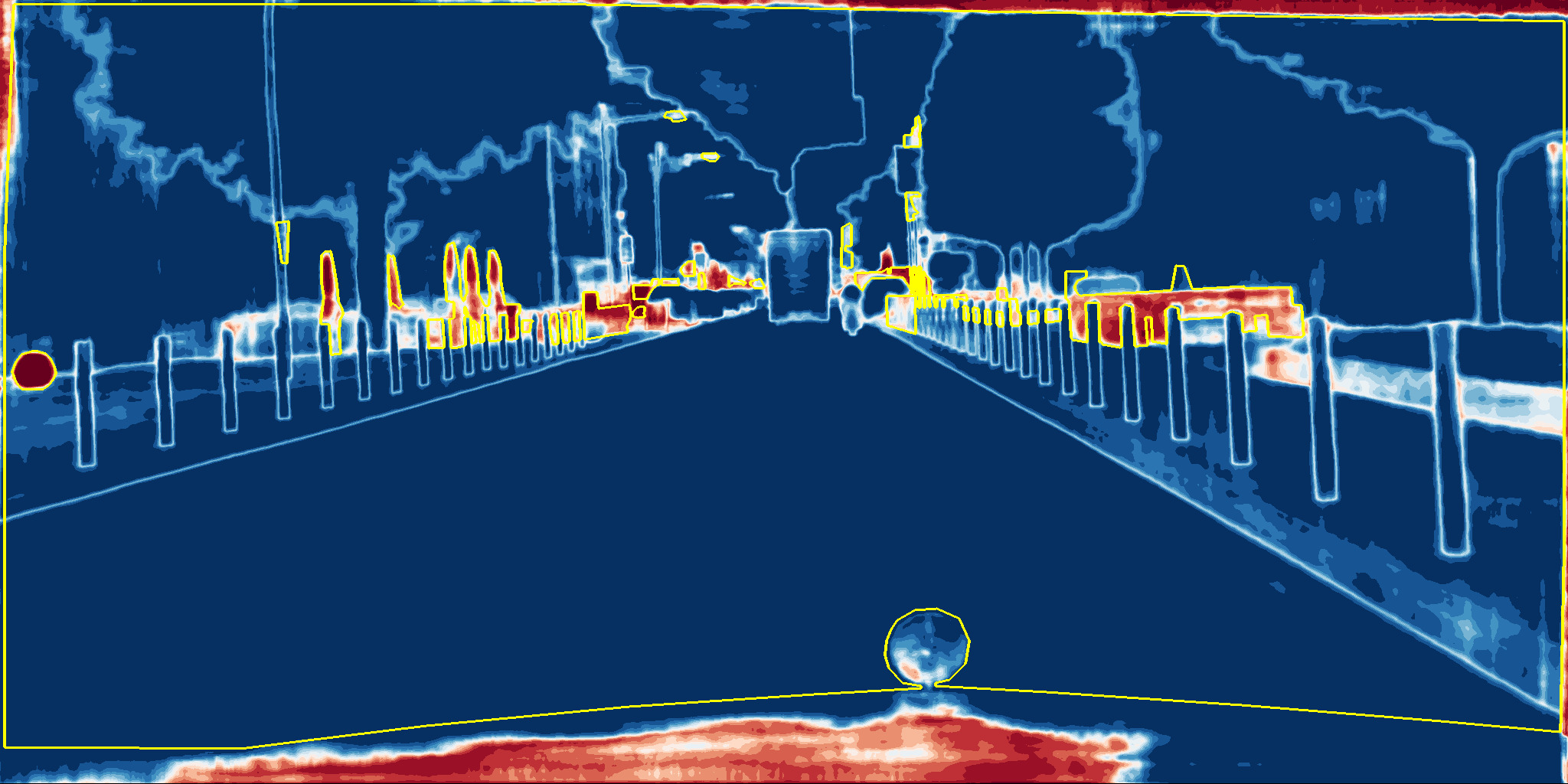}} ~
    \subfloat[Void OoD training segmentation]{\includegraphics[width=0.24\textwidth]{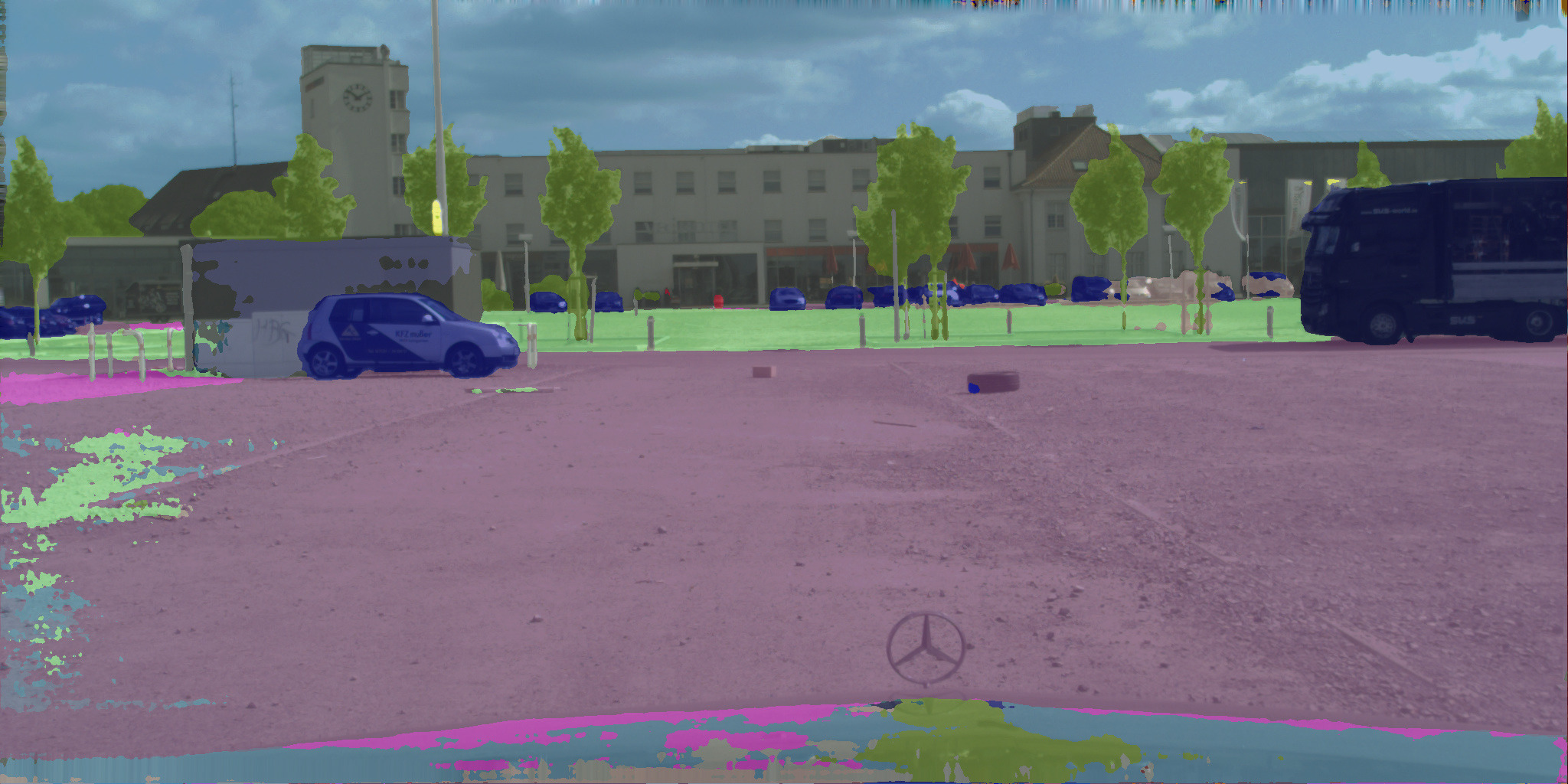}}~
    \subfloat[Void OoD training entropy]{\includegraphics[width=0.24\textwidth]{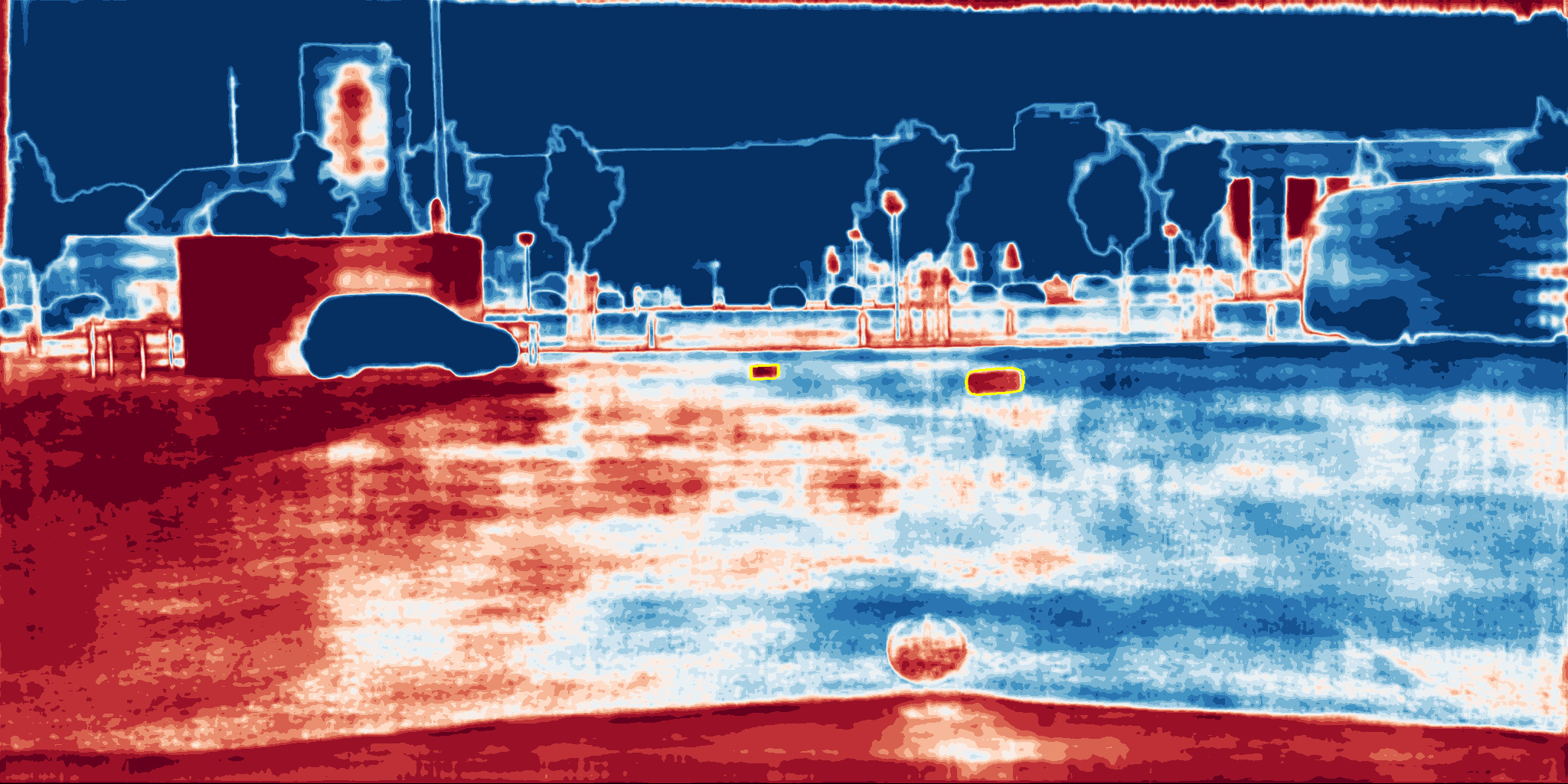}}
    \caption{Comparison between baseline model and retrained model, with entropy maximization on Cityscapes unlabeled objects, for one Cityscapes and one LostAndFound scene. The first and third column displays the segmentations obtained by the respective models on either a Cityscapes or LostAndFound input image, the second and forth column displays the corresponding entropy heatmaps. In the entropy heatmaps, the OoD objects are marked with yellow lines.} \label{fig:void_comp}
\end{figure*}

Before using the COCO dataset as OoD proxy, we conducted some experiments with the Cityscapes void class as OoD proxy for $\mathcal{D}_{out}$ in order to perform entropy maximization. This class includes objects that cannot be assigned to any of the Cityscapes training classes, therefore they remain unlabeled and are ignored during training. We refer to this retraining approach using the Cityscapes unlabeled objects as OoD proxy as \emph{void OoD training}. We find the best results in our experiments for the DeepLabv3+ as baseline model after 8 epochs of void OoD training and out-distribution loss weight of $\lambda=0.05$. With respect to the Cityscapes validation dataset, the retrained model clearly improves at identifying unseen unlabeled objects, see \cref{fig:cs_void_violins}.

However, the same retrained model fails to generalize to unseen OoD objects available in the LostAndFound dataset, see \cref{fig:laf_void_violins}. Not only the softmax entropy of OoD pixels is boosted but also the entropy of a significant amount of in-distribution pixels. This is even more considerable due to the strong class imbalance in LostAndFound. With respect to the AUROC, the void OoD training decreases the OoD detection score by 5 percent points down to 0.88, while decreasing the more relevant metric AUPRC by even 29 percent points down to 0.17 compared to the baseline model.

\begin{figure*}[t!]
    \centering
    \subfloat[LostAndFound]{\includegraphics[width=0.24\textwidth]{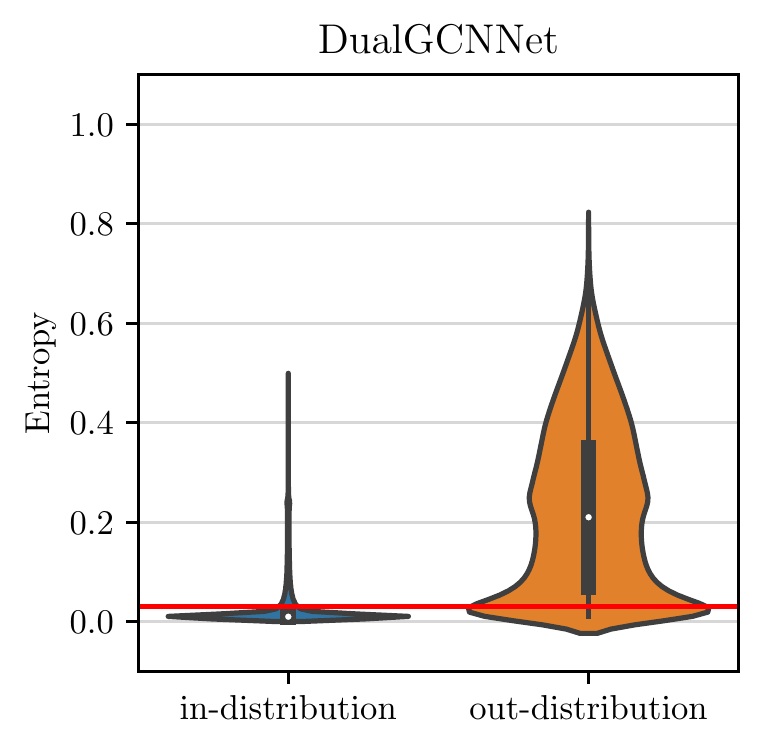} \includegraphics[width=0.24\textwidth]{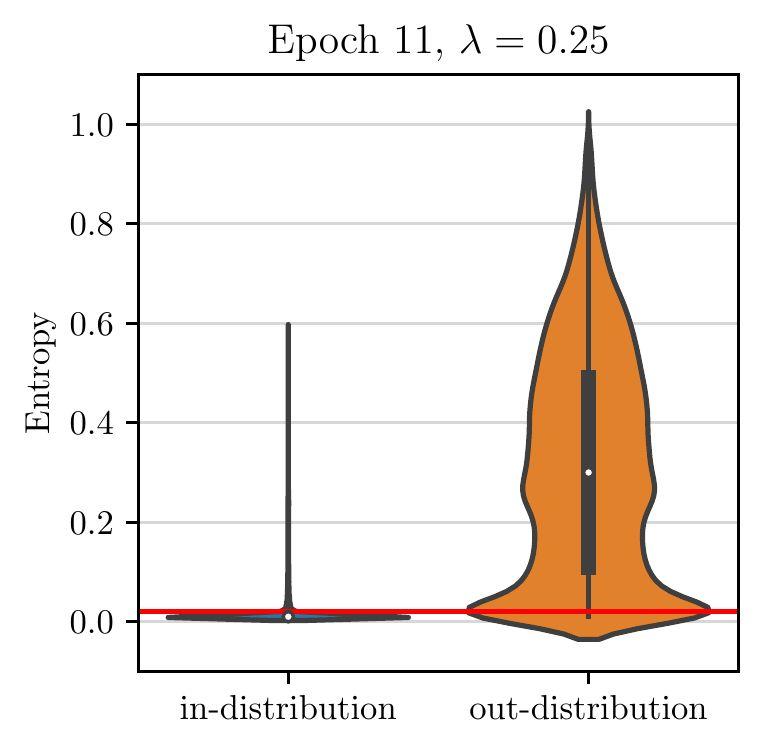}}~
    \subfloat[Fishyscapes]{\includegraphics[width=0.24\textwidth]{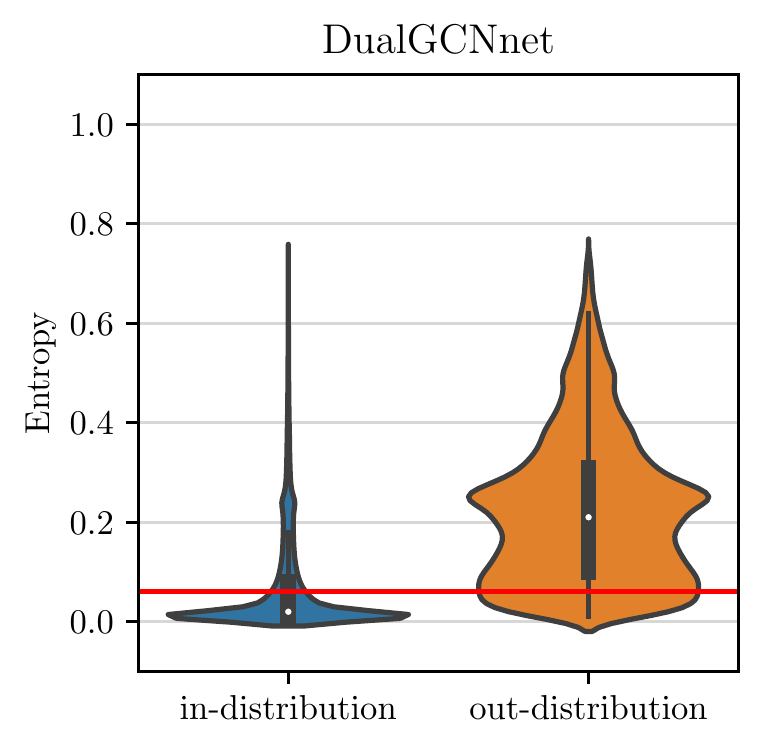} \includegraphics[width=0.24\textwidth]{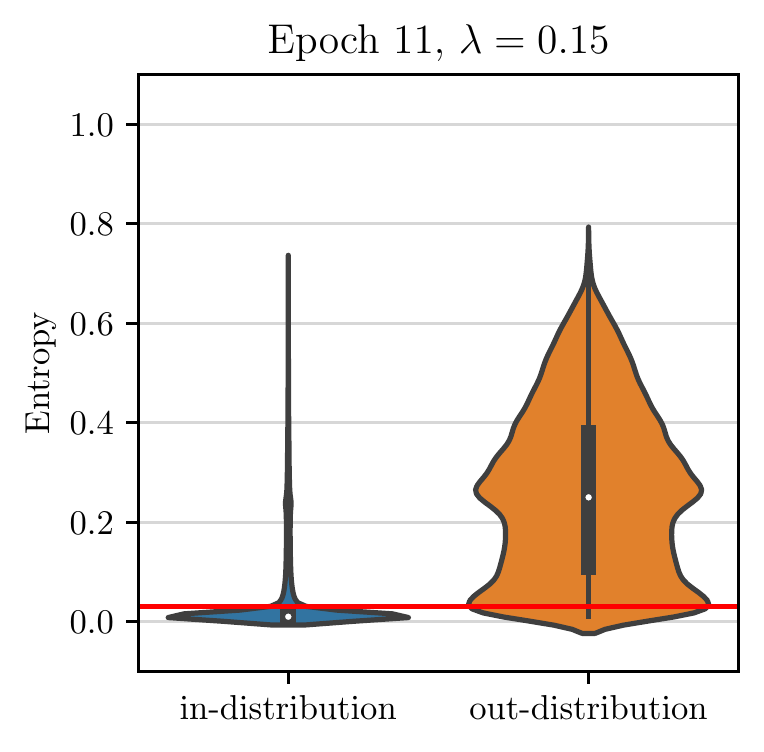}}~
    \caption{Relative pixel frequencies of LostAndFound (a) and Fishyscapes (b) OoD pixels, respectively, at different entropy values for the baseline model, \ie, before OoD training (\emph{a \& b left}), and after OoD training (\emph{a \& b right}). The red lines indicate the thresholds of highest accuracy.}
    \label{fig:violins_dualgcnnet}
\end{figure*}

A visual comparison of the effects of void OoD training is shown is \cref{fig:void_comp}. The retraining does not noticeably impact the segmentation performance, neither for Cityscapes nor LostAndFound. In particular for the segmentation of the Cityscapes scenes, there are only minor differences visible, \ie, the difference in performance for the original task is marginal. This is in line with the observation that retraining with the multi-criteria loss function, see equation \cref{eq:obj}, and the COCO dataset as OoD proxy leads only to a marginal loss of mIoU for the Cityscapes validation dataset. With respect to the Cityscapes images, the softmax entropy inside unlabeled objects is clearly boosted due to void OoD training. This makes identifying such objects easier in comparison to the baseline model.

Regarding the LostAndFound the differences in segmentations are more visible although still not being significant. On the contrary, by comparing the entropy heatmaps for the baseline model and the model after void OoD training, one observes that not only the entropy of pixels inside the OoD objects is boosted but also many in-distribution pixels. This detrimentally impacts the discrimination performance between in-distribution and out-distribution pixels as these two classes cannot be separated well via entropy thresholding. 
This supports the impression of the pixel-wise evaluation that void OoD training is not suitable for the detection of objects other than the Cityscapes unlabeled objects.

\begin{figure*}[t]
    \captionsetup[subfigure]{labelformat=empty}
    \centering
    \subfloat[Before OoD Training]{\includegraphics[width=0.48\textwidth]{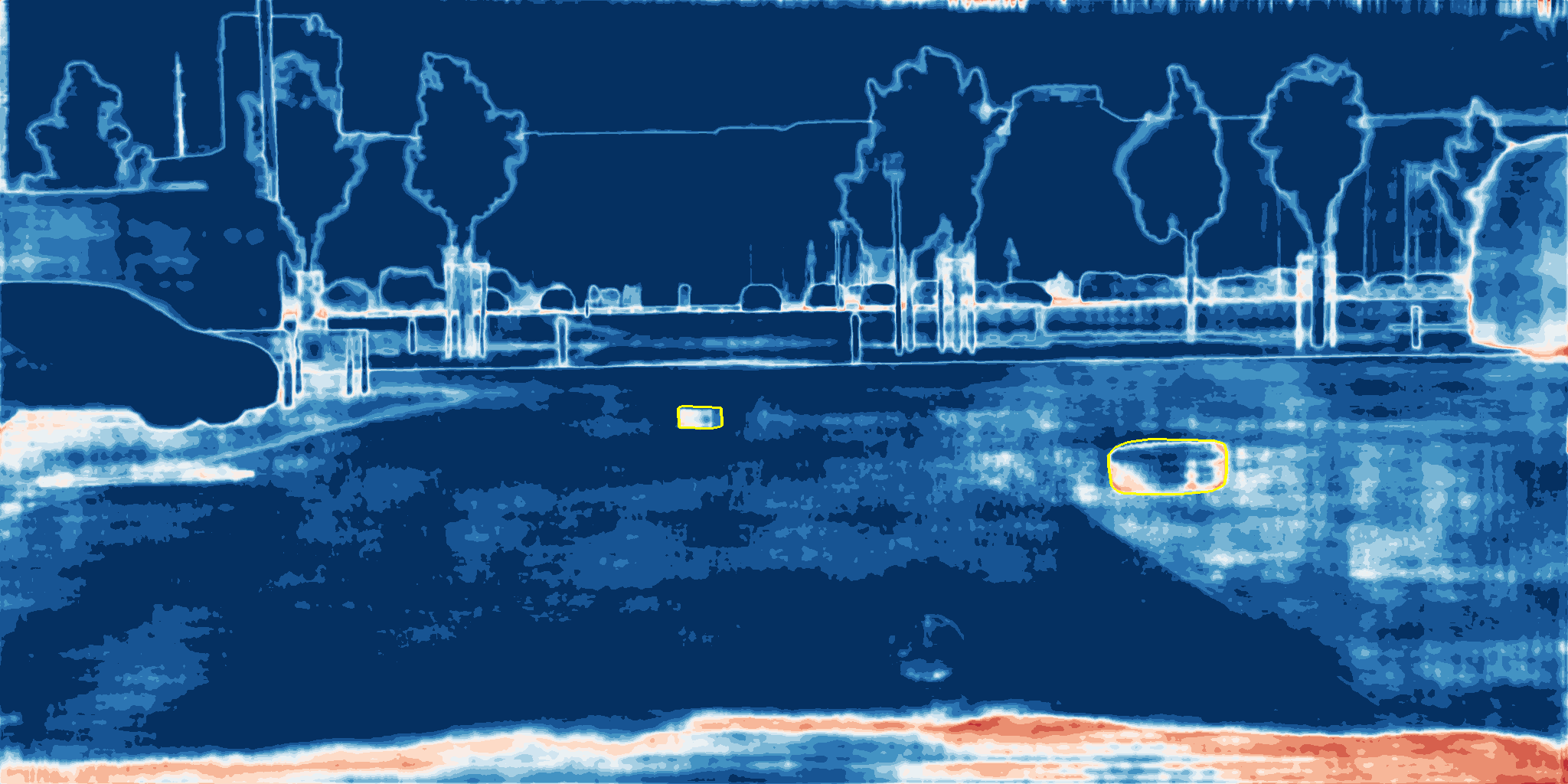}}~
    \subfloat[After OoD training]{\includegraphics[width=0.48\textwidth]{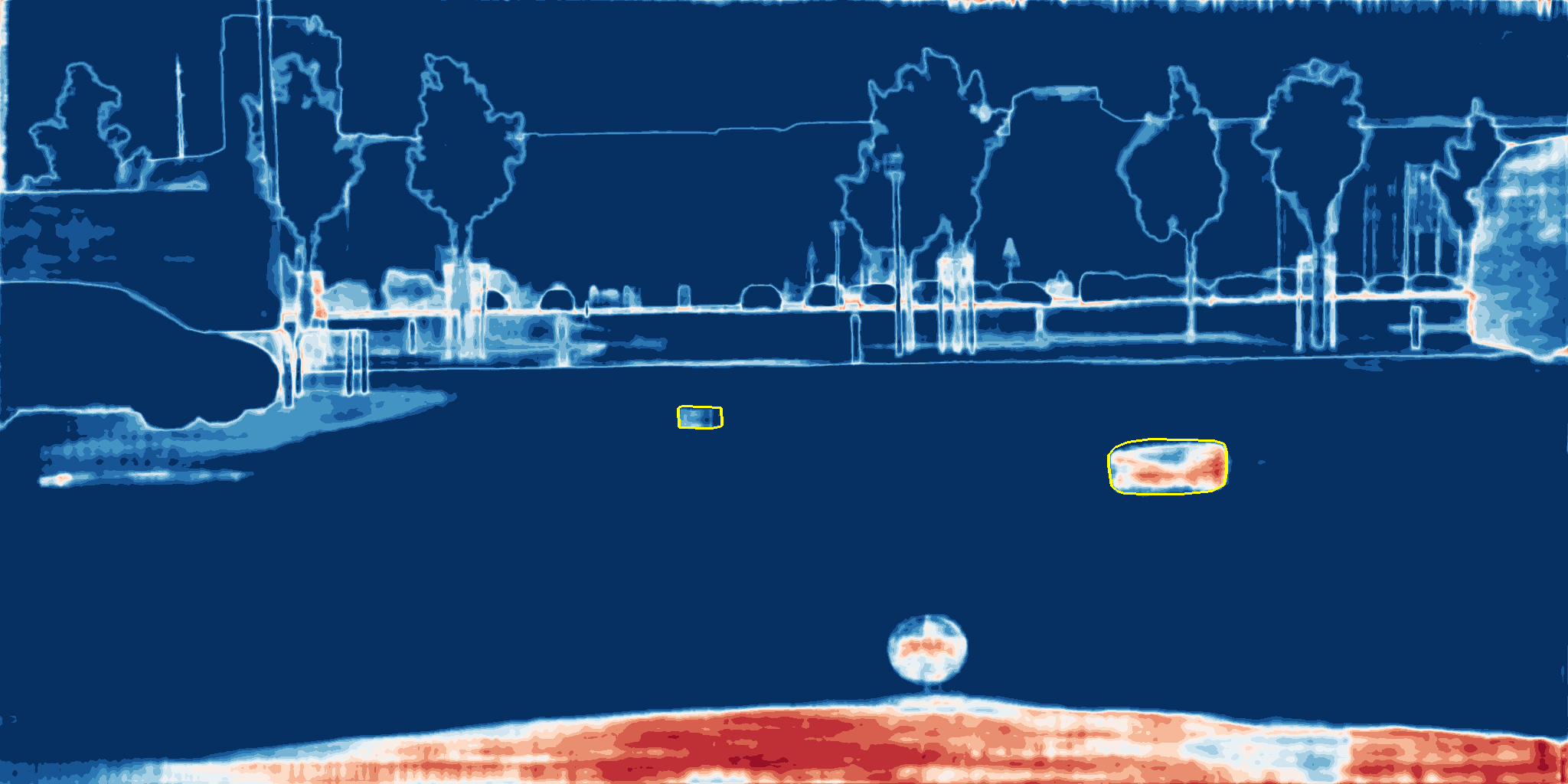}}
    \caption{Comparison of softmax entropy heatmaps before (left) and after OoD training (right). The yellow lines mark the OoD objects according to their ground truth annotation.}
    \label{fig:ent_ood_comp}
\end{figure*}

\section{OoD Training for DualGCNNet}\label{app:4}
\begin{figure*}[t]
    \centering
    ~~~~~~~\includegraphics[width=0.4\textwidth]{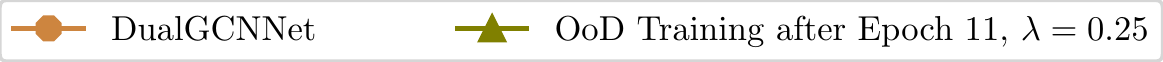}\\
    \vspace{-.3cm}
    \subfloat[LostAndFound (\emph{left:} AUROC, \emph{right:} AUPRC)]{\includegraphics[width=0.235\textwidth]{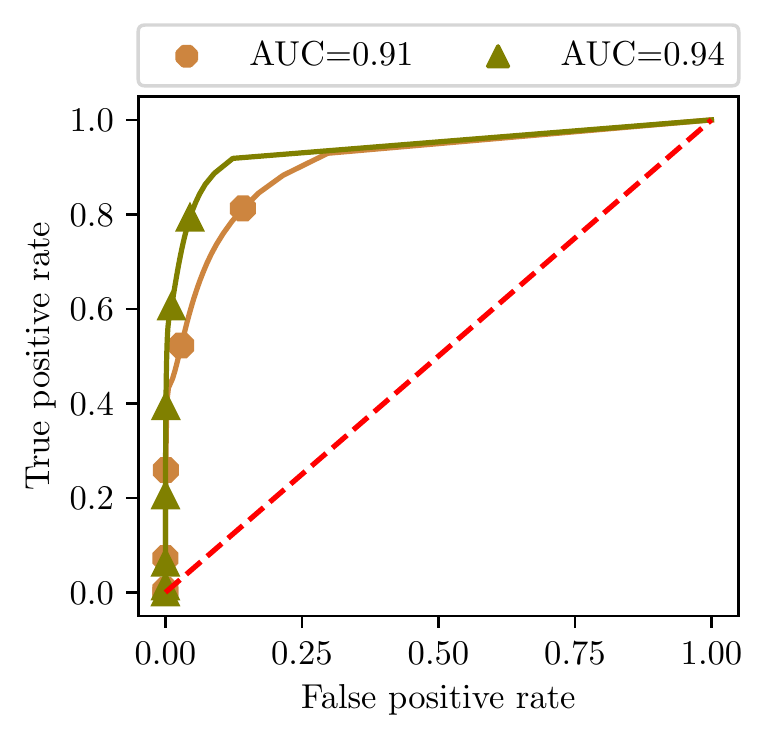} \includegraphics[width=0.235\textwidth]{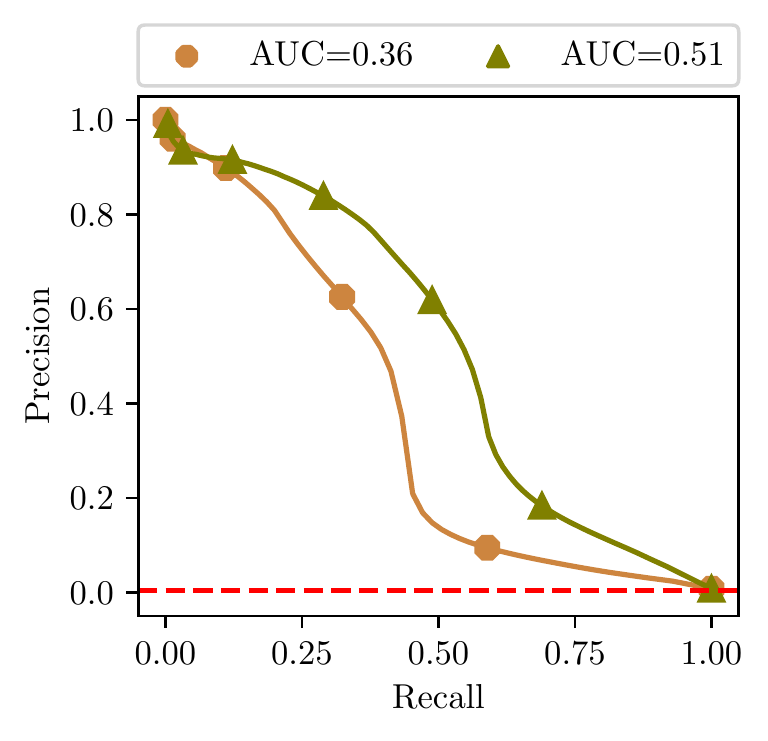}}~
    \subfloat[Fishyscapes (\emph{left:} AUROC, \emph{right:} AUPRC)]{\includegraphics[width=0.235\textwidth]{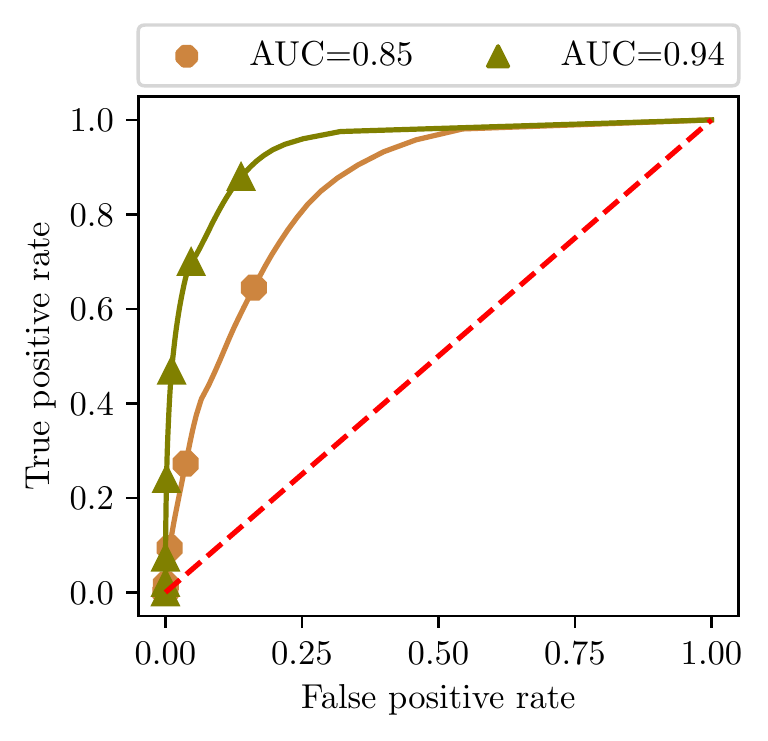} \includegraphics[width=0.235\textwidth]{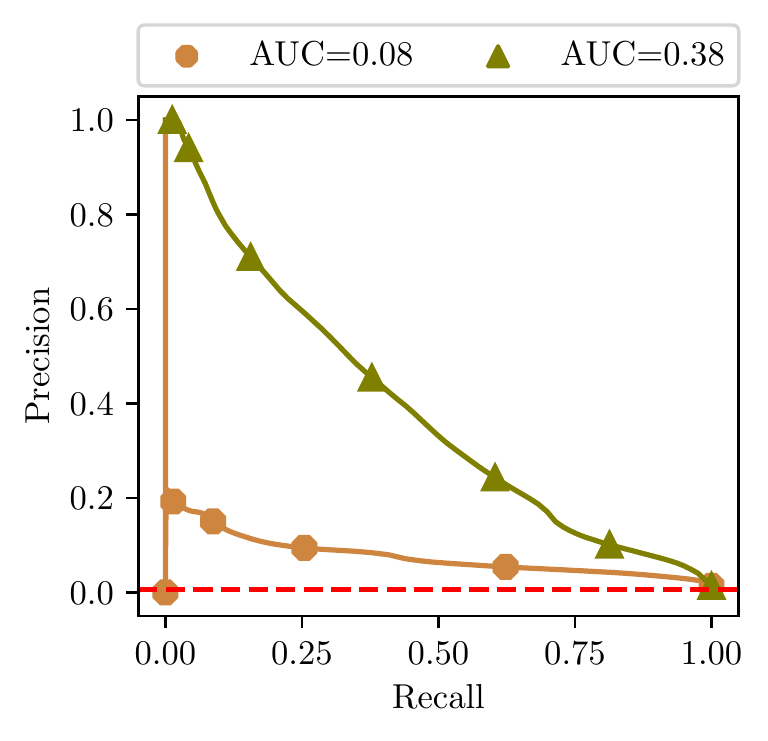}}
    \caption{Detection ability of LostAndFound (a) and Fishyscapes (b) OoD pixels, respectively, evaluated by means of receiver operating characteristic curve (\emph{a \& b left}) and precision recall curve (\emph{a \& b right}). The red lines indicate the performance according to random guessing.}
    \label{fig:auc_dualgcnnet}
\end{figure*}

As a second model complementary to the DeepLabv3+ model, we conducted same experiments of OoD training, \ie, retraining with the COCO dataset as OoD proxy, with the DualGCNNet which is a weaker and more lightweight network compared to the state-of-the-art DeepLabv3+ segmentation network. We find the best results after 11 epochs of OoD training with out-distribution loss weight of $\lambda=0.25$. As optimizer we used Adam with a learning rate of $10^{-6}$. 

We evaluated the OoD detection for the LostAndFound test and Fishyscapes Validation dataset in a similar manner as in the experiments for DeepLabv3+. For the DualGCNNet model, however, we only compare OoD training against entropy thresholding with the original model. Entropy thresholding with DualGCNNet in its original version is a weak OoD detector with AUPRC-scores of $0.36$ and $0.07$ for LostAndFound and Fishyscapes, respectively, see \cref{tab:benchmark}.
We observe that OoD training is not as effective as for the DeepLabv3+ model in terms of absolute performance gain. However, we still observe a decent improvement in separability. By applying OoD training, the AUROC increases by 3 percent points for LostAndFound and even 9 percent points for Fishyscapes up to a score of 0.94 for both datasets. With respect to the PR curves, the AUC improves by 15 percent points up to 0.51 for LostAndFound and by 20 percent points up to 0.38 for Fishyscapes. Noteworthy, these AUC scores after OoD training are higher than for the plain DeepLabv3+ (baseline) model which is already a strong OoD detection model.

These results for the weaker DualGCNNet model further demonstrate the positive effect on the OoD detection ability when performing OoD training with the COCO dataset as OoD proxy. The pixel-wise evaluation results are reported by means of the violin plots in \cref{fig:violins_dualgcnnet} and by ROC as well as PR curves in \cref{fig:auc_dualgcnnet}.

\section{OoD Training Visualization}\label{app:5}
The improved separation ability due to OoD training is not only achieved by increasing the softmax entropy of OoD pixels but also by decreasing the softmax entropy for in-distribution pixels. This can be also observed by means of the in-distribution violins, for instance in \cref{fig:violins_dualgcnnet}. By comparing the shapes of the violins corresponding to the DualGCNNet plain model and the model after OoD training, we notice that the violin shapes remain similar in large parts. The median and the upper quartile, however, decrease down to lower entropy values after OoD training. This indicates that after entropy maximization the model is on the one hand more uncertain at OoD pixel locations and on the other hand more certain about its prediction at in-distribution pixel locations. The same observation also holds for the DeepLabv3+ model, see \cref{fig:violins}. 
This is in line with the observation made in \cite{zhang17universum} that training with an OoD proxy may have a regularizing effect.

An illustration is provided in \cref{fig:ent_ood_comp}. For comparison purposes, we refer to the entropy heatmaps provided in \cref{fig:void_comp} as both figures show the same scene. The visualization of heatmaps clearly shows that due to OoD training pixels with high entropy are more concentrated inside OoD objects. Moreover, the in-distribution objects, especially the pixels corresponding to the road, have lower entropy values than before OoD training. This makes the road seem cleaner with respect to the possible occurrence of OoD objects. After entropy maximization the OoD objects are (visibly) better recognizable within the softmax entropy heatmaps. Therefore, we expect that the meta classification performance is improved as the meta classifiers are able to estimate the shape of OoD objects even better. Moreover, higher entropy values are stronger correlated with the presence of OoD objects.

\section{Found Objects due to OoD Training}
\begin{figure}[t]
    \centering
    \includegraphics[width=0.95\linewidth]{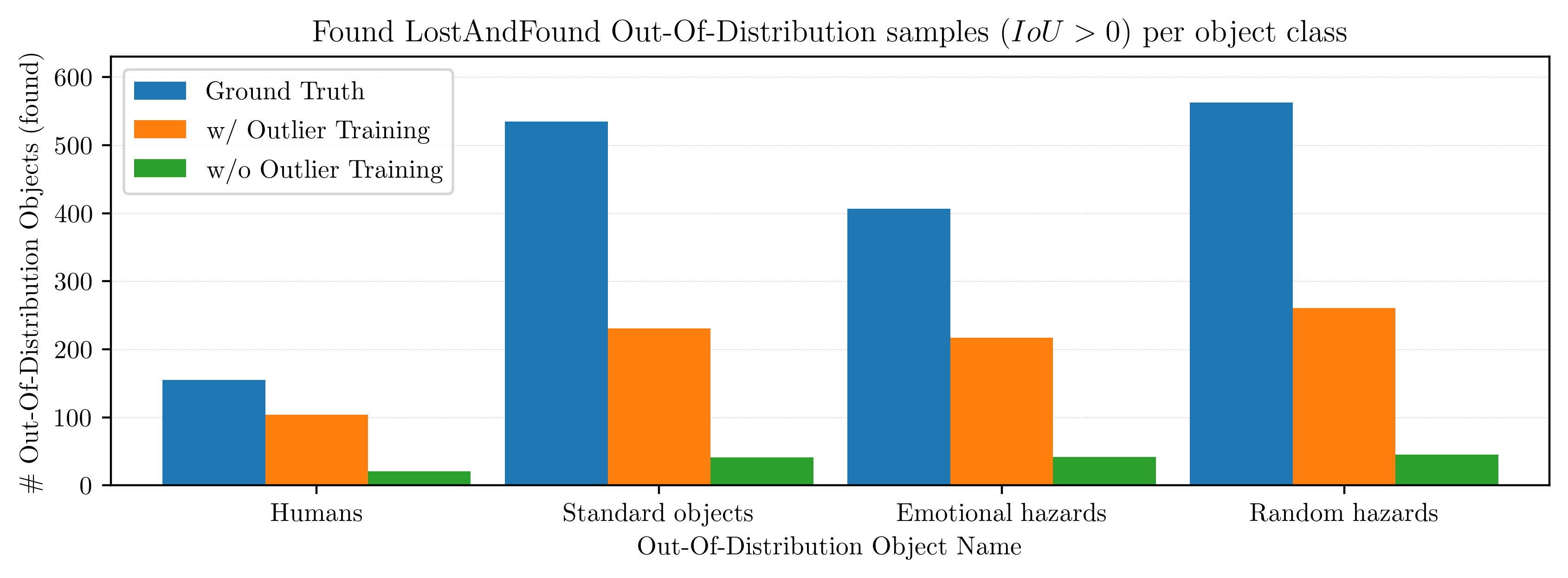}\\
    \caption{Overview of detected LostAndFound objects (per object class) with $t=0.7$ before (green) and after OoD Training (orange). The blue bar indicates the number of ground truth instances that can be found in total.}
    \label{fig:found}
\end{figure}

The objects in the LostAndFound dataset comprises four road harzard classes:
\begin{itemize}
    \item \emph{humans}: kids (with toys) on the road
    \item \emph{standard object}: crates in different shapes and colors
    \item \emph{emotional hazards}: bobby car, ball, dog, \etc
    \item \emph{random hazards}: bumper, euro pallet, pylon, tire \etc
\end{itemize}
\Cref{fig:found} illustrates the found objects via softmax entropy thresholding with the baseline model and also the model after OoD Training. One clearly notices that the number of overlooked OoD objects from all classes is significantly reduced due to OoD training.

\section{Course of OoD Training}\label{app:7}

\begin{figure}[t]
    \centering
    \includegraphics[width=0.9\linewidth, trim=0 6.4cm 0 0, clip]{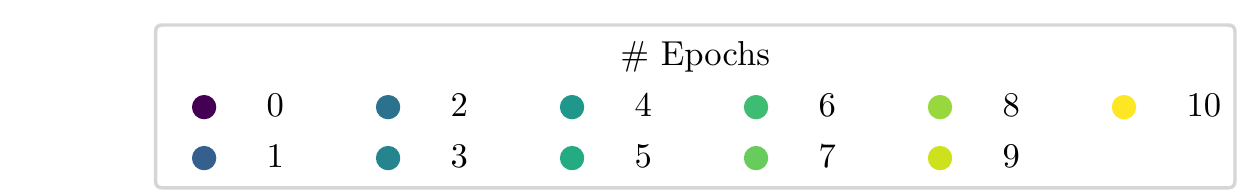}
    \includegraphics[width=0.95\linewidth]{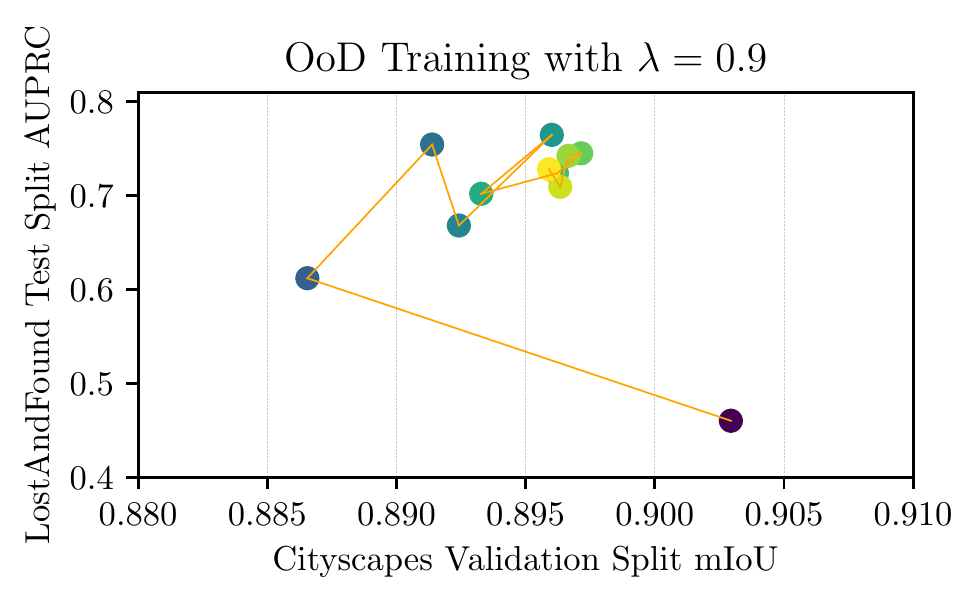}
    \caption{Mean intersection over union (mIoU) for the Cityscapes validation dataset split over the course of OoD training.}
    \label{fig:miou_over_training}
\end{figure}

In order to monitor that the baseline model does not unlearn its original task due to OoD training, we evaluate the model's original task performance over the training epochs. We evaluate the mIoU on the Cityscapes validation dataset against the AUPRC on the LostAndFound test dataset, displayed in \cref{fig:miou_over_training}. The state-of-the-art DeepLabv3+ model, which serves as baseline throughout our experiments, achieves an mIoU of $90.30\%$ when equipped only with the standard maximum a posteriori (MAP) decision principle while the same model has an entropy based OoD detection performance of $46.01\%$ in AUPRC. By fine tuning the neural network with entropy maximization on OoD inputs, we on the one hand sacrifice only little in mIoU (of the original task). On the other hand, we observe improved AUPRC scores over the course of training epochs peaking at $76.45\%$. This considerable gain at detecting OoD samples in LostAndFound comes with a marginal loss in Cityscapes validation mIoU of less than $1$ percent point. Moreover, the course of the OoD training illustrates convergence around the best AUPRC score with an mIoU loss that is in the same range as for the best score after OoD training. Concerning the overall performance of perception systems that rely on semantic segmentation, \eg, in applications like automated driving, this is a favorable trade-off in terms of safety that comes with very little computational overhead.

\section{Meta Classification Visualization}\label{app:8}

\begin{figure*}[t!]
    \begin{minipage}[t]{0.47\textwidth}
    \captionsetup[subfigure]{labelformat=empty}
    \centering
    \subfloat[Baseline plain]{\includegraphics[width=0.49\textwidth]{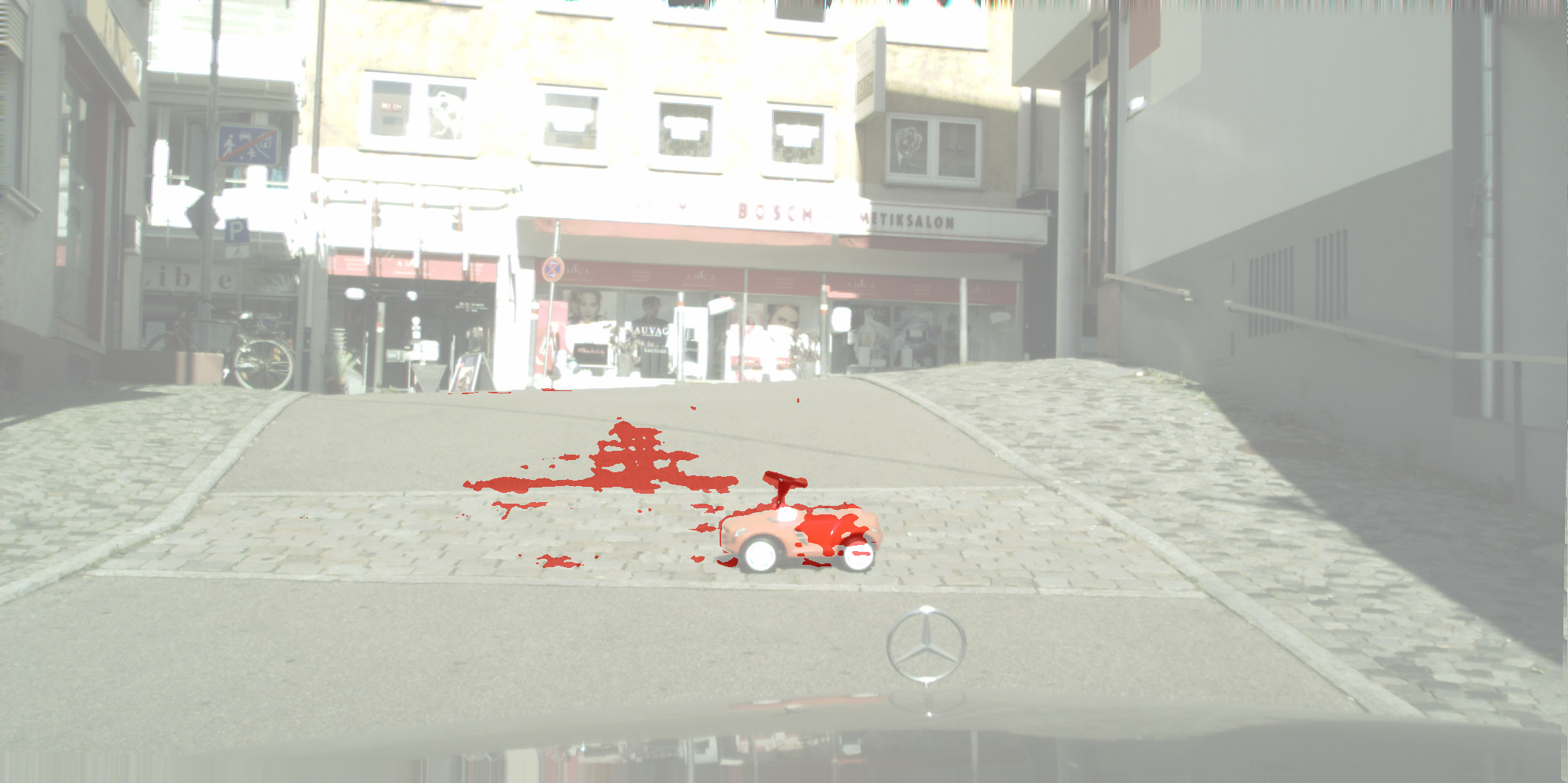}}~
    \subfloat[Baseline plain + meta classifier]{\includegraphics[width=0.49\textwidth]{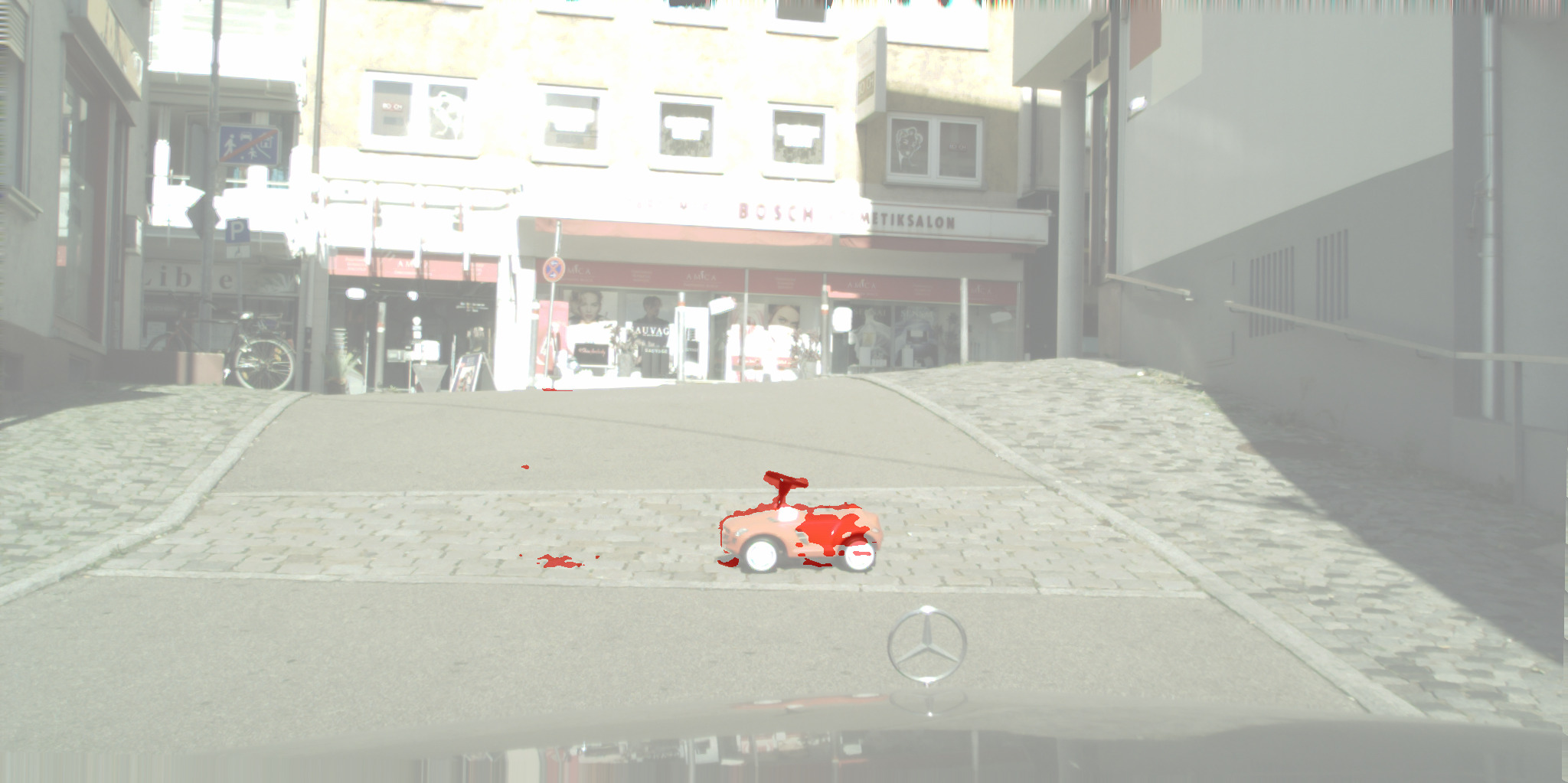}} \\
    \subfloat[OoD training]{\includegraphics[width=0.49\textwidth]{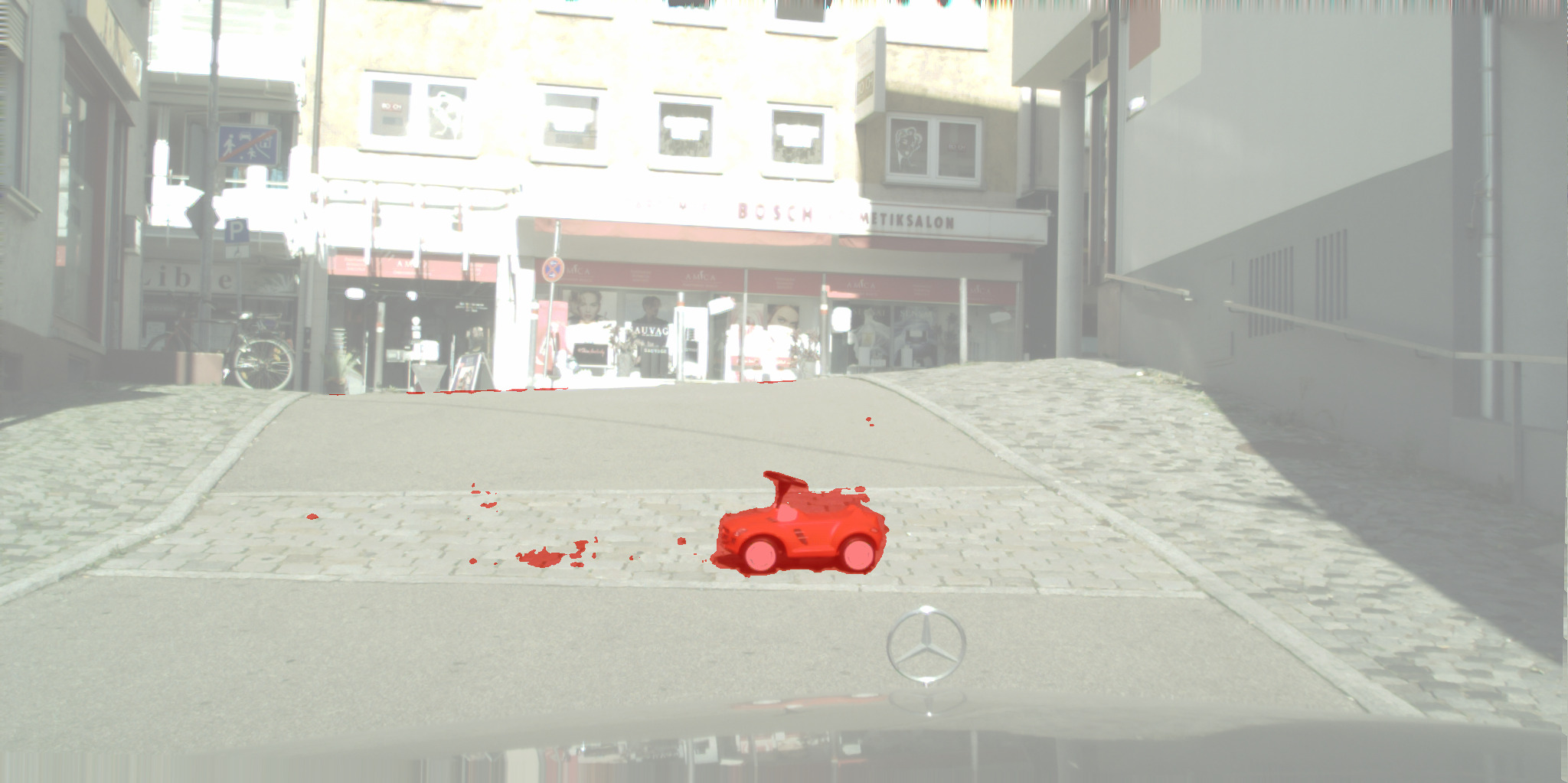}}~
    \subfloat[OoD training + meta classifier]{\includegraphics[width=0.49\textwidth]{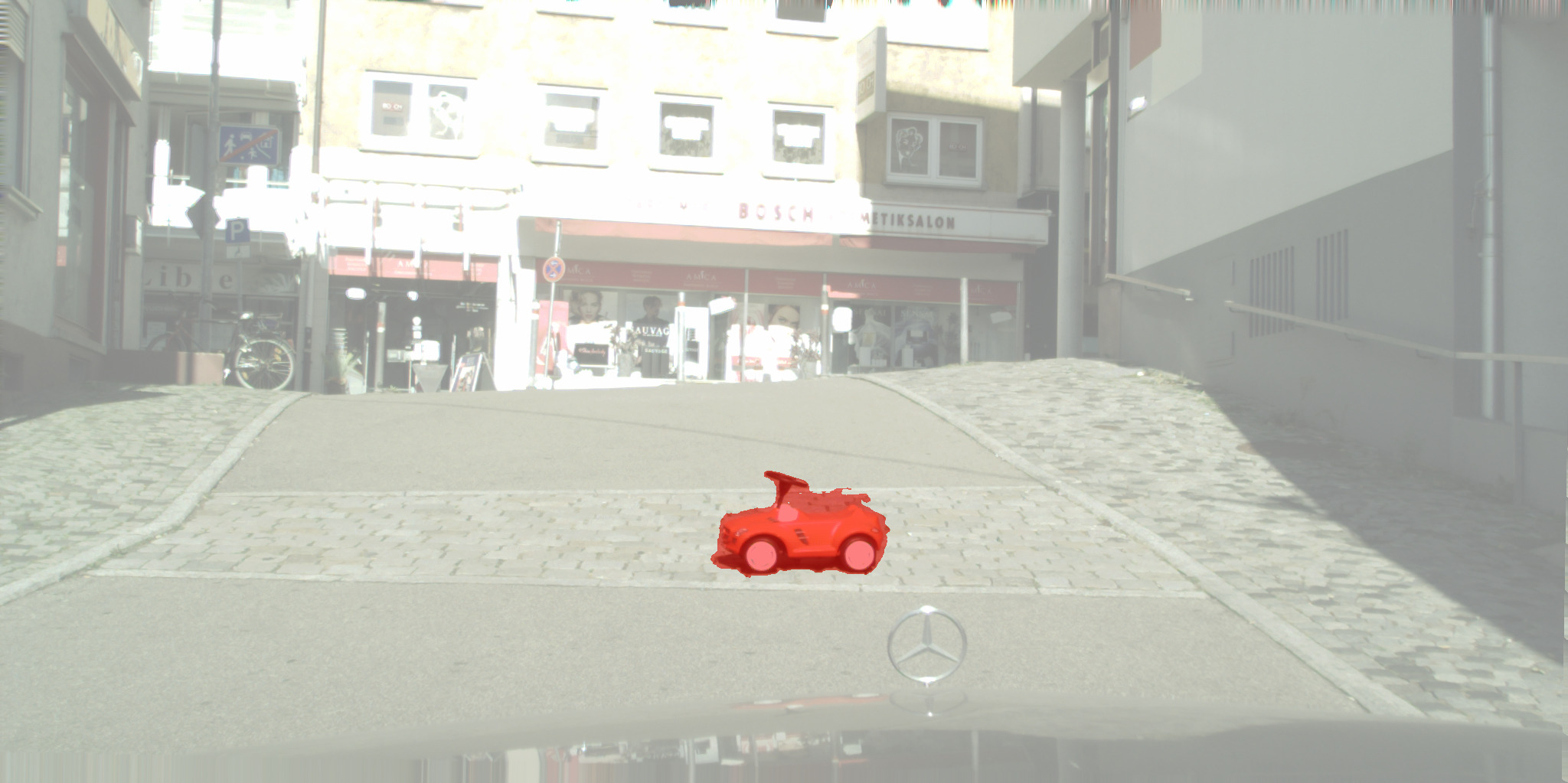}} \\
    \caption{OoD detection for one scene with different combinations of entropy thresholding for the plain model, entropy thresholding after OoD training and meta classification. For all the OoD predictions the same threshold score of $t=0.5$ was used. The red segments indicate OoD object predictions.}
    \label{fig:meta_classif}
    \end{minipage}
    \hfill
    \begin{minipage}[t]{0.47\textwidth}
    \captionsetup[subfigure]{labelformat=empty}
    \centering
    \subfloat[OoD training]{\includegraphics[width=0.49\textwidth]{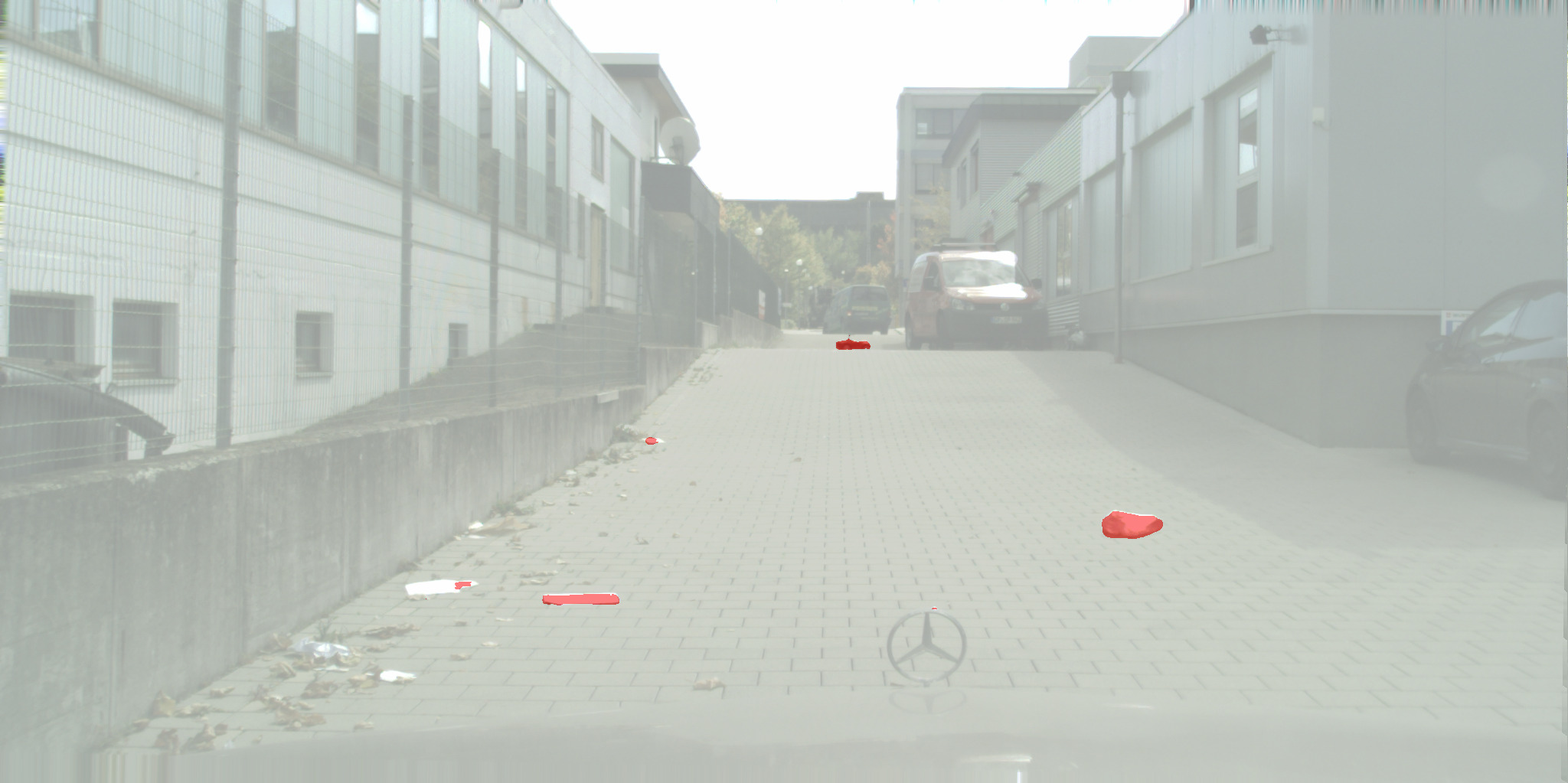}}~
    \subfloat[OoD training + meta classifier]{\includegraphics[width=0.49\textwidth]{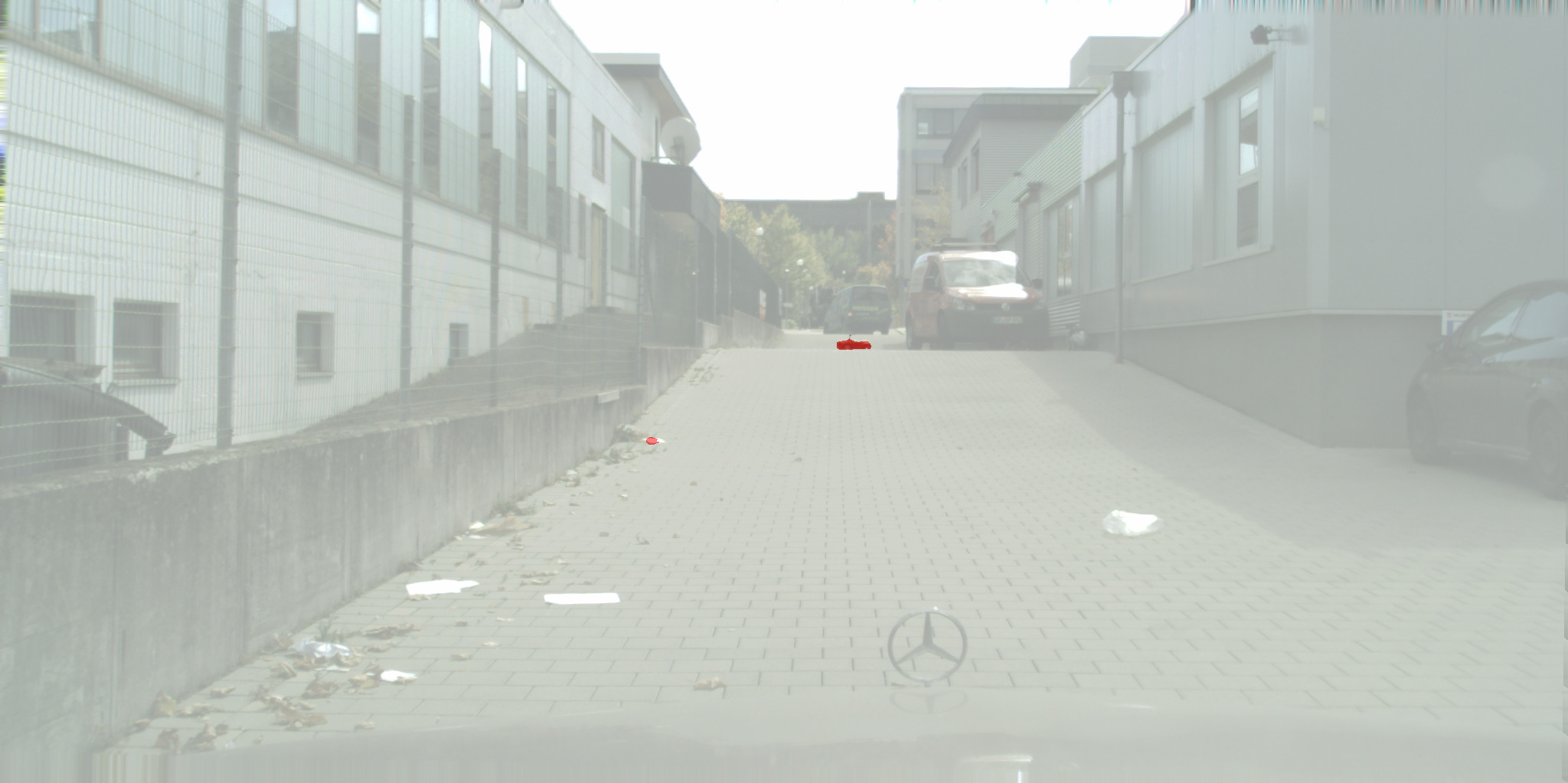}} \\
    \subfloat[OoD training]{\includegraphics[width=0.49\textwidth]{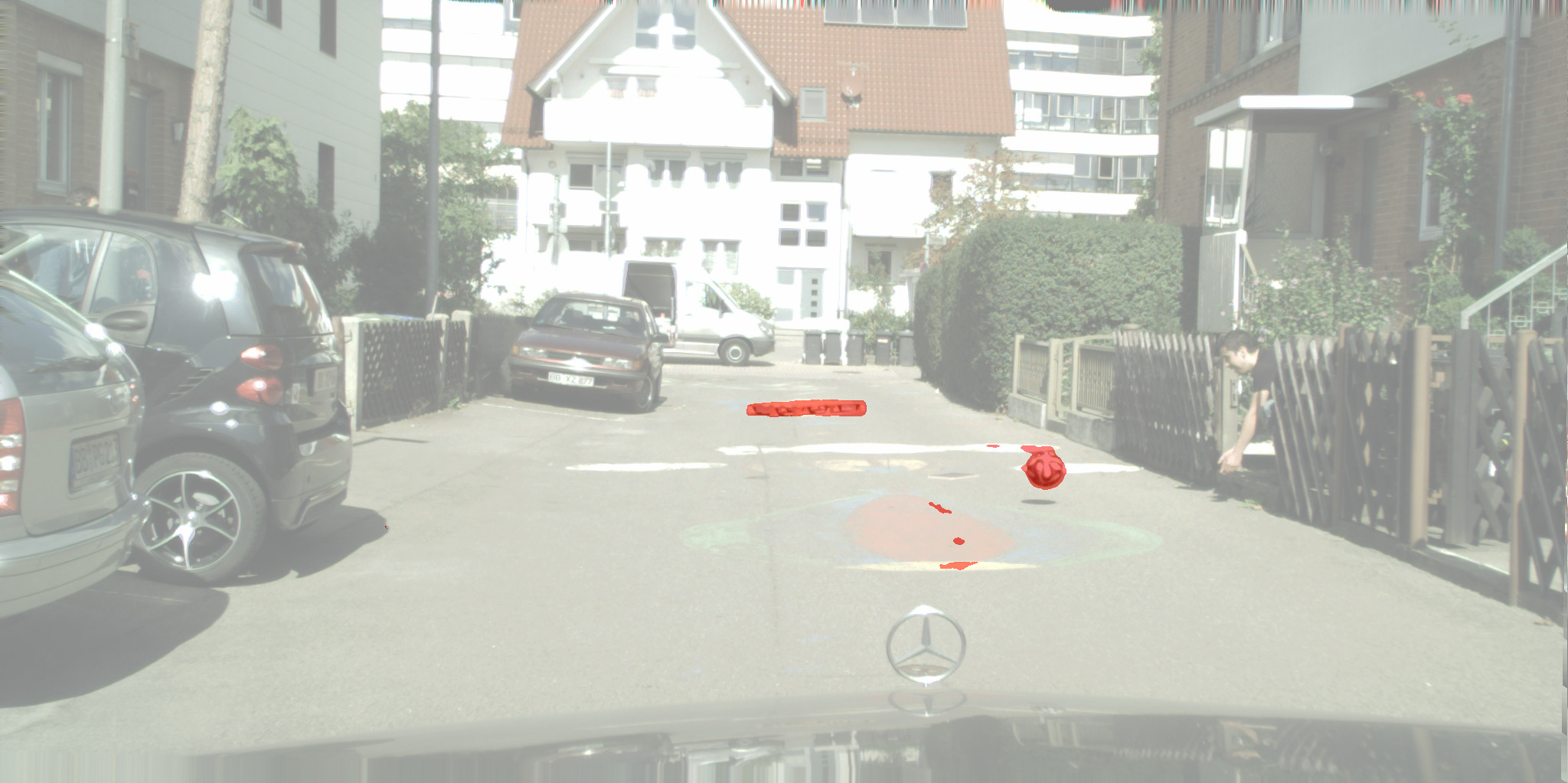}}~
    \subfloat[OoD training + meta classifier]{\includegraphics[width=0.49\textwidth]{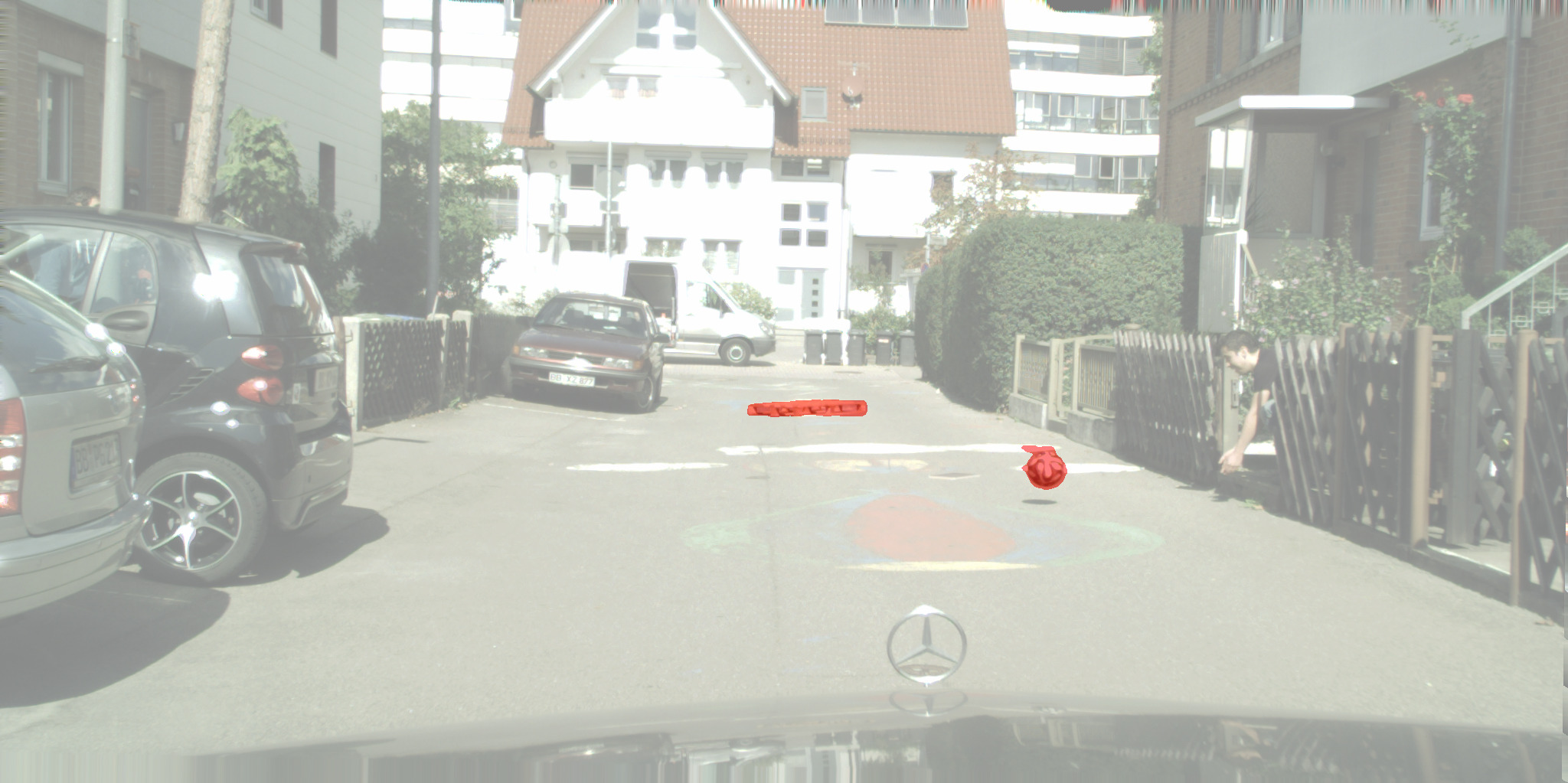}} \\
    \caption{OoD detection performed by the OoD-trained network with and without meta classifiers. The red segments indicate OoD object predictions.}
    \label{fig:meta_classif2}
    \end{minipage}
\end{figure*}

The logistic regressions as meta classifiers have proven their efficiency in identifying and afterwards removing false positive (FP) / incorrect OoD object predictions. In this section we intend to show further examples for the FP OoD removal and thus show the final output of our two-step procedure for OoD detection. 

For the plain model the meta classifiers are already able to remove FP OoD predictions reliably, see \cref{fig:meta_classif} top row. However, some false positive OoD predictions still remain. As pixels with high entropy are more concentrated inside OoD objects after the entropy maximization of the OoD training, the combination of OoD training and meta classification yields the best result in terms of the number of FP OoD predictions, see \cref{fig:meta_classif} bottom row.
The examples in \cref{fig:meta_classif2} further illustrate that the improved OoD detection performance after OoD training can even be enhanced by employing meta classifiers. The removed FP OoD predictions are rather small. However, we already consider one single pixel as FP OoD object prediction if that pixel is incorrectly predicted to be OoD. 
One could also define an OoD prediction to have a minimum amount of pixels. 
As our main focus is the reduction of overlooked OoD objects, we stick to the definition of \cref{eq:seg_tp} and consider an OoD object to be found if at least one pixel of that object is correctly classified as OoD. Therefore, small OoD segments are also fed through the meta classification model. Our two step method, consisting of entropy maximization and meta classification, extends segmentation networks by an improved OoD detection capability and unites both tasks in one model.

\section{Meta Classification Feature Analysis}\label{app:9}

\begin{figure*}[h!]
    \captionsetup[subfigure]{labelformat=empty}
    \centering
    \subfloat[Baseline model only]{\includegraphics[height=4.5cm]{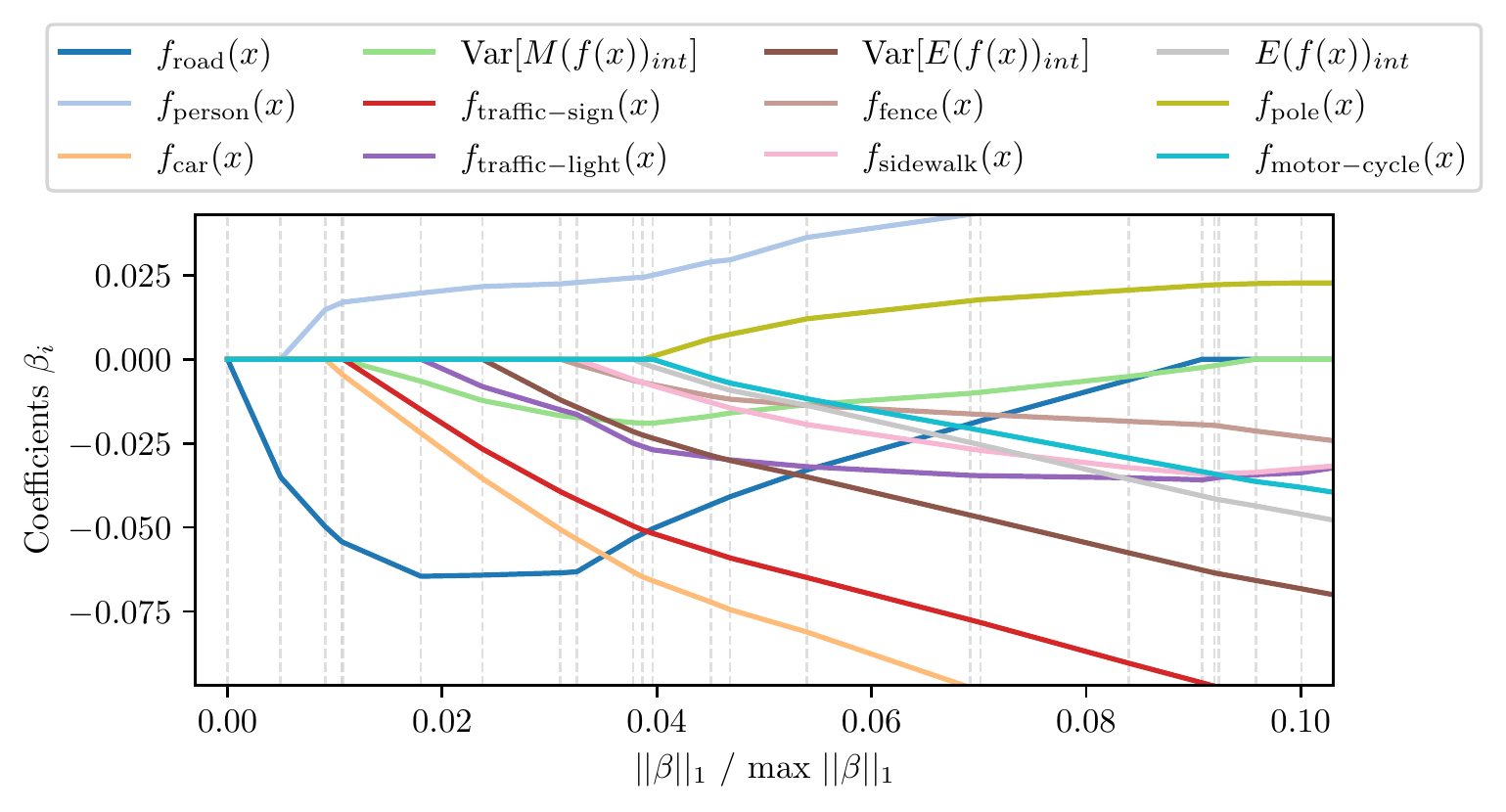}}
    \subfloat[After OoD training]{\includegraphics[height=4.5cm]{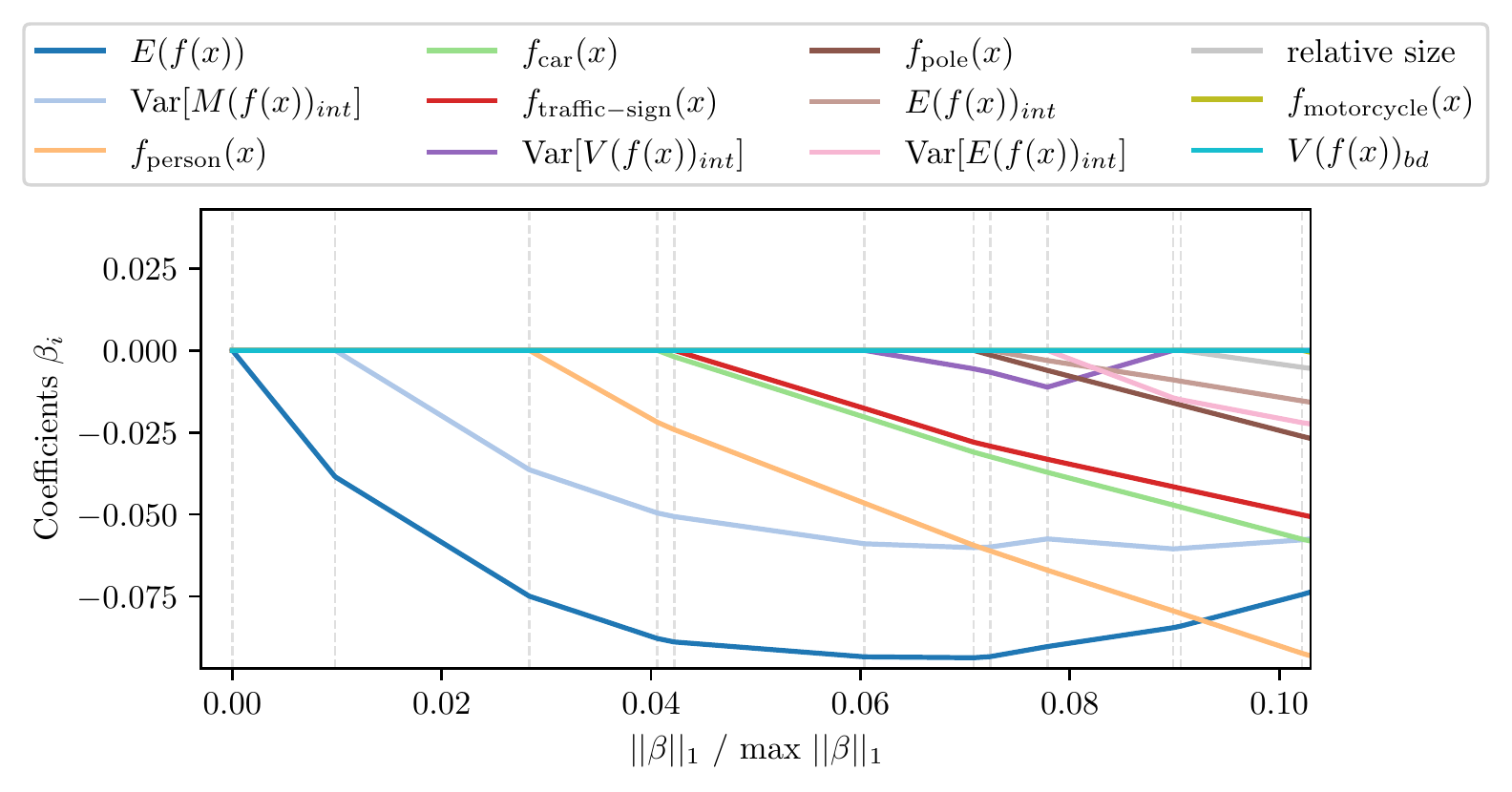}}
    \caption{Least angle regression of the meta classifier applied to the softmax output before (left) and after OoD training (right) is applied. The entropy threshold is set to $t=0.3$ as the experiments showed this threshold yielding the most efficient meta classification models. The 12 meta classification features having the highest correlation are displayed for each model.} \label{fig:lars}
\end{figure*}

Least angle regression (LARS) is a model selection algorithm. We use this algorithm to select the linear model that fits best the meta classification responses $g(\mu)$ subject to an $L_1$ penalty term for the model coefficients $\beta \in \mathbb{R}^{m} = \mathbb{R}^{83}$, \cf \cref{eq:meta-input}. LARS identifies the variables most correlated with the response. This selection method starts with all coefficients $\beta_1 = \ldots = \beta_p=0$. In each step, one variable at a time is added to the set of active predictor variables. Thus, LARS adds the best feature $\mu_i, i=1,\ldots,m$ to include in the active set, \ie features show better correlation with the responses the earlier they are added. The coefficient of the added feature variable is continuously moved from 0 to its least squares coefficient until another variable $\mu_j$ has as much correlation with the response. This procedure is then repeated with the coefficients of the features in the active set, now including $\mu_j$, until the active set reaches a predefined size. 

\Cref{fig:lars} visualizes the coefficients paths in the LARS algorithm for meta classification metrics. The correlation and therefore the impact of metrics are compared when the underlying segmentation CNN is in its plain version and also when OoD training is applied. As entropy threshold $t=0.3$ is chosen as this gives us the best linear models for both meta classification cases with an average precision of $98.84$ and $99.53$, respectively, see \cref{fig:coco-dist}.

We observe that in general features become active later when OoD training is applied, \ie $||\beta||_1 / \max ||\beta||_1$ is greater when the $i$-th variable is added to the active set. This implies that the hand-crafted metrics have higher correlations with the meta classification response and therefore have greater impact on the meta classification performance. After OoD training the entropy $E$ has become the most important metric. The entropy and its variance when restricted to the interior of a segment are among the first ten active variables as well (the eighth and ninth active variable). Without OoD training the softmax probability for the road has the highest correlation, with the entropy metric being the seventh feature in the active set. This analysis shows that the entropy boost due to OoD training has a positive effect on the meta classification performance for entropy based OoD object predictions.

Using linear models as meta classifiers, such as logistic regressions in our case, allows us to track each variable of the meta model. LARS is an effective way of analyzing the correlation of variables with the respective response in linear models. Besides being a very lightweight, such a meta classifier contributes as monitoring method to safer as well as more transparent deep learning applications.

\section{OoD Detection Methods Description and Run-time comparison}
After describing the applied OoD detection methods of our experiments in more detail, we provide a comparison of inference times in order to judge the methods' suitability as an online application.

\subsection{Methods}
Most methods for OoD detection perform on image-level. However, the state-of-the-art methods for OoD detection can be adapted to semantic segmentation in a straight-forward manner. As a reminder, we denote the pixel-wise softmax probability at pixel location $z\in\mathcal{Z}$ with $f^z(x)\in (0,1)^{|\mathcal{C}|}$ for an image $x\in\mathcal{X}$, see also \cref{sec:ood_seg}.

\paragraph{Maximum softmax probability.} The pixel-wise maximum softmax probability is a commonly used baseline for OoD detection. We apply this metric as OoD score for each pixel $z\in\mathcal{Z}$:
\begin{equation}
     1 - \max_{j\in\mathcal{C}} f_j^z(x) = 1 - f_{\hat{c}(z)}^z(x),~ x\in\mathcal{X} ~ . \label{eq:msp}
\end{equation}

\paragraph{ODIN.}
Let $tau\in\mathbb{R} \setminus \{0\}$ be a temperature scaling parameter and $\delta\in\mathbb{R}$ a perturbation magnitude. We first add small perturbations to each pixel $z\in\mathcal{Z}$ of image $x\in\mathcal{X}$: 
\begin{equation}
\tilde{x}^z = x^z - \delta \text{sign}\left( - \frac{\partial}{\partial x^z} \log f_{j^*}^z(x) \right)\,. 
\end{equation}
Then, the OoD score is obtained similar to the maximum softmax probability:
\begin{equation}
    1 - \max_{j\in\mathcal{C}} f_j^z(\tilde{x})/\tau ~.
\end{equation}

\paragraph{Mahalanobis distance.}
Let $h(\cdot)$ denote the output of the penultimate layer of a CNN. Under the assumption that $h(\cdot)$ is a class conditional Gaussian, \ie
\begin{equation} \label{eq:gauss}
    P(h^z(x)~|~y^z(x)=j) = \mathcal{N}(h^z(x)~|~\mu_j, \Sigma_j)~\forall x\in\mathcal{X}
\end{equation}
we compute the Mahalanobis distance as OoD score for each pixel $z \in \mathcal{Z}$:
\begin{equation}
    \min\limits_{j\in\mathcal{C}} \left\{ (h^z(x) - \hat{\mu}_j)^T {\hat{\Sigma}}^{-1}_j (h^z(x) - \hat{\mu}_j) \right\}
\end{equation}
where $\hat{\mu}_j$ and $\hat{\Sigma}_j$ are estimates for class mean $\mu_j$ and class covariance $\Sigma_j$, respectively, of the latent features in the penultimate layer (see \cref{eq:gauss}). 

\paragraph{Monte Carlo dropout.} Let $S\in\mathbb{N}$ denote the number of Monte Carlo sampling rounds and let $\hat{p}_j^z(x) = (f_j^z(x))_{s=1}^S$ denote the softmax probabilities of class $j\in\mathcal{C}$ for samples $s\in \{ 1,\ldots, S\}$. 
We consider the sum of variances of each class as OoD score for each pixel $z \in \mathcal{Z}$:
\begin{equation}
    \sum_{j\in\mathcal{C}}\mathrm{Var}\left(\hat{p}_j^z(x)\right), x\in\mathcal{X}
\end{equation}
where $\mathrm{Var}(\cdot)$ is the empirical variance function. Regarding Monte Carlo dropout as baseline, we conducted experiments also with the mutual information. However, we observed worse anomaly detection performance compared to the sum over variances.

\subsection{Inference Time Comparison}
Methods that estimate uncertainty are relevant for many applications involving deep learning. In practice, monitoring systems need to compute uncertainty in real time in order to provide online applicability. Therefore, one crucial factor is the run-time of OoD detection methods. In this subsection we compare the inference time, \ie the time from feeding an image through a model to obtaining pixel-wise OoD scores. We report the results in \cref{tab:run-time}. For reasons of comparison, we choose the same input and the same segmentation network architecture for all evaluated methods. We observe that our OoD training approach is highly efficient in terms of run time, only being outperformed by the weak maximum softmax probability baseline. However, the gap is less than one second. Compared to the remaining methods, the time difference is more substantial, ranging from 17 seconds for ODIN up to 70 seconds for the Mahalanobis distance.

\begin{table}[t]
    \begin{center}
    \scalebox{0.8}{
    \begin{tabular}{l|r}
    \toprule
    & time in s $\downarrow$ \\
    OoD detection method & per image \\ 
    \midrule
    \midrule
    Maximum Softmax & \textbf{1.52} \\
    Entropy Thresholding & 2.39 \\
    ODIN  & 19.63 \\
    Monte Carlo Dropout & 33.48 \\
    Mahalanobis Distance & 72.54 \\
    \rowcolor{Gray} Ours: OoD training + entropy thresholding  & 2.38 \\
    \midrule
    \end{tabular}
    }
    \end{center}
    \caption{Run time comparison of the different OoD detection methods. The inference time for one image is reported in seconds. For all methods the same input as well as the same underlying segmentation network is used.}\label{tab:run-time}
\end{table}

\section{Entropy Maximization with Different Seeds}
For entropy maximization, a subset of images from the COCO dataset is used. This subset is randomly sampled, \cf \cref{sec:setup}. To investigate how different random seeds affect the presented results of this work, we included average performance scores over multiple seeds and the corresponding standard deviations in \cref{tab:std-dev}, where we report the major metric for all considered datasets. In these additional experiments, we even observe scores better than those reported in the paper.

\begin{table}[t]
    \begin{center}
    \scalebox{0.8}{
    \begin{tabular}{l|c|c|c}
    \toprule
    & LaF test AUPRC & Fishy val AUPRC & City val mIoU\\ 
    \midrule
    \midrule
    \rowcolor{Gray} OoD train & $\mathbf{72.58} \pm 04.30$ & $\mathbf{73.51} \pm 06.12$ & $\mathbf{88.95} \pm 00.41$ \\
    Baseline & $46.00 \pm 00.00$ & $27.54 \pm 00.00 $ & $90.30 \pm 00.00$ \\
    \midrule
    \end{tabular}
    }
    \end{center}
    \vspace{-.5cm}
    \caption{Averaged performance scores over 8 random seeds (for COCO subsampling) and their corresponding standard deviations on LostAndFound, Fishyscapes and Cityscapes. For each run, 10 epochs of OoD training were executed with DeepLabV3+ and $\lambda=0.9$.}\label{tab:std-dev}
\end{table}

\end{appendices}

\end{document}